\pdfoutput=1

\documentclass[11pt]{article}

\makeatletter
\providecommand{\@LN@col}[1]{}
\providecommand{\@LN}[2]{}
\makeatother

\usepackage{style}

\usepackage[T1]{fontenc}
\usepackage[utf8]{inputenc}
\usepackage{times}
\usepackage{latexsym}
\usepackage{microtype}
\usepackage{inconsolata}

\usepackage{booktabs}
\usepackage{longtable}
\usepackage{multirow}
\usepackage{array}
\usepackage{tabularx}
\usepackage{amsmath}
\usepackage{amsfonts}
\usepackage{amssymb}
\usepackage{graphicx}
\graphicspath{{images/}}
\usepackage{subcaption}
\usepackage{float}
\usepackage{afterpage}
\usepackage{wrapfig}

\usepackage[table]{xcolor}
\usepackage{colortbl}
\usepackage{listings}
\usepackage{tcolorbox}
\usepackage{enumitem}

\usepackage{tikz}
\usepackage{pgfplots}
\usetikzlibrary{matrix}
\pgfplotsset{compat=1.18}

\usepackage{natbib}
\usepackage[colorlinks=true,linkcolor=black,citecolor=black,urlcolor=black]{hyperref}
\usepackage{url}
\usepackage{wrapfig}
\usepackage{booktabs}
\usepackage{wrapfig}
\usepackage{caption}

\title{Do VLMs Reason Like Engineers? A Benchmark and a Stage-wise Evaluation}

\author{%
  {\bfseries
  Syed Wasiq\textsuperscript{*},
  Syed Mohamad Tawseeq\textsuperscript{*},
  Yashwant Pravinrao Bangde,
  and Debaditya Roy}\\[0.75em]
  \texttt{syedwasiq12@kgpian.iitkgp.ac.in},
  \texttt{tawseeq@kgpian.iitkgp.ac.in},\\
  \texttt{yashwant@kgpian.iitkgp.ac.in},
  \texttt{debaditya@cse.iitkgp.ac.in}\\[0.5em]
  Indian Institute of Technology Kharagpur\\
  \thanks{Equal contribution (author order determined by a coin toss)}%
}

\begin{document}
\maketitle
\begin{abstract}
Vision-Language Models (VLMs) demonstrate strong performance on general multimodal reasoning benchmarks, yet their ability to perform engineering reasoning remains largely unexplored. Unlike general visual question answering, engineering problem solving requires interpreting technical diagrams, selecting governing physical principles, and maintaining physically consistent multi-step reasoning. These capabilities are increasingly important for AI systems used in engineering education, scientific assistance, and technical decision-making, where reasoning failures may produce physically invalid yet superficially plausible solutions. Existing benchmarks primarily evaluate final answers and provide limited assessment of intermediate reasoning processes. We introduce \textbf{EngVQA}, a multimodal benchmark for evaluating engineering reasoning across 5 engineering subjects containing 696 problems. We introduce an 8-stage automatic evaluation framework for assessing VLM-generated solutions. 
The framework independently evaluates each stage of the solution, enabling fine-grained analysis of reasoning failures.
We benchmark multiple state-of-the-art open and closed source VLMs on our evaluation framework and demonstrate substantial limitations in current engineering reasoning capabilities. Human evaluation shows strong agreement with our automated framework, achieving a Pearson correlation of 0.975 and a mean absolute error of 0.67 on a 10-point grading scale. Our results highlight the importance of process-oriented evaluation for reliable assessment of multimodal engineering reasoning systems.
\end{abstract}

\section{Introduction}
\label{sec:introduction}

Recent Vision-Language Models (VLMs) such as GPT-4V \citep{achiam-et-al:gpt}, and Gemini Pro \citep{geminiteam2025geminifamilyhighlycapable} have significantly advanced multimodal reasoning and visual understanding. However, existing benchmarks including VQA \citep{agrawal2016vqavisualquestionanswering}, GQA \citep{8953451}, ScienceQA \citep{lu2022learn}, and MathVista \citep{lu2024mathvistaevaluatingmathematicalreasoning} primarily evaluate symbolic reasoning and final-answer correctness, providing limited assessment of physically grounded engineering reasoning involving technical diagrams, governing equations, and multi-stage analytical workflows.

Recent engineering-oriented benchmarks such as EngiBench \citep{zhou2026engibenchbenchmarkevaluatinglarge}, EEE-Bench \citep{11094253},  move toward engineering-oriented evaluation,
yet important limitations remain. EngiBench primarily
focuses on textual engineering reasoning and
capability-level rubric evaluation without explicit
stage-wise decomposition of intermediate reasoning
stages. EEE-Bench emphasizes multimodal understanding
in electrical and electronics engineering, but still
largely evaluates final-answer correctness rather
than structured engineering workflows. Recent reasoning-aware and process-oriented
evaluation frameworks including G-Eval \citep{liu-etal-2023-g}, Prometheus \citep{kim2024prometheusinducingfinegrainedevaluation}, ProcessBench \citep{zheng2025processbenchidentifyingprocesserrors}, and Thinking-LLM-as-a-Judge~\citep{saha2025learningplanreason}
demonstrate that structured reasoning-aware
evaluation improves automated assessment reliability.

To address these limitations, we introduce \textbf{EngVQA}, a benchmark of authentic engineering problems spanning 5 subjects: Fluid Mechanics, Heat and Mass Transfer, Dynamics, Mechanics of Materials, and Thermodynamics. The benchmark requires joint reasoning over technical diagrams, physical principles, symbolic derivations, and multi-step quantitative analysis.

\begin{table}[!htbp]
\centering
\small

\begin{tabular}{lcccccc}
\toprule

\textbf{Benchmark}
&
\textbf{Engineering}
&
\textbf{Multimodal}
&
\textbf{Technical}
&
\textbf{Process}
&
\textbf{Stage-wise}
&
\textbf{Physics}
\\

&
\textbf{Reasoning}
&
\textbf{Reasoning}
&
\textbf{Diagrams}
&
\textbf{Evaluation}
&
\textbf{Evaluation}
&
\textbf{Constraints}
\\

\midrule

MMMU
& Partial & Yes & Partial & No & No & No \\

SciBench
& No & Limited & No & No & No & No \\

EngiBench
& Yes & No & No & Partial & No & Yes \\

EEE-Bench
& Yes & Yes & Yes & No & No & Partial \\

SeePhys
& Partial & Yes & Yes & No & No & No \\
\midrule
\textbf{EngVQA + EngJudge}
& Yes
& Yes
& Yes
& Yes
& Yes
& Yes
\\
\bottomrule
\end{tabular}
\vspace{0.3cm}
\caption{
Comparison of multimodal scientific and engineering reasoning benchmarks along dimensions of engineering realism, technical diagram understanding, process-level evaluation, and physics-aware reasoning constraints.
}
\label{tab:benchmark_comparison}
\end{table}

Building on the benchmark, we propose \textbf{EngJudge}, an 8-stage process-oriented evaluation framework that decomposes engineering solutions into interpretable reasoning stages while modeling dependency-aware error propagation across interrelated reasoning stages. EngJudge independently evaluates localized reasoning stages, improving interpretability and reducing evaluator ambiguity compared to holistic evaluation approaches. 

To validate the reliability of EngJudge, we also conduct a human validation study with engineering students. Our findings show that the framework's automated scores align closely with human expert grading philosophies, demonstrating its potential as a reliable tool for structured, process-oriented evaluation. Table~\ref{tab:benchmark_comparison} summarizes the differences between existing benchmarks and our approach. Our contributions are as follows:












\begin{itemize}

\item We introduce \textbf{EngVQA}, a multimodal benchmark of 696 authentic engineering problems requiring reasoning over technical diagrams, physical principles, symbolic derivations, and multi-step quantitative analysis.


\item We propose \textbf{EngJudge}, an 8-stage process-oriented evaluation framework that models dependency-aware reasoning failures and achieves strong agreement with expert human evaluators.


\item We demonstrate that SOTA VLMs exhibit substantial weaknesses in engineering reasoning, particularly in diagram interpretation, equation selection, assumption validation, and physically consistent multi-stage analysis.

\end{itemize}

\section{Related Work}
\label{sec:related_work}
\paragraph{Engineering and Scientific Reasoning Benchmarks}

Recent benchmarks have explored scientific and engineering reasoning in large language and vision-language models across diverse domains. General multimodal reasoning benchmarks such as MMMU~\citep{10656299} evaluate broad-domain visual reasoning across university-level subjects, while scientific reasoning datasets such as SciBench~\citep{SciBench}, ScienceQA~\citep{lu2022learn}, and MathVista~\citep{lu2024mathvistaevaluatingmathematicalreasoning} focus primarily on scientific problem solving, symbolic reasoning, and final-answer correctness. 
More recent engineering-oriented benchmarks have extended multimodal evaluation into STEM and applied engineering domains. EngiBench \citep{zhou2026engibenchbenchmarkevaluatinglarge} evaluates engineering-focused question answering tasks, while EEE-Bench~\citep{11094253} introduces multimodal reasoning problems in electrical engineering involving circuit diagrams and technical schematics. SeePhys \citep{xiang2025seephysdoesseeinghelp} studies visually grounded physics reasoning, and CSVQA~\citep{jian2025csvqachinesemultimodalbenchmark} explores multimodal STEM reasoning in educational settings. Existing benchmarks predominantly evaluate reasoning through final-answer correctness or holistic solution-level scoring.
\vspace{-0.20cm}
\paragraph{Stage-wise Reasoning and Process Evaluation}

Recent work has shown that final-answer correctness alone is insufficient for evaluating reasoning quality in modern language and vision-language models \citep{Lightman2023LetsVS, golovneva2023roscoe}. As models increasingly generate long-form chain-of-thought reasoning, evaluating intermediate reasoning behavior has become important for understanding logical consistency, factual correctness, and reasoning reliability.

ROSCOE~\citep{golovneva2023roscoe} introduces fine-grained metrics for evaluating
generated reasoning traces across semantic and
logical consistency dimensions. ~\cite{Lightman2023LetsVS} demonstrate that process supervision can improve mathematical reasoning by rewarding intermediate reasoning correctness rather than relying solely on final answers. ProcessBench~\citep{zheng2025processbenchidentifyingprocesserrors} studies process-level reasoning failures in mathematical settings through stage-wise verification of intermediate reasoning chains.
\paragraph{LLM-as-a-Judge and Structured Evaluation}

Recent work has increasingly explored the use of large language models as automated evaluators for reasoning and generation tasks \citep{10.5555/3666122.3668142, liu2023geval, kim2024prometheusinducingfinegrainedevaluation}. G-Eval \citep{liu2023geval} demonstrates that rubric-guided evaluation can improve the reliability and interpretability of LLM-based assessment, while Prometheus \citep{kim2024prometheusinducingfinegrainedevaluation} explores fine-grained rubric-conditioned evaluation strategies for scalable automatic assessment. More recent frameworks investigate structured reasoning strategies for evaluation itself. Thinking-LLM-as-a-Judge \citep{saha2025learningplanreason} proposes planning-oriented judging strategies in which evaluators explicitly reason through structured evaluation plans before assigning scores.

\section{EngVQA Benchmark}
\label{sec:benchmark}
\begin{table}[!htbp]
\centering
\small
\renewcommand{\arraystretch}{1.1}

\begin{tabular}{lccll}
\toprule
\textbf{Subject} &
\textbf{\#Qs} &
\textbf{ATPQ} &
\textbf{Representative Topics (Total)} &
\textbf{Diagram Types}
\\
\midrule

Dyn.
&
171
&
3.57
&
Rigid-body motion, impact, vibration (54)
&
FBDs, kinematic schematics
\\

Thermo.
&
236
&
9.28
&
Cycles, entropy, steady-flow systems (53)
&
Property plots, thermodynamic systems
\\

FM
&
93
&
5.81
&
Pipe flow, hydrostatics, drag/lift (59) 
&
Flow schematics, pressure diagrams
\\

HMT
&
96
&
13.89
&
Conduction, convection, diffusion (60)
&
Thermo circuits, boundary-layer diagrams
\\

MoM
&
100
&
9.46
&
Stress analysis, torsion, buckling (60)
&
Beam loading, stress distributions
\\

\midrule
\textbf{Total}
&
\textbf{696}
&
\multicolumn{3}{c}{Multimodal engineering reasoning problems}
\\

\bottomrule
\end{tabular}
\vspace{0.3cm}
\caption{
Subject-wise statistics of EngVQA. ATPQ (Average Topics per question), Fluid Mechanics
(FM), Heat and Mass Transfer (HMT), Mechanics of Materials (MoM), Thermodynamics (Thermo), and Dynamics (Dyn). 
}
\label{tab:subject_stats}
\end{table}

\subsection{Benchmark Principles}
We chose problems-solution pairs that align with EngVQA's benchmark design principles: (1) \textbf{Diagram-Grounded Analytical Reasoning:} Problems require extracting geometry, boundary conditions, force directions, flow structure, material interfaces, and spatial constraints directly from technical figures such as free-body diagrams, thermodynamic plots, flow schematics, stress distributions, and engineering layouts. (2) \textbf{Structured Multi-Stage Reasoning:}
Solutions involve multiple interdependent reasoning stages including problem characterization, assumption formulation, visual interpretation, equation selection, symbolic derivation, algebraic computation, and physical validation. (3) \textbf{Physics-Constrained Engineering Workflows:}
Problems require physically valid reasoning under domain-specific engineering constraints. Models must maintain consistency between visual interpretation, governing equations, simplifying assumptions, and final quantitative predictions throughout the solution process.

\subsection{Benchmark Statistics}

Table~\ref{tab:subject_stats} summarizes the subject composition and reasoning diversity of EngVQA. The benchmark spans five foundational engineering subjects and contains 696 problems requiring multimodal reasoning over technical diagrams, governing equations, symbolic derivations, and physical constraints. Rather than concentrating on a small set of narrow
problem templates, the benchmark includes diverse
question distributions across different topics.  The subject distribution reflects both breadth and diversity in engineering reasoning workflows. Dynamics and Mechanics of Materials emphasize free-body analysis, force interactions, and rigid-body reasoning, while Thermodynamics and Heat \& Mass Transfer involve physically constrained energy-system analysis, property relationships, and transport phenomena. Fluid Mechanics problems additionally require spatial reasoning over flow structures, pressure distributions, and conservation laws. Across all subjects, technical diagrams play a central role in downstream analytical formulation, making visual interpretation a necessary component of successful problem solving rather than an auxiliary context. The average number of topics per question measures reasoning density across the domains. This metric underscores that solving a typical problem in our benchmark requires the simultaneous integration and synthesis of multiple physical concepts, which fundamentally heightens the difficulty of problems. Full list of topics per subject in Appendix \ref{app:benchmark_details}.

\section{EngJudge: Stage-wise Evaluation}
\label{sec:eng_judge}

Evaluating engineering reasoning requires more than final-answer correctness because failures often emerge within intermediate reasoning stages such as assumption formulation, diagram interpretation, equation selection, and computation. To capture this process-level behavior, we develop \textbf{EngJudge}, a stage-wise evaluation framework motivated by an error analysis of \texttt{gemini-2.0-flash-exp} solutions across 3 subjects - Fluid Mechanics, Heat and Mass Transfer (HMT), and Mechanics of Materials (MOM).

Our analysis shows that engineering reasoning failures are highly multifaceted and rarely occur in isolation. We additionally observe that visual interpretation forms a structurally distinct failure mode, while error correlations suggest that engineering reasoning is best modeled as a \textit{partially dependent} process rather than a fully independent one (Appendix~\ref{appendix:error_analysis}). These findings motivate two key design choices in EngJudge: independent evaluation of localized reasoning stages and dependency propagation along empirically observed reasoning edges. 
\begin{figure}
    \centering
    \includegraphics[width=0.85\linewidth]{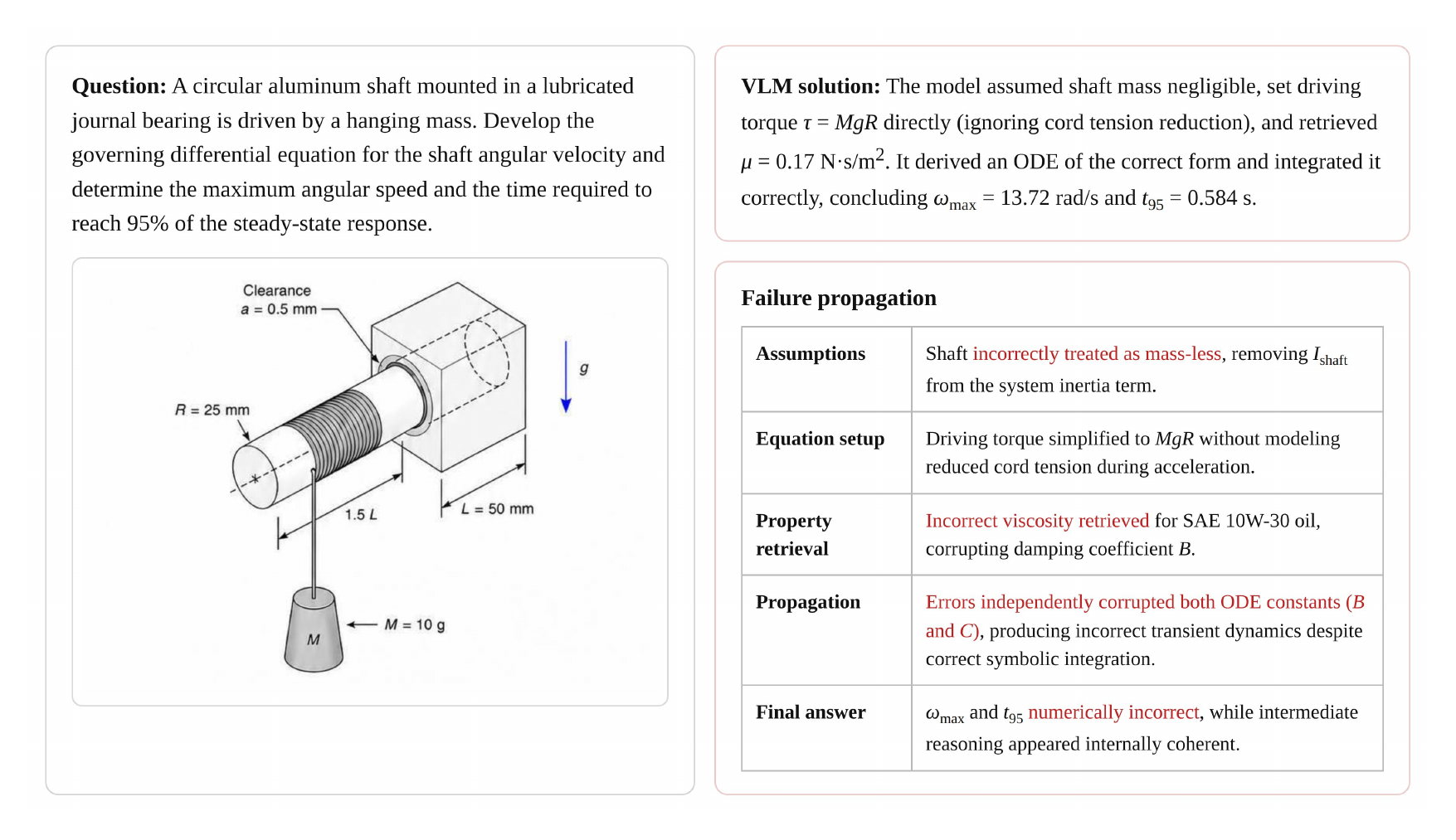}
    \vspace{-0.3cm}
    \caption{A representative example problem with LLM generated solution showing error propagation from incorrect assumptions affects the rest of the solution.}
    \label{fig:annotated_example}
\end{figure}

Figure~\ref{fig:annotated_example} illustrates a failure case in a fluid mechanics problem. Here, a simple step-by-step average metric would penalize the integration step for having incorrect numerical coefficients, failing to recognize that the math was internally coherent but corrupted by prior errors. By decoupling these stages and tracking the failure chain ($\text{Assumptions} \to \text{Equation Selection} \to \text{Algebraic Accuracy} \to \text{Final Answer}$), EngJudge pinpoints the root cause of failure in physical modeling rather than mathematical execution.

\subsection{Evaluation Framework}
\label{sec:evaluation}
\begin{figure}[t]
    \centering
    \includegraphics[width=\textwidth]{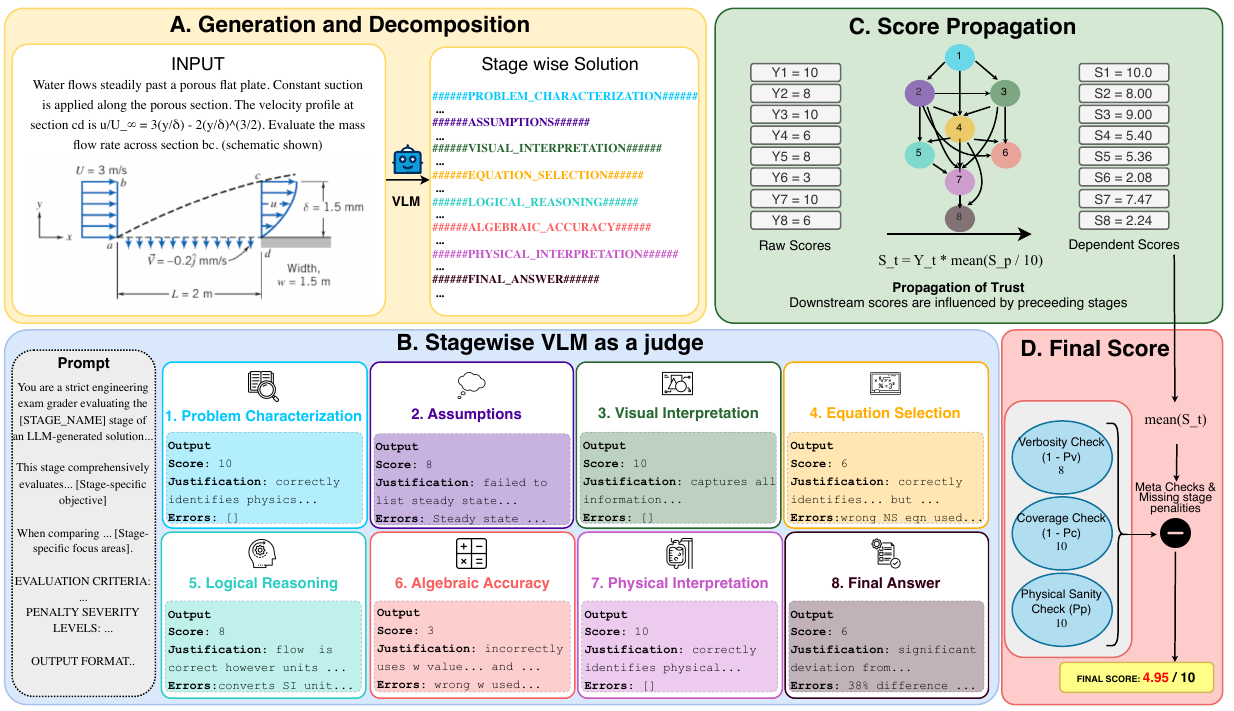}
    \caption{Overview of the proposed EngJudge evaluation framework. A. VLM generates a structured step-wise solution for an engineering problem containing both text and technical diagram. B. Solution is decomposed into eight reasoning stages, each independently evaluated using rubric-guided LLM-as-a-judge prompts (App~\ref{app:evaluation_prompts}) with penalty-based scoring and fatal-error detection. C. Stage scores are then propagated through a dependency graph (colors correspond to steps in B.) and D. aggregated using meta-evaluation checks to produce a final interpretable score on a 0–10 scale.}
    \label{fig:pipeline}
\end{figure}

We introduce a multi-stage, penalty-based automatic evaluation framework, \textbf{EngJudge}, for grading LLM-generated solutions to graduate-level engineering problems. Rather than relying solely on holistic or
single-score evaluation~\cite{10.5555/3666122.3668142, liu2023geval}, our framework decomposes each solution into eight structured reasoning stages and evaluates them individually through an LLM-as-a-judge. The framework combines \textbf{penalty-based stage scoring}, \textbf{graph-based dependency propagation}, \textbf{meta-evaluation checks}, and \textbf{exam-style partial credit}~\citep{mertler2001designing} to produce a single interpretable score on a 0--10 scale. The full pipeline is illustrated in Figure~\ref{fig:pipeline}. The evaluation pipeline consists of the following stages:

\paragraph{Solution Parsing:}
We prompt the LLM to generate solutions using a
fixed 8-stage reasoning structure designed to
separate different components of engineering
problem-solving, consistent with chain-of-thought prompting conventions~\cite{chain_of_thought2022, 10.5555/3600270.3601883}. Each stage is enclosed within
explicit tags so that the generated solution can be
parsed automatically (See Figure\hyperref[fig:pipeline]{~\ref*{fig:pipeline}A}).

\begin{tcolorbox}[
colback=gray!3,
colframe=black!25,
boxrule=0.3pt,
arc=1.5mm,
left=1mm,
right=1mm,
top=0.8mm,
bottom=0.8mm,
width=0.95\textwidth
]
\footnotesize
\#\#\#\#\#\# STAGE\_NAME \#\#\#\#\#\# \newline
[stage content] \newline
\#\#\#\#\#\# END\_STAGE \#\#\#\#\#\#
\end{tcolorbox}

\noindent A rule-based parser extracts each tagged block and
produces an ordered list of reasoning stages.
The required stages are:
Problem Characterization,
Assumptions,
Visual Interpretation,
Equation Selection,
Logical Reasoning,
Algebraic Accuracy,
Physical Interpretation,
and Final Answer.
\paragraph{Stage-wise Evaluation:}
Each reasoning stage is evaluated independently using
a dedicated LLM-as-a-judge prompt (see Appendix~\ref{app:prompts}) tailored to that
stage. Every stage begins with a score of 10, and
penalties are deducted for detected reasoning errors.
The penalties are distributed into four main categories shown in (Table~\ref{tab:penalties}).

\begin{wraptable}{r}{0.5\columnwidth}
\vspace{-10pt}
\centering
\small
\begin{tabular}{p{1.2cm}p{0.8cm}p{3.6cm}}
\toprule
\textbf{Severity} & \textbf{Penalty} & \textbf{Examples} \\
\midrule
Minor    & 2  & Small rounding differences \\
Moderate & 4  & Unit conversion errors \\
Major    & 7  & Wrong boundary conditions \\
Critical & 10 & Dimensional inconsistency \\
\bottomrule
\end{tabular}
\caption{Penalty severity scale.}
\label{tab:penalties}
\vspace{-10pt}
\end{wraptable}

\noindent The raw stage score ($Y_t$) is computed as:
\begin{equation}
Y_t = \max\!\left(0,\ 10 - \textstyle\sum_{i} p_i\right)
\label{eq:raw_score}
\end{equation}
where $p_i$ is the penalty assigned to the $i^{th}$ identified error in stage $t$.



\paragraph{Fatal Errors:}
Certain critical mistakes cap the maximum score for a
stage regardless of the accumulated penalties.
For example, if the governing equation is fundamentally
incorrect, the \texttt{Equation Selection} stage
immediately receives a score of 0. Similarly,
dimensionally inconsistent formulations or physically
impossible results apply fixed score caps (See Appendix~\ref{app:fatal_error_caps}).

For the \texttt{Final Answer} stage, predicted numerical
values are compared against the ground-truth solution
using relative error. Errors within 5\% receive no
penalty, errors between 5\% and 10\% receive a
moderate penalty, and errors above 10\% receive a
major penalty. Missing units, incomplete answers,
or physically impossible results (e.g., negative
absolute temperature) receive additional penalties
or score caps.

\paragraph{Graph-Based Dependency Propagation:}
\label{ssec:graph}

After computing the stage-wise scores, we propagate
upstream reasoning failures through a dependency
graph. This design is motivated by the error
analysis discussed in Section~\ref{sec:eng_judge},
which showed that many engineering reasoning errors
propagate across stages.

We model these relationships using the dependency
graph shown in Figure\hyperref[fig:pipeline]{~\ref*{fig:pipeline}C}.
Each stage score is adjusted using the scores of its
parent stages.
Let $Y_t$ denote the raw score of stage $t$ obtained
from the penalty-based evaluation. Let $P(t)$ denote
the set of parent stages connected to stage $t$, and
let $N_t$ be the number of parent stages ($N_t = \text{len}(P(t))$).
Let $S_t$ and $S_p$ denote the final propagated scores
of stage $t$ and its parent stage(s) $p \in P(t)$, respectively.
For stages with no parents i.e. Stage 1, we set $S_1 = Y_1$.
For all other stages, the dependency score is computed as:
\begin{equation}
S_t =
Y_t
\times
\frac{1}{N_t}
\sum_{p \in P(t)}
\frac{S_p}{10}
\label{eq:dep_score}
\end{equation}

\paragraph{Score Aggregation and Final Penalty Gates:}
\label{ssec:aggregation}
The final score is computed from the dependency-aware
stage scores obtained after propagation. Let $S_t$
denote the propagated score for stage $t$, and let
$N$ denote the number of reasoning stages. The base
score is computed as:

\begin{equation}
S_{\text{base}}
=
\frac{1}{N}
\sum_{t=1}^{N} S_t
\end{equation}

Missing reasoning stages receive fixed deductions: 

\begin{equation}
S_{\text{blend}}
=
\max(0,\,
S_{\text{base}} - \text{MP})
\end{equation}

where MP denotes the total missing-stage penalty. 
The blended score is then adjusted using three
meta-evaluation checks applied to the complete
solution:

\begin{itemize}[leftmargin=4mm,itemsep=1pt]
\item \texttt{VERBOSITY} penalizes excessive
repetition, filler reasoning, and unnecessarily long
solutions that do not contribute meaningful
engineering reasoning~\citep{10.5555/3666122.3668142}.

\item \texttt{COVERAGE} checks whether all
requested quantities, sub-parts, assumptions, and
reasoning steps are addressed in the final solution.

\item \texttt{SANITY\_FAIL} detects physically
impossible or internally inconsistent outputs, such
as negative absolute temperature, invalid units, or
violations of conservation laws.
\end{itemize}

The resulting penalties are applied multiplicatively
during final score aggregation.
Final score is calculated as:
\begin{equation}
S_{\text{final}} = S_{\text{blend}} \cdot (1 - \mathtt{COVERAGE}) \cdot (1 - \mathtt{VERBOSITY}) \cdot 0.9^{\mathtt{SANITY\_FAIL}}
\end{equation}


\section{Experiments}
\label{sec:experiments}
\subsection{Benchmark Construction Pipeline}
\label{subsec:construction}
Question PDFs are processed using a semi-automated extraction pipeline combining Adobe PDF Extract API\footnote{\url{https://developer.adobe.com/document-services/apis/pdf-extract/}}, PyMuPDF\footnote{\url{https://pymupdf.readthedocs.io/en/latest/}}, and Gemini-3.1-Flash-Lite to recover layouts, figures, equations, and symbolic notation while preserving multimodal structure. All extracted questions, diagrams, metadata, and solutions are manually verified against the reference material.

\subsection{Experimental Settings}
We evaluate two vision-language models as solution generators: \texttt{Qwen3-VL-8B} and \texttt{Gemini-2.5-Flash}, representing compact open-weight and proprietary frontier models, respectively. We used Chain-of-Thought (CoT) prompting \cite{chain_of_thought2022} to trace step-by-step physical and mathematical reasoning (see Appendix \ref{app:prompts}). We use \texttt{Gemini-3.1-Pro-Preview} as the primary evaluator due to its strong multimodal reasoning capabilities, and additionally employ \texttt{Qwen3-VL-32B-Instruct} to assess cross-model agreement and reduce potential self-evaluation bias.

\paragraph{Baseline Evaluator Setup}
To quantify the benefits of EngJudge's structured design, we implement two single-pass baselines. In both baseline configurations, the evaluator receives the entire problem statement, the VLM's solution, and the ground-truth reference, and predicts scores for all eight reasoning stages in a single LLM call (see Appendix~\ref{app:baseline_evaluation_prompts}). The first baseline (\textit{SinglePass}) averages these raw stage scores ($Y_t$ in Equation~\ref{eq:raw_score}) directly. The second baseline (\textit{SinglePass + DP}) applies our Directed Acyclic Graph dependency propagation (DP) rules to obtain dependency scores ($S_t$ in Equation~\ref{eq:dep_score}) before averaging.

\subsection{Results}

\begin{figure}[H]
    \centering
    \includegraphics[width=1\linewidth]{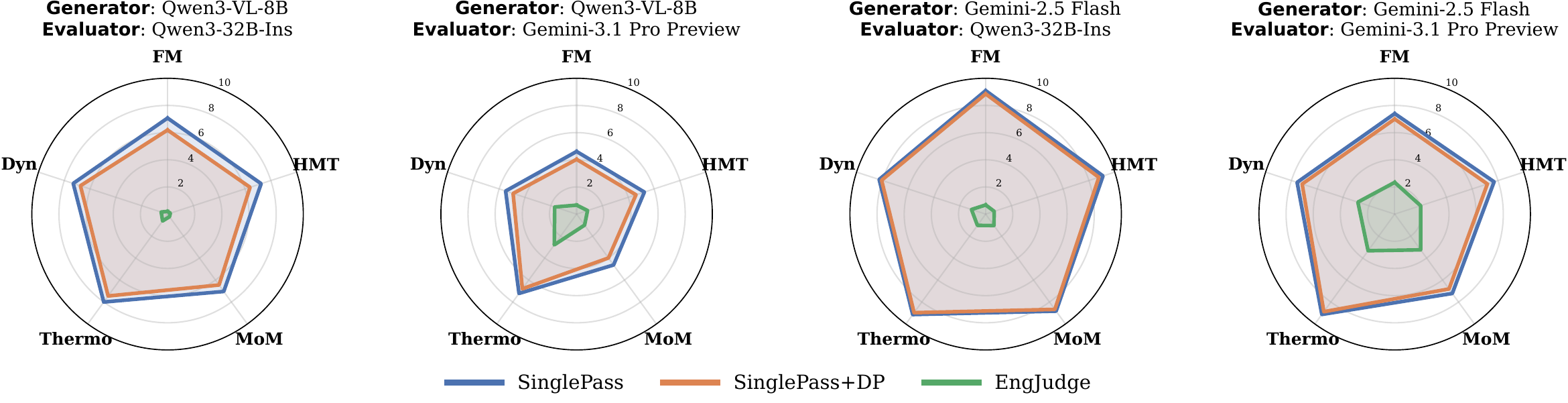}
\caption{
Cross-evaluator comparison across the five engineering subjects in EngVQA. Each radar plot represents a generator–evaluator pair. SinglePass uses a single LLM call and averages stage-wise raw scores $Y_t$ (Equation~\ref{eq:raw_score}), while SinglePass+DP averages stage-wise dependency scores $S_t$(Equation~\ref{eq:dep_score}). All scores range from 0--10.
}
    \label{fig:main_results_radar}
\end{figure}

\vspace{-0.4cm}

The cross-evaluator performance comparison across the five engineering subjects is summarized in Figure~\ref{fig:main_results_radar}. 
The quantitative results clearly indicate that \texttt{Gemini-2.5-Flash} significantly outperforms \texttt{Qwen3-VL-8B} across all evaluated subjects. 
This performance gap is consistent across all subjects.
Both models struggle severely in Fluid Mechanics and Heat \& Mass Transfer, which represent the lowest-scoring areas. 
This difficulty arises from the complex multi-phase physics, differential boundary conditions, and spatial geometry integration required in these fields. 
Conversely, performance is higher in Thermodynamics and Mechanics of Materials, where problems tend to rely more on algebraic path-following and discrete-state equations. 
Nonetheless, no generator model breaches a mean score of $4$ in any subject under EngJudge, highlighting the extreme difficulty of engineering questions for current state-of-the-art VLMs.

We also observe a massive performance discrepancy between the baseline and the EngJudge evaluation framework. 
When evaluated by the \texttt{Gemini-3.1-Pro-Preview (SinglePass)} one-shot judge, \texttt{Gemini-2.5-Flash} receives a highly optimistic overall score of $8.001$. 
However, when subjected to the structured, dependency-aware grading of \textbf{EngJudge}, its score drops to $2.869$. 
This behavior is driven by the fact that standard one-shot LLMs suffer from a leniency bias, often overlooking mathematical errors, missing justifications, or misinterpreting visuals if the final text appears superficially plausible.

\begin{wrapfigure}{r}{0.45\columnwidth}
\vspace{-10pt}
\centering

\begin{tikzpicture}
\begin{axis}[
    height=4cm,
    scale only axis,
    width=0.80\linewidth,
    ymin=0,
    ymax=10,
    ylabel={Score},
    xlabel={Stage},
    symbolic x coords={PC,AS,VI,ES,LR,AA,PI,FA},
    xtick=data,
    xticklabel style={rotate=45,anchor=east,font=\tiny},
    legend style={
        at={(0.5,-0.35)},
        anchor=north,
        legend columns=3,
        draw=none,
        font=\tiny
    },
    grid=major,
    mark size=1.5pt
]

\addplot[
    blue,
    thick,
    dashed,
    mark=o
]
coordinates{
(PC,9.82)
(AS,9.64)
(VI,8.49)
(ES,9.02)
(LR,7.90)
(AA,6.89)
(PI,7.93)
(FA,5.50)
};
\addplot[
    orange,
    thick,
    densely dotted,
    mark=triangle
]
coordinates{
(PC,9.82)
(AS,9.53)
(VI,8.27)
(ES,8.48)
(LR,7.47)
(AA,6.24)
(PI,7.32)
(FA,5.05)
};
\addplot[
    red,
    thick,
    mark=square*
]
coordinates{
(PC,7.33)
(AS,6.01)
(VI,5.81)
(ES,5.48)
(LR,4.71)
(AA,2.74)
(PI,4.53)
(FA,2.91)
};
\legend{
SinglePass,
SinglePass + DP,
EngJudge
}

\end{axis}
\end{tikzpicture}

\caption{Stage-wise scores.}
\label{fig:stepwise_scores_breakdown}

\vspace{-10pt}
\end{wrapfigure}
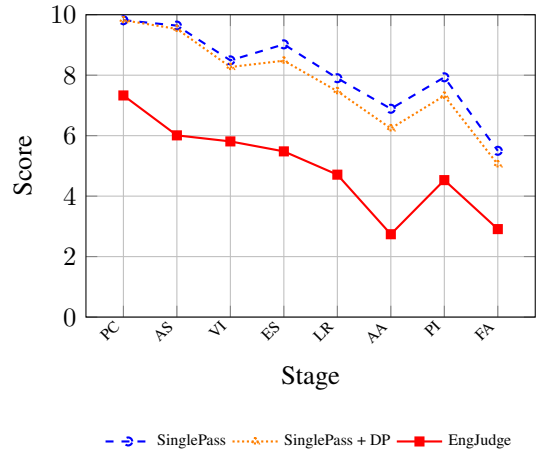

We observe that the gap between \textbf{SinglePass} and \textbf{SinglePass + DP} is remarkably small and relatively constant (ranging from $0.11$ to $0.65$ points). This indicates that mathematical error propagation alone is insufficient to address the overestimation of LLM capabilities. Instead, the massive drop to the \textbf{EngJudge} curve (a gap of $2.5$ to $4.1$ points across all stages) is primarily driven by the framework's granular penalty rubrics, fatal error caps, and global meta-checks. 




We also observe that the curves reveal a dichotomy between the \textit{problem setup} and \textit{execution}. During the initial setup stages, scores remain relatively high across all configurations (remaining above $5.4$ even under EngJudge). However, once the model transitions to the execution stages - Logical Reasoning (\texttt{LR}), Algebraic Accuracy (\texttt{AA}), Physical Interpretation (\texttt{PI}), and the Final Answer (\texttt{FA}) - performance drops significantly. Under EngJudge, the score bottoms out at $2.74$ for \texttt{AA}, indicating that algebraic calculation and sequential execution represent the \emph{primary cognitive bottlenecks} for the generator model.

\vspace{-0.3cm}
\begin{wraptable}{r}{0.5\columnwidth}
\centering
\scriptsize
\renewcommand{\arraystretch}{1.1}

\begin{tabular}{p{3.6cm}cc}
\toprule
\textbf{Generator Variant} &
\textbf{Qwen3-VL-8B} &
\textbf{Gemini 2.5 Flash} \\
\midrule
\textbf{EngJudge (Full Framework)} & \textbf{1.67} & \textbf{2.87} \\
EngJudge w/o DP & 2.48 & 4.12 \\
EngJudge w/o Meta Eval Checks & 2.35 & 4.75 \\
EngJudge w/o Missing Stage Penalty & 2.01 & 2.96 \\
\bottomrule
\end{tabular}

\caption{Component ablation of EngJudge using \texttt{Gemini 3.1 Pro Preview} as the evaluator.}
\label{tab:ablation}
\end{wraptable}

\paragraph{Component Ablation of EngJudge.}
Table~\ref{tab:ablation} quantifies the contribution of each framework component. Removing dependency propagation based on our causal graph (\emph{w/o DP}) substantially increases scores for both Gemini 2.5 Flash and Qwen3-VL-8B, showing that dependency-aware propagation is the primary source of grading rigor by penalizing downstream reasoning built on incorrect upstream formulations. Removing meta-evaluation checks further increases scores, indicating frequent violations of coverage and verbosity constraints in generated solutions. In contrast, removing missing-stage penalties has minimal impact on Gemini, suggesting strong adherence to the required structured reasoning format.



\begin{figure}[!htbp]
    \centering
    \includegraphics[width=0.5\linewidth]{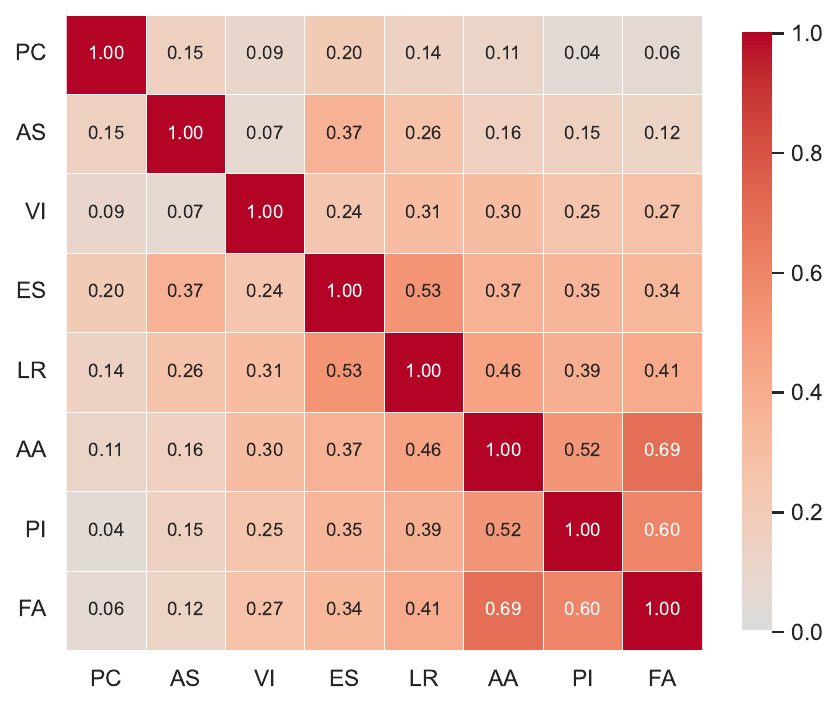}
    \caption{Pearson correlation matrix between stage-wise raw scores.}
    \label{fig:app_corr_heatmap}
\end{figure}

We also conduct an empirical analysis of stage-wise score correlations using the Pearson correlation matrix of raw scores across all stages (Figure~\ref{fig:app_corr_heatmap}). We observe strong correlations between adjacent reasoning stages, notably between Algebraic Accuracy (\texttt{AA}) and the Final Answer (\texttt{FA}) ($r=0.69$), and between Equation Selection (\texttt{ES}) and Logical Reasoning (\texttt{LR}) ($r=0.53$). 

Conversely, correlations between early conceptual stages and final calculations are negligible (e.g., Problem Characterization (\texttt{PC}) vs. Final Answer yields $r=0.06$). These strong, localized correlations reflect the physical reality of error propagation, empirically validating our dependency graph design: errors in prerequisite stages cascade to compromise downstream execution. Detailed analysis of these correlations and their relationship in the dependency graph is provided in Appendix~\ref{sec:appendix_correlation}.

\section{Human Study for Validating EngJudge}
\label{sec:human_study}
To rigorously validate the automated evaluations produced by the EngJudge (Sec \ref{sec:eng_judge}), we conducted a blinded human study to assess the alignment between EngJudge's structured scoring and human expert intuition. The evaluation panel consisted of 9 undergraduate/graduate students possessing strong backgrounds and practical experience in the chosen engineering subjects.

Evaluators were presented with randomized engineering problems, generated solutions from \texttt{Gemini 2.5 flash}, and EngJudge assigned scores for each stage through a web interface (see Appendix Figs. \ref{fig:human_study_ss_1_and_2}-\ref{fig:human_study_ss_12_and_13}). 
They rated EngJudge's stagewise and overall scores on 4 randomly selected questions.
EngJudge demonstrated exceptional alignment with human judgment, achieving a Pearson correlation coefficient of \textbf{0.975} and a Mean Absolute Error (MAE) of just 0.67 (on a 10-point scale). When assessing EngJudge's overall score, 66.7\% of the evaluators rated the overall system as ``Good'' or ``Very Good'' and 22.3\% rated it as average. 

In another setting, we asked human evaluators to compare EngJudge's dependency-aware score against a naive average, obtained by simply averaging the scores assigned to the eight stages in a blind A/B format. Evaluators explicitly favored the dependency score in 66.7\% of cases. This strongly validates our hypothesis that engineering reasoning evaluation requires hierarchical, cascading penalization rather than isolated stage-wise grading. A comprehensive breakdown of the methodology, graphical visualizations, and granular stage-level metrics is provided in Appendix \ref{app:human_study}.

\section{Conclusion}
\label{sec:conclusion}
In this paper, we introduced EngVQA, a challenging benchmark of graduate-level engineering physics problems spanning five core engineering disciplines. Second, we introduce EngJudge, an evaluation framework that uses dependency-aware score propagation to capture compounding physical and mathematical errors in engineering reasoning. Evaluation of SOTA VLMs on EngJudge highlights reasoning collapses due to visual-geometric and algebraic failures. EngJudge shows excellent alignment with human expert judgment, and its dependency-aware framework is preferred over averaging. We believe EngJudge can be a robust diagnostic tool to guide the development of physically grounded multimodal reasoning agents for engineering applications and interactive tutoring systems.


\section*{Limitations}
\label{sec:limitations}
\paragraph{Data Contamination.}
We cannot fully rule out benchmark contamination from model pretraining data. However, the benchmark emphasizes multimodal understanding and multi-step engineering reasoning, making direct memorization alone insufficient. 

\paragraph{Computational Cost of Evaluation.}
Our framework requires 11 LLM judge calls per solution, making evaluation significantly more computationally expensive. Although KV caching can reduce redundant computation \citep{pope2022efficientlyscalingtransformerinference}, evaluation cost may still limit scalability in large-scale or resource-constrained settings.

\paragraph{LLM Judge Reliability and Bias.}
EngJudge inherits known limitations of LLM-as-a-judge evaluation, including potential biases and imperfect alignment with human judgment. Although human validation demonstrates strong agreement, such biases cannot be fully eliminated \citep{huang-etal-2025-empirical}.

\section*{Ethics Statement}
This work complies with ethical guidelines concerning data sourcing, human studies, and computational deployment. The evaluation dataset comprises standard academic engineering physics problems.
We conducted the human study with volunteer graduate engineering students; participation was entirely voluntary, all data was fully anonymized, and no PII or demographic data was collected. Prior to the study, participants were presented with explicit instructions explaining how their evaluation data would be used, and they provided informed consent before proceeding. Furthermore, all evaluated models were accessed in accordance with their developer terms of service and license agreements, using local configurations on consumer-grade hardware to minimize computational energy consumption and environmental footprint.

\newpage
\label{sec:references}
\bibliographystyle{unsrtnat}
\bibliography{main}

\clearpage
\appendix
\label{sec:appendix}

\section{Error Analysis}
\label{appendix:error_analysis}

Before designing the evaluation framework, we conducted a systematic
pilot error analysis across three engineering subjects: Fluid
Mechanics, Heat and Mass Transfer (HMT), and Mechanics of Materials
(MOM). Rather than imposing evaluation steps heuristically, we first
analyzed how VLMs fail on engineering reasoning problems and whether
those failures exhibit consistent structural patterns across domains.

For each pilot subject, a frontier VLM (\texttt{Gemini-2.0-flash-exp}) was prompted to solve a subset
of the problems. A separate LLM judge then identified and
categorized all reasoning errors relative to the ground-truth
solutions, producing subject-specific error taxonomies grounded in
engineering domain knowledge. The errors and their categories were manually inspected as well. 

Unless otherwise noted, figures visualize representative pilot
subjects; analogous trends were consistently observed across all other subjects.

\paragraph{Key Finding 1: Errors Are Multi-Faceted and Co-occurring.}

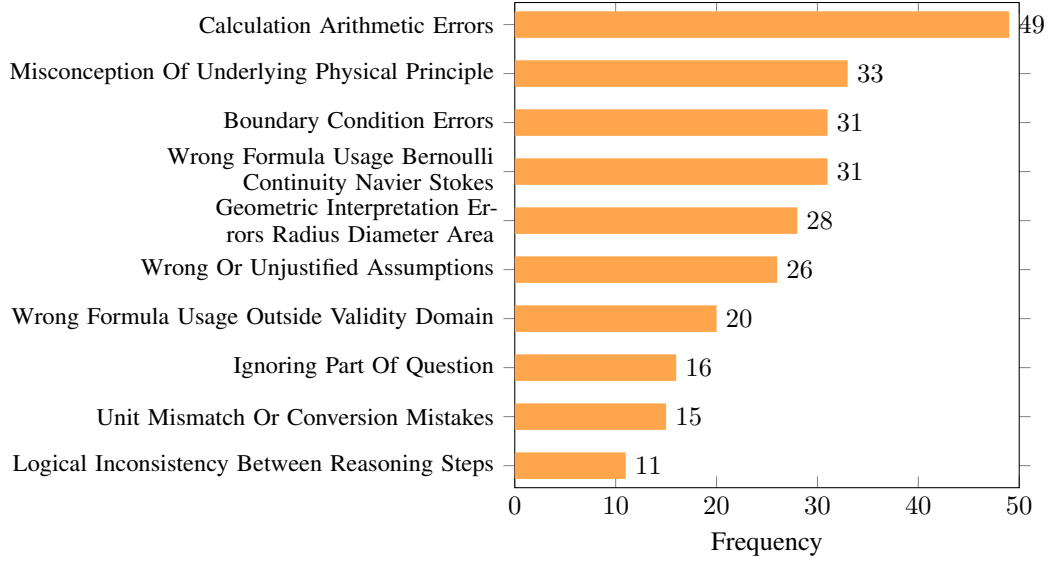
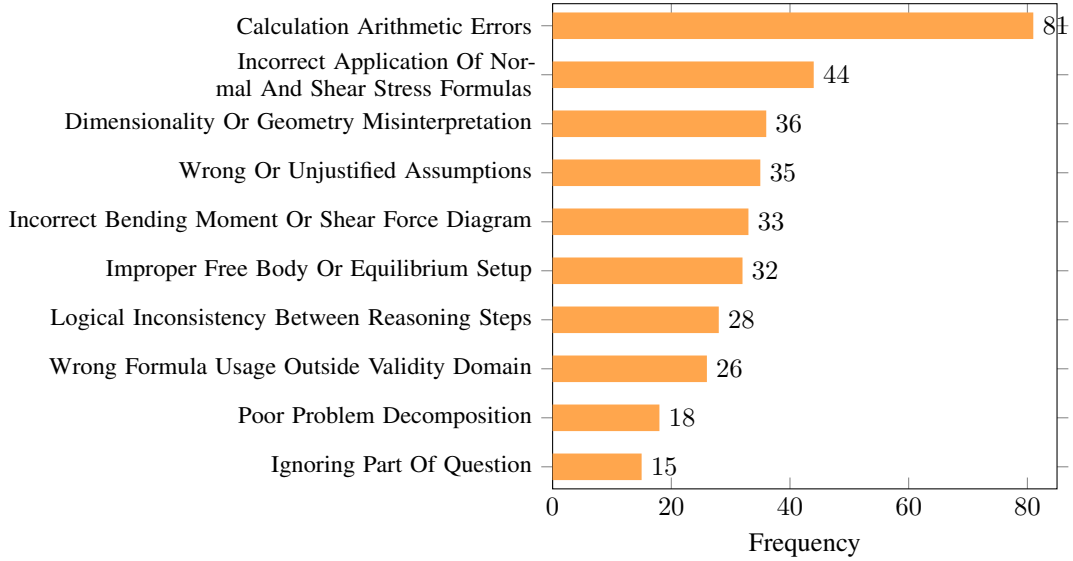
\begin{figure}[!t]
\centering
\begin{subfigure}{\textwidth}
\centering
\begin{tikzpicture}
\begin{axis}[
    xbar,
    width=0.50\textwidth,
    height=8cm,
    xmin=0,
    xmax=50,
    ytick=data,
    enlarge y limits=0.05,
    symbolic y coords={
        Logical Inconsistency Between Reasoning Steps,
        Unit Mismatch Or Conversion Mistakes,
        Ignoring Part Of Question,
        Wrong Formula Usage Outside Validity Domain,
        Wrong Or Unjustified Assumptions,
        Geometric Interpretation Errors Radius Diameter Area,
        Wrong Formula Usage Bernoulli Continuity Navier Stokes,
        Boundary Condition Errors,
        Misconception Of Underlying Physical Principle,
        Calculation Arithmetic Errors
    },
    yticklabel style={
        text width=7cm,
        align=right,
        font=\footnotesize
    },
    nodes near coords,
    nodes near coords align={horizontal},
    xlabel={Frequency}
]

\addplot[
    fill=orange!70,
    draw=none
]
coordinates {
(11,Logical Inconsistency Between Reasoning Steps)
(15,Unit Mismatch Or Conversion Mistakes)
(16,Ignoring Part Of Question)
(20,Wrong Formula Usage Outside Validity Domain)
(26,Wrong Or Unjustified Assumptions)
(28,Geometric Interpretation Errors Radius Diameter Area)
(31,Wrong Formula Usage Bernoulli Continuity Navier Stokes)
(31,Boundary Condition Errors)
(33,Misconception Of Underlying Physical Principle)
(49,Calculation Arithmetic Errors)
};

\end{axis}
\end{tikzpicture}
\caption{Fluid Mechanics}
\end{subfigure}
\vspace{1em}

\begin{subfigure}{\textwidth}
\centering
\begin{tikzpicture}
\begin{axis}[
    xbar,
    width=0.50\textwidth,
    height=8cm,
    xmin=0,
    xmax=85,
    ytick=data,
    enlarge y limits=0.05,
    symbolic y coords={
        Ignoring Part Of Question,
        Poor Problem Decomposition,
        Wrong Formula Usage Outside Validity Domain,
        Logical Inconsistency Between Reasoning Steps,
        Improper Free Body Or Equilibrium Setup,
        Incorrect Bending Moment Or Shear Force Diagram,
        Wrong Or Unjustified Assumptions,
        Dimensionality Or Geometry Misinterpretation,
        Incorrect Application Of Normal And Shear Stress Formulas,
        Calculation Arithmetic Errors
    },
    yticklabel style={
        text width=7cm,
        align=right,
        font=\footnotesize
    },
    nodes near coords,
    nodes near coords align={horizontal},
    xlabel={Frequency}
]
\addplot[
    fill=orange!70,
    draw=none
]
coordinates {
(15,Ignoring Part Of Question)
(18,Poor Problem Decomposition)
(26,Wrong Formula Usage Outside Validity Domain)
(28,Logical Inconsistency Between Reasoning Steps)
(32,Improper Free Body Or Equilibrium Setup)
(33,Incorrect Bending Moment Or Shear Force Diagram)
(35,Wrong Or Unjustified Assumptions)
(36,Dimensionality Or Geometry Misinterpretation)
(44,Incorrect Application Of Normal And Shear Stress Formulas)
(81,Calculation Arithmetic Errors)
};

\end{axis}
\end{tikzpicture}
\caption{Mechanics of Materials}
\end{subfigure}
\caption{
Most frequent error categories across representative engineering domains. Arithmetic-related failures are consistently common, while secondary error distributions remain domain-specific. Heat and Mass Transfer exhibited similar trends.
}
\label{fig:top_errors}
\end{figure}

Across all three pilot subjects, arithmetic-related failures were
among the most frequent observed error categories, occurring 49 times
in Fluid Mechanics, 45 times in HMT, and 80 times in MOM
(Figure~\ref{fig:top_errors}). However, the secondary distribution of
errors varied substantially across domains.

Fluid Mechanics failures frequently involved wrong formula usage (Bernoulli, continuity, Navier-Stokes: 31 occurrences) and boundary
condition errors (31 occurrences). HMT failures showed a high rate of wrong or unjustified assumptions (31 occurrences), particularly misidentificaion of steady-state vs. transient regime. MOM failures were dominated by stress-formula misuse(44 occurrences),
free-body setup errors, and geometric misinterpretation(36 occurrences).

These results indicate that engineering reasoning failures are not
well-described by a single generic failure mode. Instead, different
engineering domains exhibit distinct reasoning error signatures,
motivating domain-aware and step-level evaluation.

\paragraph{Key Finding 2: Errors Cluster into Distinct Reasoning
Steps.}

When we grouped the subject-specific error types by the reasoning
operation they implicate, a consistent structure emerged across all
three subjects shown in (Table~\ref{tab:error_step_mapping}).

\begin{table}[H]
\centering
\scriptsize
\begin{tabular}{@{}p{3.1cm}p{10.9cm}@{}}
\toprule
\textbf{Stage} & \textbf{Representative Error Types} \\
\midrule
Problem Characterization.
  & Ignoring part of question; poor decomposition \\[2pt]
Assumptions
  & Wrong/unjustified assumptions; steady vs.\ transient
    failure; lumped capacitance misuse \\[2pt]
Visual Interpretation
  & Geometric misinterpretation; incorrect FBD or shear
    force diagram; boundary condition misread from figure \\[2pt]
Equation Selection
  & Fourier/Fick misapplication; wrong Bernoulli form;
    formula outside validity domain; wrong stress relation \\[2pt]
Logical Reasoning
  & Inconsistency between steps; incorrect sub-result
    dependency \\[2pt]
Algebraic Accuracy
  & Arithmetic errors; unit mismatch; incorrect
    dimensionless parameter definition \\[2pt]
Physical Interpretation
  & Physical principle misconception; material property
    misuse; incorrect property evaluation temperature \\[2pt]
Final Answer
  & Numerical error vs.\ ground truth; physically
    impossible result; partial answer \\
\bottomrule
\end{tabular}
\vspace{0.3cm}
\caption{Mapping of observed error types to evaluation steps. Each
  step captures a qualitatively distinct failure mode not detectable
  by inspecting other steps.}
\label{tab:error_step_mapping}
\end{table}

\paragraph{Key Finding 3: Engineering Errors Exhibit Sparse Structured Dependencies.}

\begin{figure}[t]
  \centering
  \includegraphics[width=0.7\textwidth]{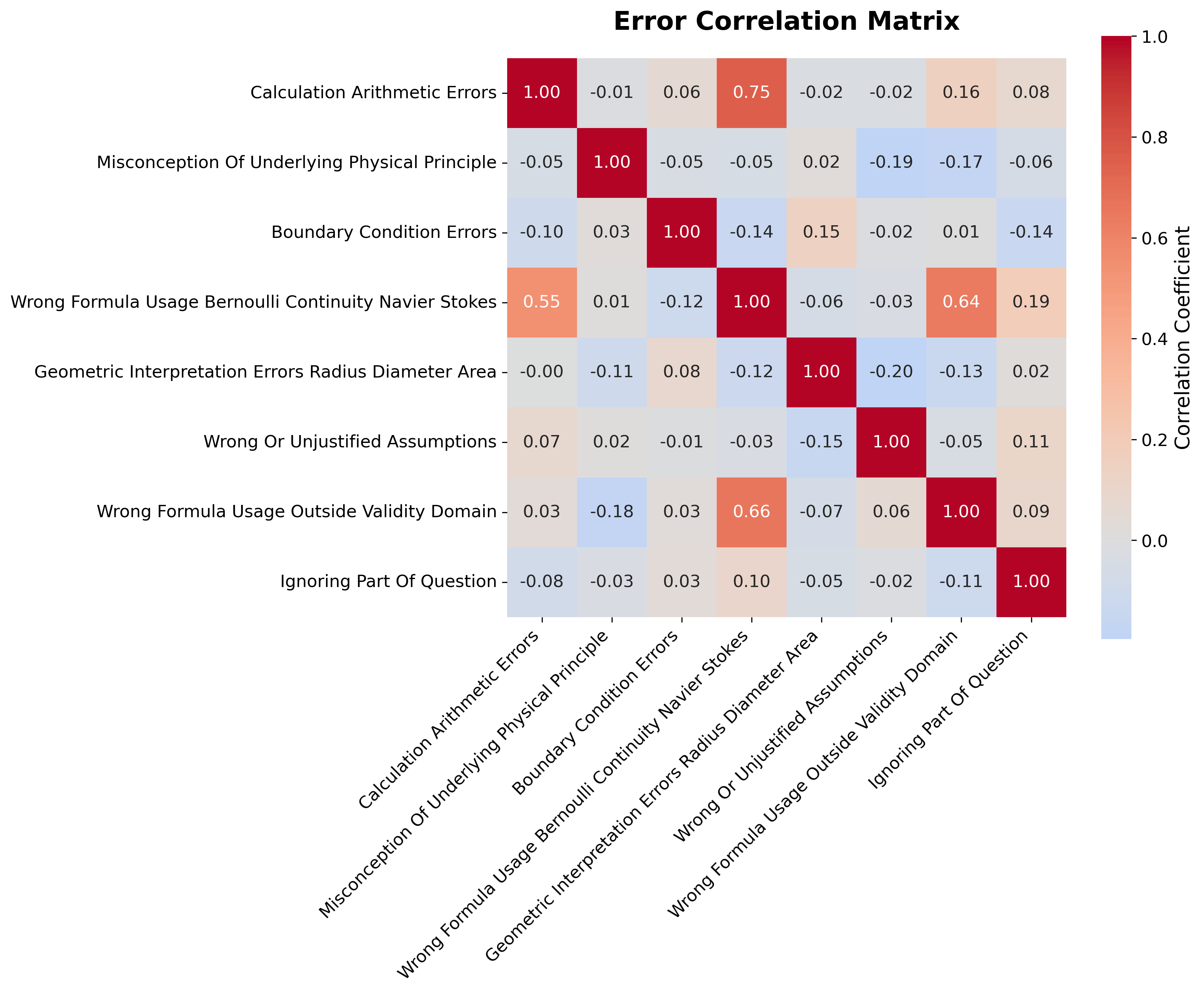}
  \caption{Error correlation matrix for Fluid Mechanics. Most
    off-diagonal entries are near zero, confirming step
    independence. The exception, formula misapplication and
    arithmetic errors ($r = 0.75$), reflects a cascading
    dependency that directly motivates the propagation edge in the
    DAG.}
  \label{fig:error_correlation}
\end{figure}

Figure~\ref{fig:error_correlation} shows error
correlation matrices from the pilot analysis. The matrices reveal
that engineering reasoning failures are not fully independent. Several
dependencies consistently emerge across errors in different reasoning stages.

The strongest and most consistent dependency appears between equation
misuse and arithmetic errors ($r = 0.55$--$0.75$ across pilot
subjects), reflecting cascading failures where incorrect governing
equations propagate into downstream computation. Additional moderate
dependencies were observed between visual interpretation and equation
selection in diagram-centric problems, as well as between assumption
formulation and downstream logical reasoning.

This sparse dependency structure has a direct methodological
implication: engineering reasoning should not be evaluated using a
single score, it also cannot be fully
factorized into completely independent steps. Instead, evaluation
must combine step-level assessment with selective dependency-aware
propagation along empirically justified reasoning edges.

\paragraph{Key Finding 4: Visual Reasoning Is a Distinct and
Persistent Failure Mode.}

\begin{figure}[t]
\centering
\begin{tikzpicture}
\begin{axis}[
    xbar,
    bar width=10pt,
    width=0.98\textwidth,
    height=6.5cm,
    xmin=0, xmax=0.68,
    xtick={0.0,0.1,0.2,0.3,0.4,0.5},
    xticklabel style={font=\scriptsize},
    ytick={0,1,2,3,4,5,6,7,8,9},
    yticklabels={
        {Hydrostatic forces},
        {Conservation of mass},
        {Energy eq.\ (First Law)},
        {Stability},
        {Bernoulli's eq.},
        {Archimedes' principle},
        {Viscosity \& class.},
        {Buoyancy},
        {Hydrostatic pressure dist.},
        {Hydrostatic pressure dist.}
    },
    yticklabel style={
        font=\scriptsize,
        align=right,
        text width=2.6cm,
    },
    enlarge y limits=0.06,
    nodes near coords,
    nodes near coords align=horizontal,
    every node near coord/.append style={
        font=\scriptsize\bfseries,
        anchor=west,
        xshift=3pt,
    },
    axis x line=bottom,
    axis y line=left,
    xlabel={\scriptsize Visual Error Susceptibility Score},
    xlabel style={yshift=-3pt},
    clip=false,
    tick align=outside,
]
\addplot[draw=none, fill=red!55!black!70] coordinates {
    (0.40, 0)
    (0.43, 1)
    (0.45, 2)
    (0.45, 3)
    (0.45, 4)
    (0.47, 5)
    (0.49, 6)
    (0.49, 7)
    (0.56, 8)
    (0.56, 9)
};
\end{axis}
\end{tikzpicture}
\caption{Topics with highest visual error susceptibility scores
  (Fluid Mechanics). Scores above $0.5$ indicate that the majority
  of errors in that topic originate from diagram misinterpretation,
  confirming visual reasoning as a structurally distinct failure mode.}
\label{fig:visual_susceptibility}
\end{figure}
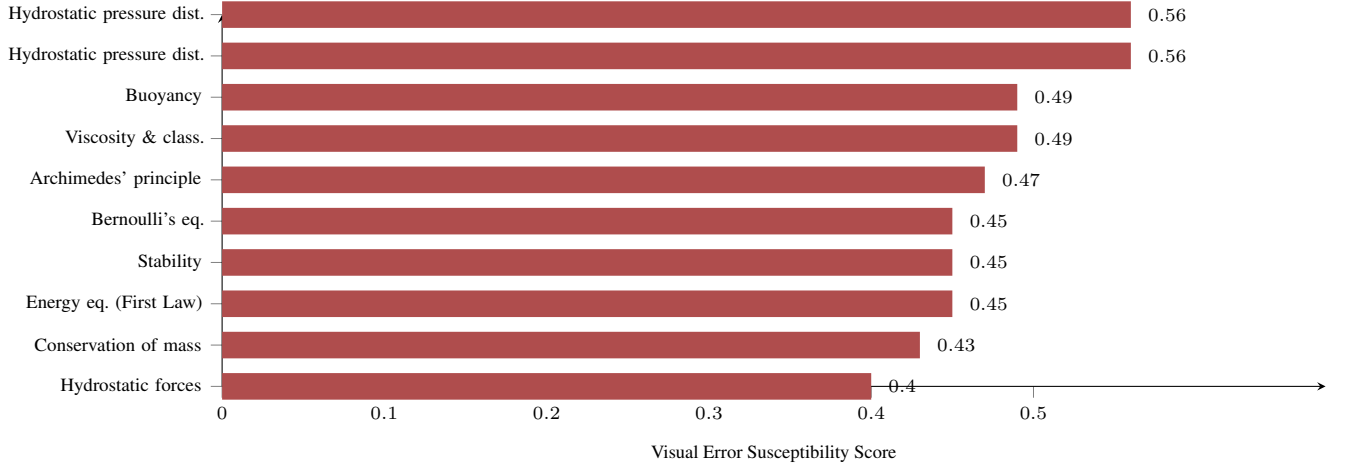

We computed a \textit{visual error susceptibility score} for each
topic within each subject---defined as the fraction of errors due to misinterpretation of the diagram rather than to
purely textual or computational reasoning
(Figure~\ref{fig:visual_susceptibility}). Several topics scored above
0.55: \textit{hydrostatic pressure distributions} and
\textit{buoyancy} in Fluid Mechanics (0.56 each); \textit{thermal
resistance networks} in HMT (0.57); and \textit{stress resultants
and equilibrium}, \textit{torsion}, and \textit{bending} in MOM
(0.59, 0.55, 0.55). This led us to concluding that visual misinterpretation is a structurally
independent failure mode requiring its own evaluation step.

\paragraph{From Error Analysis to Evaluation Design.}

The pilot error analysis directly motivated the design of our
evaluation framework. Frequent propagation from equation mistakes to
arithmetic errors motivated the use of \textit{dependency
propagation}, where upstream mistakes reduce the score of downstream
steps. Similarly, the large number of diagram-related failures
motivated a dedicated \textit{Visual Interpretation} step, while the
different reasoning patterns across subjects motivated the use of
\textit{domain-specific prompts}.

\begin{wrapfigure}{r}{0.48\columnwidth}
\vspace{-10pt}
\centering
\includegraphics[width=0.46\linewidth]{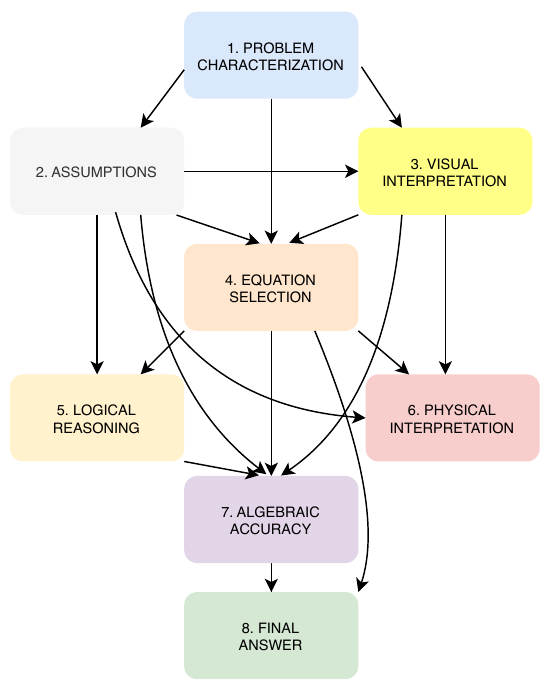}
\caption{Dependency DAG structure used for trust propagation across reasoning steps.}
\label{fig:dependency_graph}
\vspace{-10pt}
\end{wrapfigure}

The resulting DAG structure
(Figure~\ref{fig:dependency_graph}) follows the natural flow of an
engineering solution. \textit{Problem Characterization} and
\textit{Assumptions} appear first because they affect all later
steps. \textit{Visual Interpretation} feeds into
\textit{Equation Selection}, since correct equations often depend on
properly reading diagrams and boundary conditions.
\textit{Equation Selection} and \textit{Logical Reasoning} influence
\textit{Algebraic Accuracy}, because calculations based on incorrect
equations or inconsistent logic usually produce incorrect numerical
results. \textit{Physical Interpretation} runs alongside symbolic
computation and contributes to the \textit{Final Answer}, ensuring
that the final result is also physically meaningful.

The error correlation matrix (Figure~\ref{fig:error_correlation})
provides more insight into this propagation of errors.
This cascade pattern is precisely what the dependency propagation
formula captures: a correct downstream step built on a flawed
upstream step should not receive full credit, because the apparent
correctness is contingent on an invalid foundation.




\section{Benchmark details}
\label{app:benchmark_details}
To ensure broad curriculum-level coverage and reduce
over-specialization toward narrow reasoning patterns,
EngVQA was curated across five major engineering
domains spanning dynamics, thermodynamics,
transport phenomena, structural mechanics, and
physics-constrained analytical reasoning. The
benchmark intentionally includes both foundational
engineering principles and advanced multi-topic
analytical problems commonly encountered in
undergraduate and graduate-level engineering
curricula.

Many questions require simultaneous reasoning
across multiple conceptual categories, including
diagram interpretation, governing equation
selection, physical constraint validation, symbolic
derivation, and numerical computation. Rather
than evaluating isolated single-step calculations,
the benchmark emphasizes interconnected
engineering workflows involving coupled physical
processes, system-level reasoning, and multi-stage
analytical dependencies.

The benchmark additionally contains substantial
cross-topic overlap within each subject area. For
example, a single problem may simultaneously
involve visual interpretation, conservation laws,
material properties, boundary conditions, and
multi-stage quantitative reasoning. This structure
was intentionally designed to evaluate robustness
across heterogeneous engineering concepts,
mathematical formulations, and physical settings.

The full list of representative topics covered within
each engineering domain is summarized below.

\vspace{3pt}

\begin{tcolorbox}[
colback=blue!2,
colframe=blue!35,
title=\textbf{Dynamics Topic Coverage},
fonttitle=\scriptsize\bfseries,
boxrule=0.25pt,
arc=1pt,
left=1pt,right=1pt,
top=0.5pt,bottom=0.5pt
]
\tiny

\textbf{Particle Kinematics:}
Rectilinear motion with constant and variable
acceleration; dependent motion of particles;
projectile motion; normal--tangential coordinates;
polar and cylindrical coordinates; relative motion;
absolute and relative velocity/acceleration;
curvilinear trajectories; coordinate transformations;
rotating reference frames.

\textbf{Particle Kinetics:}
Newton's second law; equations of motion in
Cartesian, normal--tangential, and cylindrical
coordinates; work--energy methods; conservative
and non-conservative forces; impulse--momentum;
angular impulse and momentum; impact analysis;
central-force motion; power and efficiency.

\textbf{Rigid-Body Dynamics:}
Plane motion; fixed-axis rotation; rolling constraints;
instantaneous centers; rigid-body kinetics;
mass moment of inertia; radius of gyration;
gyroscopic motion; free and damped vibration;
force and moment analysis; rigid-body impact;
connected rigid-body systems; cable tension;
planetary and orbital motion.
\end{tcolorbox}

\vspace{2pt}

\begin{tcolorbox}[
colback=red!2,
colframe=red!35,
title=\textbf{Thermodynamics Topic Coverage},
fonttitle=\scriptsize\bfseries,
boxrule=0.25pt,
arc=1pt,
left=1pt,right=1pt,
top=0.5pt,bottom=0.5pt
]
\tiny

\textbf{Fundamental Thermodynamics:}
Open and closed systems; intensive and extensive
properties; pressure and temperature;
thermodynamic equilibrium; specific volume and
density; work and heat transfer mechanisms;
energy conservation; mechanical and thermal
equilibrium; property relations; phase-change
processes; saturated mixtures; superheated vapor;
subcooled liquid; property diagrams and tables.

\textbf{Energy and First-Law Analysis:}
Closed-system and control-volume energy balances;
steady and unsteady flow; nozzles and diffusers;
turbines and compressors; throttling valves;
pumps and heat exchangers; polytropic processes;
specific heats; ideal-gas and real-gas relations;
compressibility effects; flow work and enthalpy.

\textbf{Entropy and Exergy:}
Second-law analysis; entropy generation;
isentropic processes; reversible and irreversible
processes; exergy destruction; thermal efficiency;
COP of refrigeration and heat-pump systems;
availability and work potential; entropy balances.

\textbf{Thermodynamic Cycles:}
Carnot cycle; Rankine cycle with regeneration
and reheat; Brayton cycle and gas turbines;
Otto and Diesel cycles; refrigeration cycles;
combined cycles; jet propulsion systems.
\end{tcolorbox}

\vspace{2pt}

\begin{tcolorbox}[
colback=teal!2,
colframe=teal!35,
title=\textbf{Fluid Mechanics Topic Coverage},
fonttitle=\scriptsize\bfseries,
boxrule=0.25pt,
arc=1pt,
left=1pt,right=1pt,
top=0.5pt,bottom=0.5pt
]
\tiny

\textbf{Fluid Statics and Properties:}
Hydrostatic pressure distributions; manometry;
fluid properties; shear stress and viscosity;
surface tension and capillarity; buoyancy and
stability; atmospheric pressure variation;
hydrostatic forces on submerged surfaces.

\textbf{Fluid Dynamics and Conservation Laws:}
Conservation of mass and momentum;
Bernoulli analysis; integral and differential
control-volume analysis; continuity equations;
energy equations; velocity and acceleration fields;
vorticity and circulation; dimensional analysis;
Buckingham-$\Pi$ theorem; Reynolds, Mach,
Froude, Weber, and Euler numbers.

\textbf{Internal and External Flows:}
Laminar and turbulent pipe flow; entrance-region
development; boundary layers; Reynolds-number
effects; head losses and Moody diagram;
velocity profiles; hydraulic jumps; flow separation;
drag and lift; wake formation; control surfaces;
pump and turbine performance; cavitation.

\textbf{Compressible and Computational Flows:}
Compressible nozzle flows; isentropic flow;
Fanno and Rayleigh flow; shock waves;
potential flow theory; Navier--Stokes equations;
Euler equations; CFD discretization and
numerical convergence; flow visualization;
transport phenomena and environmental flows.
\end{tcolorbox}

\vspace{2pt}

\begin{tcolorbox}[
colback=orange!2,
colframe=orange!40,
title=\textbf{Heat \& Mass Transfer Topic Coverage},
fonttitle=\scriptsize\bfseries,
boxrule=0.25pt,
arc=1pt,
left=1pt,right=1pt,
top=0.5pt,bottom=0.5pt
]
\tiny

\textbf{Heat Conduction:}
Fourier's law; steady-state and transient
conduction; thermal conductivity;
thermal resistance networks; lumped-capacitance
analysis; plane-wall, cylindrical, and spherical
conduction; composite walls; fins and extended
surfaces; thermal contact resistance;
two-dimensional conduction; conduction
shape factors; boundary conditions in conduction.

\textbf{Convection and Boundary Layers:}
Forced and natural convection;
convection heat-transfer coefficients;
velocity and thermal boundary layers;
laminar and turbulent flow;
Reynolds, Prandtl, Nusselt, Grashof,
Rayleigh, and Biot numbers;
internal and external flow convection;
mixed convection; boiling and condensation;
hydrodynamic and thermal entry lengths.

\textbf{Thermal Radiation:}
Blackbody radiation; emissivity;
Stefan--Boltzmann law; Planck's law;
Wien's displacement law; radiation exchange;
view factors; participating media radiation;
surface energy balance.

\textbf{Mass Transfer and Coupled Transport:}
Fick's law; mass diffusivity;
species conservation; evaporation;
Raoult's and Henry's laws;
heat exchangers; LMTD and NTU methods;
thermal energy generation and storage;
thermoelectric effects; coupled heat and
mass-transfer analogies.
\end{tcolorbox}

\vspace{2pt}

\begin{tcolorbox}[
colback=violet!2,
colframe=violet!35,
title=\textbf{Mechanics of Materials Topic Coverage},
fonttitle=\scriptsize\bfseries,
boxrule=0.25pt,
arc=1pt,
left=1pt,right=1pt,
top=0.5pt,bottom=0.5pt
]
\tiny

\textbf{Stress and Deformation:}
Normal and shear stress; axial loading;
stress--strain relations; Hooke's law;
Poisson's ratio; thermal stresses;
combined loading; pressure vessels;
principal stresses; Mohr's circle;
strain transformations; failure theories.

\textbf{Structural Analysis:}
Beam bending; torsion of circular and
thin-walled sections; free-body diagrams;
equilibrium analysis; stress resultants;
shear and bending moment diagrams;
neutral axis; centroids and moments of inertia;
section modulus; shear center;
buckling and elastic stability.

\textbf{Advanced Material Behavior:}
Stress concentration effects;
composite beams and transformed sections;
unsymmetric bending;
energy methods and Castigliano theorem;
deflection analysis;
impact loading and strain energy;
plastic bending;
thin-walled structures;
material yielding and fracture behavior.
\end{tcolorbox}

\subsection{Topic Generation and Assignment Methodology}
\label{appendix:topic_methodology}

To systematically classify the conceptual coverage of the benchmark, we established a structured vocabulary of topics for each engineering subject and mapped them to individual problems. This appendix details our topic extraction process, validation matching pipeline, and the resulting conceptual density statistics.

\subsubsection{Master Topic List Synthesis}
Rather than relying on manually predefined keywords, we utilized an LLM-driven topic synthesis pipeline to generate a realistic representation of university-level engineering curricula. For each subject, the text statements of all questions in the dataset were concatenated and fed into \texttt{Gemini 2.5 Flash}. The model was prompted to analyze the question corpus and synthesize a master vocabulary of $50$--$60$ generalized topics per domain (e.g., $53$ for Thermodynamics and $54$ for Dynamics). These synthesized topics were structured to match standard course syllabi.

\subsubsection{Multimodal Topic Assignment and Validation}
With the master topic list established, we ran a multimodal classification pipeline to map topics to each problem. For each question:
\begin{enumerate}
    \item The model was provided with the question statement, its master topic list, and the corresponding question diagram.
    \item The model selected all topics necessary to formulate or solve the problem.
    \item To ensure data integrity and prevent hallucinated labels, the model's output was processed by an automated validation script. The validator matched each output string against the master topic list using case-insensitive transformations and fuzzy string matching (with a similarity cutoff threshold of $0.85$). Topics failing this validation check were discarded.
\end{enumerate}

\subsubsection{Conceptual Density and Subject Variations}

The average number of topics assigned per question varies significantly by subject:
\begin{itemize}
    \item \textbf{Dynamics ($3.57$ topics/Q):} Emphasizes modular problem-solving frameworks where questions are typically isolated to specific kinematic or kinetic formulations (e.g., pure projectile motion or pure impulse-momentum analysis).
    \item \textbf{Fluid Mechanics ($5.81$ topics/Q):} Highlights moderately coupled phenomena such as pressure fields under hydrostatic conditions or head loss calculations in pipe networks.
    \item \textbf{Thermodynamics ($9.28$ topics/Q) and Mechanics of Materials ($9.46$ topics/Q):} Require concurrent evaluation of material properties, load conditions, geometric invariants, and thermodynamic state variables.
    \item \textbf{Heat \& Mass Transfer ($13.89$ topics/Q):} Exhibits the highest conceptual coupling because standard heat transfer scenarios inherently involve multiple parallel transport phenomena (conduction, convection, and radiation occurring simultaneously with thermal boundaries and resistance properties).
\end{itemize}
This multi-concept mapping highlights that solving a typical problem in our benchmark requires the simultaneous integration and synthesis of multiple physical principles, making the benchmark fundamentally more challenging than standard single-template datasets.

\section{Detailed Evaluation Framework}
\label{appendix:evaluation_framework_details}

This appendix provides the detailed mathematical and algorithmic specifications for the \textbf{EngJudge} evaluation framework, expanding on the summary in Section~\ref{sec:evaluation}.

\subsection{Penalty Scales}
\label{app:penalty_scales}

The automated evaluation framework utilizes a penalty-based grading scheme designed to mirror the grading practices of human professors in graduate-level engineering courses. Rather than evaluating a solution in a binary (correct/incorrect) fashion, each of the eight mandatory reasoning stages begins with a base score of 10. The automated grader identifies specific errors within the stage, classifies them into one of four severity levels, and subtracts the corresponding penalty points.

The four severity levels consist of : \textbf{Minor} (2 points), \textbf{Moderate} (4 points), \textbf{Major} (7 points), and \textbf{Critical} (10 points). The spacing of this scale (2, 4, 7, 10) is designed to ensure that minor arithmetic or formatting slips do not excessively penalize a student's score, while severe conceptual errors or invalid physical formulations are heavily penalized, often resulting in an immediate zero for that stage. This configuration provides a flexible yet robust mechanism for partial credit distribution.

Table~\ref{tab:penalty_scales} summarizes the criteria and provides concrete illustrative examples from the evaluation of the benchmark questions.

\begin{table}[htbp]
\centering
\small
\renewcommand{\arraystretch}{1.3}
\begin{tabular}{lcp{4.5cm}p{5.5cm}}
\hline
\textbf{Penalty Level} & \textbf{Points} & \textbf{Grading Criteria \& Description} & \textbf{Illustrative Benchmark Example} \\ \hline
Minor & 2 & 
Insignificant arithmetic or rounding slips, notation issues, or minor dimension misreads with negligible downstream effects. & 
In a forced convection chip cooling problem (Heat \& Mass Transfer, Problem 71), using a slightly rounded air velocity ($30.8$ m/s converted to $31.29$ m/s) in the Reynolds number calculation, which does not alter the analytical method. \\ \hline
Moderate & 4 & 
Unit conversion errors, sub-optimal coordinate system selection, or missing an important visual boundary condition. & 
In a piston-cylinder compression problem (Thermodynamics, Problem 5-131), failing to explicitly state the ``large surroundings'' assumption needed to simplify the radiation heat transfer terms. \\ \hline
Major & 7 & 
Formulating boundary conditions incorrectly, using equations outside their valid limits, or severe derivation slips. & 
In a reheat-regenerative Rankine cycle analysis (Thermodynamics, Problem 10-107), misidentifying state points and extraction pressures for a closed feedwater heater, leading to incorrect enthalpy lookups. \\ \hline
Critical & 10 & 
Chosing fundamentally wrong governing equations, dimensionally inconsistent boundary conditions, or physically impossible results. & 
In a pipe exit plug fluid momentum analysis (Fluid Mechanics, Problem 11), applying Bernoulli's equation to calculate a non-zero static pressure at the exit of an unconfined jet discharging to atmosphere. \\ \hline
\end{tabular}
\vspace{0.3cm}
\caption{Detailed penalty scales, grading criteria, and concrete failure examples across the benchmark domains.}
\label{tab:penalty_scales}
\end{table}

At the end of the human study, evaluators were asked to rate the calibration of the penalty scheme (the specific deductions of 2, 4, 7, and 10 points for minor, moderate, major, and critical errors). The feedback from the 9 expert participants was distributed as follows:
\begin{itemize}
    \item \textbf{Well Calibrated:} \textbf{55.6\%} (5 out of 9 evaluators) expressed direct agreement that the penalty levels are well calibrated.
    \item \textbf{Minors Too Harsh:} \textbf{22.2\%} (2 out of 9 evaluators) felt that minor error penalties (2 points) were slightly too harsh.
    \item \textbf{Scale Too Coarse:} \textbf{11.1\%} (1 out of 9 evaluators) suggested a more granular scale.
    \item \textbf{Not Harsh Enough:} \textbf{11.1\%} (1 out of 9 evaluators) indicated that the penalties could be even stricter.
\end{itemize}


\subsection{Fatal Error Capping Mechanism}
\label{app:fatal_error_caps}

To prevent VLMs from receiving high scores on steps containing fundamental conceptual breakdowns, the EngJudge framework incorporates a \emph{Fatal Error Capping} mechanism. During the evaluation of a reasoning step, if the evaluator model detects any of the fatal error types defined in Table~\ref{tab:fatal_caps}, the score for that specific step is capped at the designated value, regardless of any correct algebraic manipulations or formatting compliance. If multiple fatal errors are triggered within a single step, the step score is capped at the \emph{lowest} applicable value.

The physical rationales and definition for each fatal error cap are detailed below:

\begin{itemize}
    \item \textbf{Governing Equation Incorrect (Score Cap = 3):} This error is triggered when the VLM selects a fundamental governing equation that is physically inapplicable to the problem domain. A prime example is applying Bernoulli's equation (which assumes frictionless, inviscid flow) to a high-viscosity pipe flow system where shear stress dominates, or using linear momentum equations in a system where rotational torque governs. Because selecting the incorrect governing relation invalidates the entire physical foundation of the solution, the step is capped at a low score of $3$, reflecting a failure in basic physical reasoning.
    \item \textbf{Physical Model Invalid (Score Cap = 4):} This error occurs when the VLM formulates a physical representation or simplification that directly contradicts the physical realities of the problem. Examples include assuming a transient process is at steady-state, treating a highly compressible gas as an incompressible fluid, or modeling a multi-dimensional flux as a one-dimensional flow without justification. While the model may choose the correct general equations, the invalid simplification renders the subsequent modeling physically incorrect, justifying a cap of $4$.
    \item \textbf{Dimensionally Inconsistent (Score Cap = 2):} This represents the most severe algebraic and physical violation in engineering. It is triggered when the model writes an equation where the units or physical dimensions of the left-hand side do not match those of the right-hand side (e.g., adding a force term directly to a velocity term, or equating mass flow rate to pressure). Because dimensional consistency is the most fundamental sanity check in physical sciences, any dimensional violation indicates a complete collapse of mathematical and physical consistency, resulting in the strictest cap of $2$.
    \item \textbf{Boundary Conditions Invalid (Score Cap = 4):} In graduate-level engineering physics, solving governing differential equations (such as the Navier-Stokes or heat diffusion equations) depends entirely on specifying boundary and initial conditions. This error is triggered when the model misapplies interface conditions—such as setting a free-slip condition at a stationary wall, neglecting convective heat transfer at a boundary, or using absolute pressure in momentum balances where gauge pressure is required. Since incorrect boundary conditions yield physically impossible solutions, the score is capped at $4$.
\end{itemize}
By enforcing these limits, the framework ensures that a solution is not graded solely on surface-level correctness or formatting compliance, but rather on its conceptual and physical integrity.

\begin{table}[h]
\centering
\scriptsize
\renewcommand{\arraystretch}{1.2}
\begin{tabular}{lc}
\toprule
\textbf{Fatal Error Type} & \textbf{Score Cap} \\
\midrule
Governing equation incorrect      & 3 \\
Physical model invalid            & 4 \\
Dimensionally inconsistent        & 2 \\
Boundary conditions invalid       & 4 \\
\bottomrule
\end{tabular}
\vspace{0.3cm}
\caption{Fatal error caps. The step score is capped at the lowest applicable value when a fatal error is detected.}
\label{tab:fatal_caps}
\end{table}

\subsection{Meta-Evaluation Checks and Deductions}
\label{app:meta_evals_detailed}

\subsubsection{Missing Stage Penalty (\text{MP})}
To ensure that models do not bypass intermediate cognitive stages (e.g., jumping directly to a final answer without stating assumptions or selecting equations), the rule-based parser scans the output for the eight required markup tags.
\begin{itemize}
    \item \textbf{Deduction Rule:} For every missing reasoning stage, a flat penalty of $2.0$ points is added to the total missing penalty ($MP$):
    \begin{equation}
        MP = 2.0 \times \left| \{ \text{absent required stages} \} \right|
    \end{equation}
    This penalty is subtracted directly from the base score(aggregated score from all the stages) before applying meta-evaluation multipliers: $S_{\text{blend}} = \max(0, S_{\text{base}} - MP)$.
\end{itemize}

Beyond step-level grading, EngJudge applies three solution-level \emph{meta-evaluations}: Coverage, verbosity and physical sanity, to assess the overall quality, completeness, and physical grounding of the derivation. These checks are implemented via LLM-as-a-judge and evaluates the full solution and not stage-wise.

\subsubsection{Coverage Penalty (\texttt{COVERAGE})}
The coverage check verifies whether the model has addressed all sub-questions (e.g., part (a) and part (b)) and calculated all requested engineering quantities. 
\begin{itemize}
    \item \textbf{Evaluation Rubric:} The evaluator model evaluates the solution on a scale of $0$ to $10$. A score of $10$ indicates complete coverage. For each missing or unaddressed sub-question or target quantity, the score is reduced. This gives us the coverage score (CS)
    \item \textbf{Penalty Mapping:} The coverage score (CS) is mapped to a fractional penalty $\mathtt{COVERAGE} \in [0,\, 0.5]$ using:
    \begin{equation}
        \mathtt{COVERAGE} = \min\left(0.5,\ (10 - \mathtt{CS}) \times 0.10\right)
    \end{equation}
    Each point below $10$ deducts $10\%$ from the final score, capped at a maximum deduction of $50\%$.
\end{itemize}

\subsubsection{Verbosity Penalty (\texttt{VERBOSITY})}
Large Language Models frequently generate excessive, redundant steps or boilerplate explanations\citep{10.5555/3666122.3668142}. The verbosity check penalizes solutions that contain excessive filler text, unnecessary restatements of the problem, or repetitive derivations.
\begin{itemize}
    \item \textbf{Evaluation Rubric:} The evaluator model rates the conciseness and technical directness of the solution from $0$ to $10$. A score of $10$ represents a clean, direct derivation with minimal filler text. This gives us the verbosity score(VS).
    \item \textbf{Penalty Mapping:} The verbosity score (VS) maps to a fractional penalty $\mathtt{VERBOSITY} \in [0,\, 0.5]$ using:
    \begin{equation}
        \mathtt{VERBOSITY} = \min\left(0.5,\ (10 - \mathtt{VS}) \times 0.10\right)
    \end{equation}
    Similar to coverage, each point of excessive verbosity deducts $10\%$, up to a maximum cap of $50\%$. We cap it at 0.5 as we don't want it to bring the overall final score down to zero, same goes for Coverage.

\end{itemize}

\subsubsection{Physical Sanity Check (\texttt{SANITY\_FAIL})}
This is a critical global filter checking for high-level reasoning and physical consistency. Even if a model's intermediate steps appear mathematically structured, the final numerical values might be physically impossible. This is different from the earlier fatal error caps, as this checks the final solution/answer.
\begin{itemize}
    \item \textbf{Evaluation Rubric:} The evaluator model performs a binary check ($\mathtt{SANITY\_FAIL} \in \{0, 1\}$) on whether the solution violates fundamental physical laws. Examples of failure conditions include:
    \begin{itemize}
        \item Violating the Second Law of Thermodynamics (e.g., heat engines exceeding Carnot efficiency).
        \item Violating mass or energy conservation (e.g., fluid outflow exceeding inflow in steady state).
        \item Generating impossible physical bounds (e.g., negative absolute temperatures in Kelvin, efficiencies greater than 1.0, or velocities exceeding the speed of light).
    \end{itemize}
    \item \textbf{Penalty Mapping:} If any physical sanity check fails, $\mathtt{SANITY\_FAIL} = 1$, which applies a global multiplicative factor of $0.9^{\mathtt{SANITY\_FAIL}} = 0.9$ (a flat $10\%$ penalty) to the final score.
\end{itemize}

Final score is calculated as:
\begin{equation}
S_{\text{final}} = S_{\text{blend}} \cdot (1 - \mathtt{COVERAGE}) \cdot (1 - \mathtt{VERBOSITY}) \cdot 0.9^{\mathtt{SANITY\_FAIL}}
\end{equation}

\section{Experiments}
The table \ref{tab:main_table_appendix} contains all the scores obtained by using different generators, evaluators, and evaluation techniques.
\begin{table}[t]
\centering
\small
\setlength{\tabcolsep}{5pt}

\begin{tabular}{llccccc|c}
\toprule
\textbf{Generator} & \textbf{Evaluator}
& \textbf{FM}
& \textbf{HMT}
& \textbf{MoM}
& \textbf{Thermo}
& \textbf{Dyn}
& \textbf{Overall} \\
\midrule

\multirow{6}{*}{Qwen3-VL-8B (CoT)}

& Qwen3-VL-32B-Ins (\textit{Baseline})
& 7.070 & 7.250 & 7.040 & 8.000 & 7.310 & 7.446 \\

& Gemini-3.1 Pro Preview (\textit{Baseline})
& 4.610 & 5.250 & 4.630 & 7.220 & 5.500 & 5.752 \\

& Qwen3-VL-32B-Ins (\textit{Baseline with Dep})
& 6.190 & 6.390 & 6.450 & 7.450 & 6.750 & 6.793 \\

& Gemini-3.1 Pro Preview (\textit{Baseline with Dep})
& 4.030 & 4.600 & 3.990 & 6.800 & 4.920 & 5.203 \\

\rowcolor{gray!12}
\cellcolor{white} & Qwen3-VL-32B-Ins (\textit{EngJudge})
& 0.199 & 0.219 & 0.239 & 0.623 & 0.495 & 0.416 \\

\rowcolor{gray!12}
\cellcolor{white} & Gemini-3.1 Pro Preview (\textit{EngJudge})
& 0.661 & 0.847 & 1.000 & 2.798 & 1.697 & 1.670 \\

\midrule

\multirow{6}{*}{Gemini-2.5-Flash (CoT)}

& Qwen3-VL-32B-Ins (\textit{Baseline})
& 9.090 & 9.090 & 8.830 & 9.140 & 8.250 & 8.855 \\

& Gemini-3.1 Pro Preview (\textit{Baseline})
& 7.400 & 7.700 & 7.220 & 9.130 & 7.560 & 8.001 \\

& Qwen3-VL-32B-Ins (\textit{Baseline with Dep})
& 8.850 & 8.770 & 8.670 & 8.970 & 8.060 & 8.646 \\

& Gemini-3.1 Pro Preview (\textit{Baseline with Dep})
& 7.000 & 7.180 & 6.810 & 8.880 & 7.170 & 7.632 \\

\rowcolor{gray!12}
\cellcolor{white} & Qwen3-VL-32B-Ins (\textit{EngJudge})
& 0.674 & 0.663 & 1.039 & 1.028 & 1.098 & 0.945 \\

\rowcolor{gray!12}
\cellcolor{white} & Gemini-3.1 Pro Preview (\textit{EngJudge})
& 2.337 & 2.008 & 3.252 & 3.334 & 2.853 & 2.869 \\

\bottomrule
\end{tabular}
\vspace{0.3cm}
\caption{
Cross-evaluator performance comparison across engineering domains. Scores are reported on a 0--10 scale. Gray rows correspond to the proposed EngJudge framework. FM = Fluid Mechanics, HMT = Heat \& Mass Transfer, MoM = Mechanics of Materials, Thermo = Thermodynamics, Dyn = Dynamics, and CoT = Chain-of-Thought~\cite{chain_of_thought2022}.
}
\label{tab:main_table_appendix}
\end{table}

\subsection{Plot against Baseline}
\begin{figure}[H]
\centering
\begin{tikzpicture}
\begin{axis}[
    ybar,
    width=\textwidth,
    height=4.5cm,
    ymin=0,
    ymax=10,
    ylabel={Score},
    symbolic x coords={FM,HMT,MoM,Thermo,Dyn},
    xtick=data,
    enlarge x limits=0.2,
    bar width=15pt,
    legend style={
        at={(0.5,1.02)},
        anchor=south,
        legend columns=2
    }
]

\addplot[
    fill=blue!30,
    draw=blue!70,
    nodes near coords,
    point meta=y,
    every node near coord/.append style={
        font=\scriptsize,
        yshift=2pt
    },
    nodes near coords={
        \pgfmathprintnumber[fixed,precision=2]{\pgfplotspointmeta}
    }
]
coordinates {
(FM,7.40)
(HMT,7.70)
(MoM,7.22)
(Thermo,9.13)
(Dyn,7.56)
};

\addplot[
    fill=red!25,
    draw=red!80,
    nodes near coords,
    point meta=y,
    every node near coord/.append style={
        font=\scriptsize,
        yshift=2pt
    },
    nodes near coords={
        \pgfmathprintnumber[fixed,precision=2]{\pgfplotspointmeta}
    }
]
coordinates {
(FM,2.337)
(HMT,2.008)
(MoM,3.252)
(Thermo,3.334)
(Dyn,2.853)
};

\legend{Baseline, EngJudge}

\end{axis}
\end{tikzpicture}
\caption{Comparison of baseline and EngJudge evaluation by \texttt{Gemini 3.1 Pro Preview}.}
\label{fig:baseline_vs_engjudge}
\end{figure}
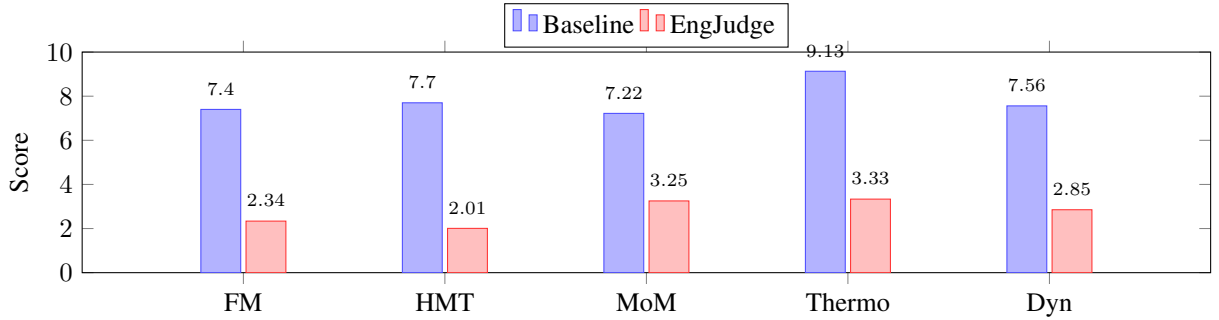

\subsection{Step wise scores}
\begin{table}[H]
\centering
\scriptsize
\renewcommand{\arraystretch}{1.1}
\begin{tabular}{lcccc}
\toprule
\textbf{Reasoning step} &
\multicolumn{2}{c}{\textbf{Qwen3-VL-8B}} &
\multicolumn{2}{c}{\textbf{Gemini 2.5 Flash}} \\
\cmidrule(lr){2-3} \cmidrule(lr){4-5}
& \textbf{Dep} & \textbf{Ind} & \textbf{Dep} & \textbf{Ind} \\
\midrule
Problem Characterization (PC) & 7.13 & 7.13 & 7.33 & 7.33 \\
Assumptions (AS)              & 4.42 & 5.85 & 6.01 & 7.93 \\
Visual Interpretation (VI)    & 4.47 & 7.49 & 5.81 & 8.72 \\
Equation Selection (ES)       & 2.85 & 4.40 & 5.48 & 8.14 \\
Logical Reasoning (LR)        & 2.17 & 4.11 & 4.71 & 7.48 \\
Algebraic Accuracy (AA)       & 1.38 & 2.30 & 2.74 & 4.34 \\
Physical Interpretation (PI)  & 2.47 & 4.96 & 4.53 & 7.51 \\
Final Answer (FA)             & 1.47 & 3.41 & 2.91 & 5.74 \\
\bottomrule
\end{tabular}
\vspace{0.3cm}
\caption{Data for average step-wise dependent (Dep) and Independent (Ind) scores, evaluated with \texttt{Gemini 3.1 Pro Preview} }
\label{tab:stepwise_breakdown}
\end{table}

\subsection{Detailed Topological and Correlation Analyses}
\label{sec:appendix_analyses}

We provide extended quantitative details on the choice of dependency structures  and the empirical correlations between reasoning steps.
\subsubsection{Influence of Dependency Topologies}
\label{sec:appendix_topologies}
To evaluate the choice of dependency mapping, we compare the proposed Directed Acyclic Graph against alternative topological structures in Table~\ref{tab:topology_ablation}. 
Excluding dependency propagation entirely (\emph{Independent/Flat}) yields optimistic base scores of $7.19$ (Gemini) and $5.18$ (Qwen). 
Conversely, imposing a sequential chain (\emph{Strict Linear Chain}), where each step $N$ depends entirely on step $N-1$, results in overly severe penalties ($4.29$ for Gemini, $2.93$ for Qwen). 
This is because a linear topology assumes that a minor mistake in an early independent step (e.g., a small assumption slip) completely invalidates parallel, unaffected tracks (such as reading dimensions from a diagram). 
Our proposed \emph{True DAG} models these parallel pathways accurately, penalizing dependent steps only when their specific prerequisites fail, resulting in balanced base scores of $5.03$ (Gemini) and $3.48$ (Qwen).
\begin{table}[H]
\centering
\scriptsize
\renewcommand{\arraystretch}{1.1}
\begin{tabular}{lcc}
\toprule
\textbf{Topology Type} &
\textbf{Qwen3-VL-8B} &
\textbf{Gemini 2.5 Flash} \\
\midrule
True DAG            & 3.48 & \textbf{5.03} \\
Strict Linear Chain & 2.93 & 4.29 \\
Independent (Flat)  & \textbf{5.18} & \textbf{7.19} \\
\bottomrule
\end{tabular}
\vspace{0.3cm}
\caption{Ablation showing the effect of different dependency structures on the average \textbf{base score} (excluding meta-evaluation checks and missing step penalties). Removing dependency propagation (Independent/Flat) yields substantially higher scores, while stricter dependency assumptions (Linear Chain) result in lower scores than the proposed DAG-based topology.}
\label{tab:topology_ablation}
\end{table}

\subsubsection{Correlation and Causal Dependency Analysis}
\label{sec:appendix_correlation}
To empirically validate the dependency relations modeled by our DAG, we analyze the Pearson correlation matrix between the raw step scores across all evaluations (shown in Figure~\ref{fig:app_corr_heatmap} in the main paper). The matrix reveals a clear, diagonal-adjacent correlation structure, where strong correlations are localized to neighboring steps of the reasoning chain. 

Specifically, we observe strong correlations between Algebraic Accuracy and the Final Answer ($r=0.69$), and between Equation Selection and Logical Reasoning ($r=0.53$). These findings reflect the physical reality of error propagation: a failure in selecting the correct governing equation or executing mathematical derivations directly compromises downstream calculations and the final numerical output. Conversely, correlations between early conceptual steps and final execution steps are negligible (e.g., Problem Characterization vs. Final Answer yields $r=0.06$). This demonstrates that a VLM's ability to classify the physics domain of a problem is independent of its ability to execute the math required to solve it, reinforcing the need for step-wise evaluation.

\paragraph{Why Statistical Correlation Differs from Causal DAG Dependencies?}
This is a key design question: why are some pairs of steps with high statistical correlation in Figure~\ref{fig:app_corr_heatmap} not connected by direct dependency edges in our DAG? For instance, Physical Interpretation (\texttt{PI}) and Final Answer (\texttt{FA}) exhibit a high correlation of $r=0.60$, and Algebraic Accuracy (\texttt{AA}) and Physical Interpretation (\texttt{PI}) show $r=0.52$, yet neither pair has a direct edge in the evaluation graph. This design choice is guided by three main principles:

\begin{enumerate}
    \item \textbf{Causality vs. Correlation:} The DAG is designed to enforce direct, physical causal prerequisites. For example, a student can mathematically compute a correct final numerical answer (\texttt{FA}) via correct algebraic manipulation (\texttt{AA}) without necessarily understanding or explaining its physical meaning (\texttt{PI}). Because \texttt{PI} is not a strict mathematical prerequisite for calculating \texttt{FA}, drawing a dependency edge from \texttt{PI} to \texttt{FA} would be causally incorrect, despite their statistical correlation.
    
    \item \textbf{Confounding by Downstream Position:} Later steps in the reasoning chain (such as \texttt{AA}, \texttt{PI}, and \texttt{FA}) are strongly correlated because they are co-dependent on the cumulative errors of early steps (like \texttt{AS} and \texttt{ES}). This shared dependency on common ancestors creates high statistical correlation (confounding) in the empirical data. Adding redundant edges between these downstream variables in the DAG would lead to duplicate penalization for the same upstream error.
    
    \item \textbf{Conceptual Independence in Rubrics:} Conceptual reasoning (such as qualitative physical interpretation) and algebraic computation are graded as independent dimensions in standard engineering pedagogy. A model may fail the algebra but perform a correct physical limit check, or vice versa. The high empirical correlation ($r=0.52$ between \texttt{AA} and \texttt{PI}) is a reflection of general model capability (i.e., stronger VLMs perform well on both, while weaker models fail on both) rather than a causal dependency between the two skills.
\end{enumerate}
Thus, this correlation does not necessarily validate our dependency graph, but it does support the claims made in the graph to some extent.

\subsection{Qualitative Case Studies}
\label{sec:appendix_qualitative_cases}

This section details the primary failure modes observed in the generated solutions, categorized by the nature of the cognitive or mathematical lapse. To illustrate these mechanisms, we present four distinct examples spanning four different engineering domains.

\subsubsection{Visual Interpretation Errors}
Despite demonstrating strong general visual recognition, vision-language models (VLMs) frequently fail to map two-dimensional diagrams of three-dimensional systems into correct mathematical formulations, or overlook subtle geometric features that alter the boundary conditions.
\begin{itemize}
    \item \textbf{Omission of Physical Constraints (Thermodynamics, Problem 5-131):} When modeling a piston-cylinder device under compression, the VLM failed to identify the physical stops inside the cylinder wall depicted in the schematic. Consequently, it modeled the entire compression process as isobaric under the assumption that the piston was free to move indefinitely. In reality, once the piston contacts the stops, the process transitions to a constant-volume phase, rendering the subsequent state and boundary work calculations invalid.
\end{itemize}

\subsubsection{Physical Principles and Boundary Conditions Errors}
A major source of low evaluation scores is the formulation of equations that violate the system's boundary conditions, flow characteristics, or conservation laws.
\begin{itemize}
    \item \textbf{Incorrect Boundary Pressure Formulation (Fluid Mechanics, Problem 11):} In analyzing the force required to hold an exit nozzle plug in place, the model applied Bernoulli's equation to calculate a non-zero exit pressure ($p_2 \approx 364\text{ kPa}$). Because the nozzle discharges directly to the atmosphere as an unconfined jet, the correct physical boundary condition is a uniform gauge pressure of zero ($p_2 = 0$). Introducing a fictitious pressure force at the boundary led to a physically invalid linear momentum balance.
\end{itemize}

\subsubsection{Dimensionless Group and Correlation Errors}
In domains involving convective transport and non-dimensional scaling, models frequently apply empirical correlations outside their valid regimes or evaluate fluid properties at incorrect reference states.
\begin{itemize}
    \item \textbf{Incorrect Reference Temperature Property Evaluations (Heat \& Mass Transfer, Problem 71):} During the forced convection cooling analysis of an electronic chip, the model evaluated the physical properties of the air stream (density, dynamic viscosity, and thermal conductivity) at the free-stream temperature of $25^\circ\text{C}$ (300 K) instead of the film temperature ($T_f = (T_s + T_\infty)/2 \approx 308\text{ K}$). This property mismatch introduced systematic errors into the Reynolds ($Re$) and Prandtl ($Pr$) numbers, which propagated through the Nusselt number correlation and yielded an incorrect convective heat transfer coefficient.
\end{itemize}

\subsubsection{Arithmetic and Unit Discrepancies}
Even when the physical formulation is sound, final numerical answers are often invalidated by basic calculation slips, incorrect coordinate transformations, or lookup errors.
\begin{itemize}
    \item \textbf{Kinematic and Component Formulation Slips (Dynamics, Problem 21-77):} When calculating the angular momentum of a precessing satellite, the model equated the satellite's spin rate directly to the total angular velocity component along the symmetry axis ($\psi_z = \omega_s$), ignoring the precession contribution and the coordinate rotation. This algebraic process error resulted in a cascading numerical slip, underestimating the satellite's angular momentum by approximately $45\%$.
\end{itemize}

\section{Detailed Human Study Results}
\label{app:human_study}

This appendix provides a detailed quantitative breakdown of the human validation study referenced in Section \ref{sec:human_study}. The study utilized a custom-built, interactive evaluation dashboard that sequentially guided evaluators through randomly assigned problems per session. Participants first performed a binary rating (correct/incorrect) on each generated step without seeing the framework's evaluation, minimizing anchoring bias \cite{tversky1974judgment}. Subsequently, they were shown the framework's dependency score and penalty justifications, and were asked to rate the severity on a 5-point Likert scale (ranging from ``Much too low'' to ``Much too high'').

\subsection{Data Point Formulation and Rating Quantification}
To perform continuous statistical analysis on qualitative feedback, we define a structured mapping to convert the qualitative human ratings into numerical scores.

\paragraph{Data Point Definition}
A single evaluation data point represents the rating of a single reasoning unit (either a stage-level evaluation or a meta-evaluation check) for a given problem instance by a human participant. 
With 9 expert participants evaluating 4 randomized problems each, the total number of data points is:
\begin{equation}
    N_{\text{total}} = N_{\text{step}} + N_{\text{meta}} = 285 + 108 = 393
\end{equation}
where $N_{\text{step}} = 285$ represents individual stage evaluations (8 stages per problem, with occasional missing steps or parsing fallbacks omitted), and $N_{\text{meta}} = 108$ represents meta-evaluation evaluations (3 checks per problem: Verbosity, Coverage, and Physical Sanity).

\paragraph{Likert-to-Numerical Mapping}
During the study, human experts did not assign raw scores directly; instead, they reviewed the framework's automated score ($S_{\text{auto}} \in [0, 10]$) and selected a qualitative rating on a 5-point Likert scale to express their level of agreement. 
To reconstruct the absolute human score ($S_{\text{human}}$), we define an adjustment mapping ($\delta$) for each Likert category:
\begin{align}
    \text{``Much too low''} &\implies \delta = +2 \nonumber \\
    \text{``Slightly too low''} &\implies \delta = +1 \nonumber \\
    \text{``About right''} &\implies \delta = 0 \label{eq:likert_mapping} \\
    \text{``Slightly too high''} &\implies \delta = -1 \nonumber \\
    \text{``Much too high''} &\implies \delta = -2 \nonumber
\end{align}
The equivalent human score is then reconstructed by adjusting the automated score and clipping the result to the valid $[0, 10]$ grading scale:
\begin{equation}
    S_{\text{human}} = \max(0, \min(10, S_{\text{auto}} + \delta))
\end{equation}
This formulation ensures a human-consistent numerical scale, where a rating of ``About right'' ($\delta=0$) corresponds to perfect agreement ($S_{\text{human}} = S_{\text{auto}}$). 
Pearson correlation ($r$) and Mean Absolute Error (MAE) are then calculated directly using the paired arrays of $S_{\text{auto}}$ and $S_{\text{human}}$ across the $393$ data points.

\subsection{Continuous Alignment Metrics}

\begin{figure}[H]
    \centering
    \includegraphics[width=0.45\textwidth]{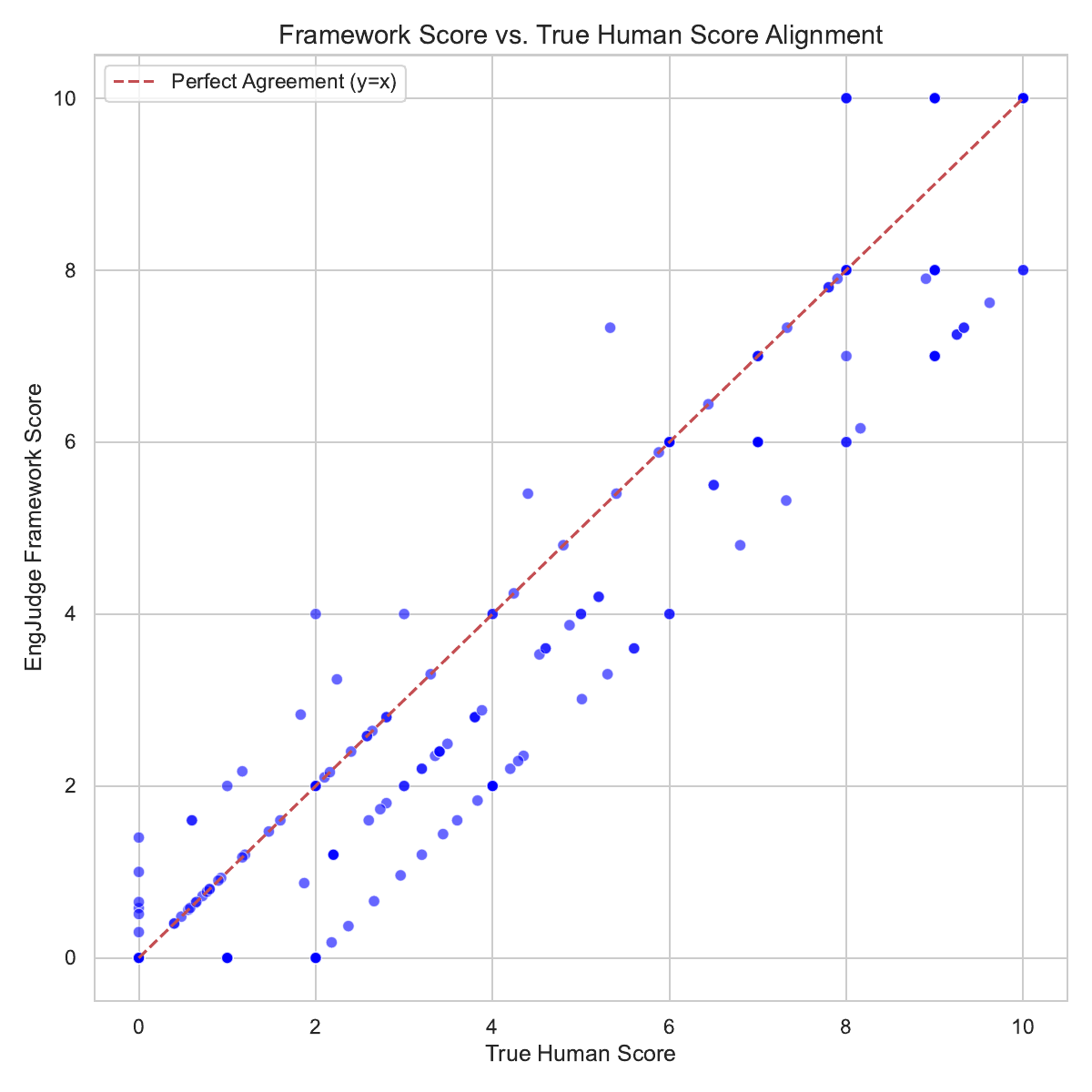}
    \caption{Alignment between synthetic human scores and automated our framework scores ($N=393$). Points tightly clustering around the $y=x$ diagonal indicate exceptionally strong agreement.}
    \label{fig:scatter_alignment}
\end{figure}

The continuous alignment between the framework's automated scores and the synthetic human scores is visually depicted in Figure \ref{fig:scatter_alignment} and quantitatively summarized in Table \ref{tab:human_study_metrics}.
\begin{table}[H]
\centering
\small

\begin{tabular}{lccc}
\toprule
\textbf{Category} &
\textbf{Data Points} &
\textbf{Pearson Corr.} &
\textbf{MAE} \\
\midrule
Step-Level & 285 & 0.9672 & 0.7068 \\
Meta-Eval. & 108 & 0.9859 & 0.5648 \\
\midrule
\textbf{Overall} & \textbf{393} & \textbf{0.9749} & \textbf{0.6678} \\
\bottomrule
\end{tabular}
\vspace{0.3cm}
\caption{Quantitative Alignment of our framework vs. Human Expert Evaluators}
\label{tab:human_study_metrics}
\end{table}

\subsection{Meta-Evaluation Breakdown}
Beyond sequential steps, the framework assesses complete solutions through three meta-checks. The human alignment for these individual checks is highly consistent:
\begin{itemize}
    \item \textbf{Verbosity ($N=36$):} $r = 0.983$, MAE = 0.94.
    \item \textbf{Coverage ($N=36$):} $r = 0.987$, MAE = 0.36.
    \item \textbf{Physical Sanity ($N=36$):} $r = 0.974$, MAE = 0.39.
\end{itemize}

\subsection{Qualitative Agreement and Evaluator Preferences}
\begin{figure}[H]
    \centering
    \includegraphics[width=0.5\linewidth]{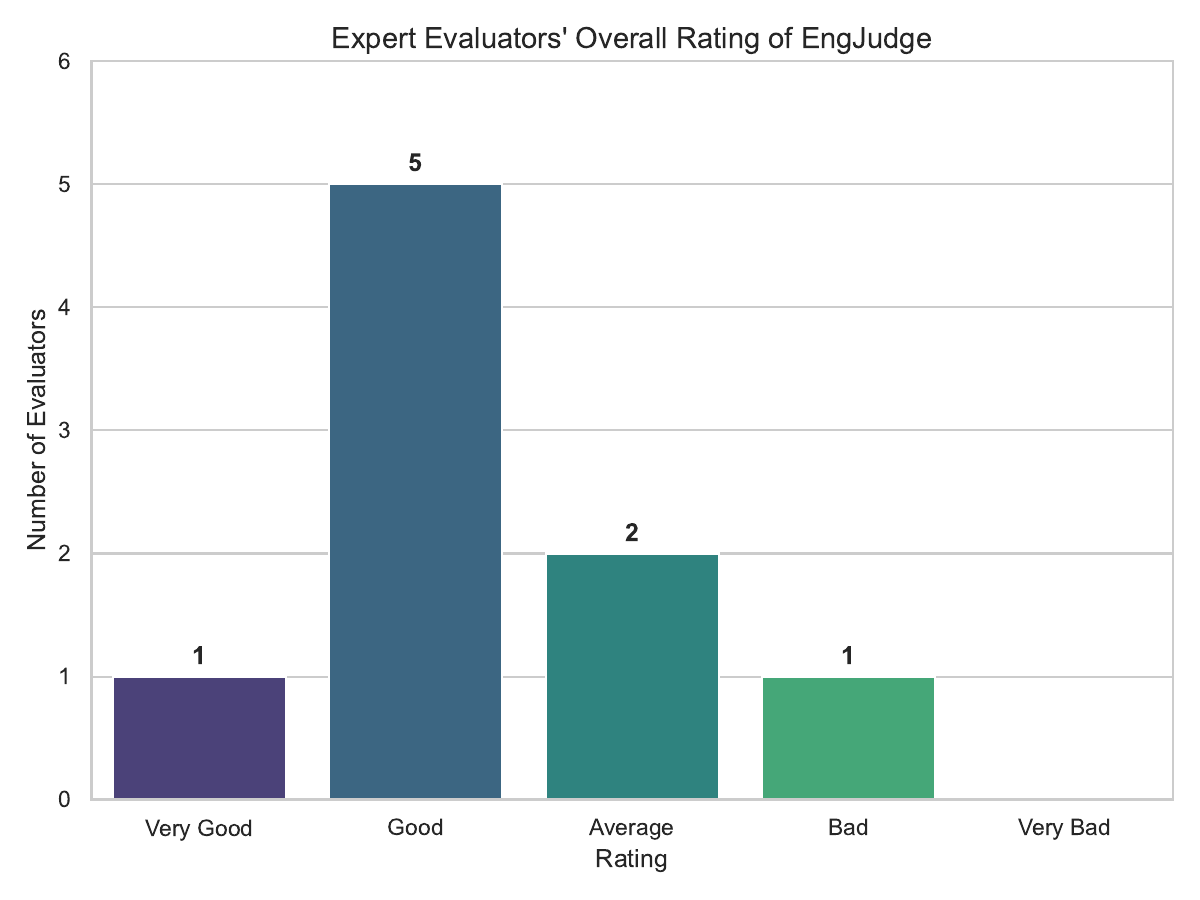}
    \caption{Distribution of overall framework ratings assigned by human expert evaluators.}
    \label{fig:overall_rating}
\end{figure}
\begin{figure}[H]
    \centering
    \includegraphics[width=0.5\linewidth]{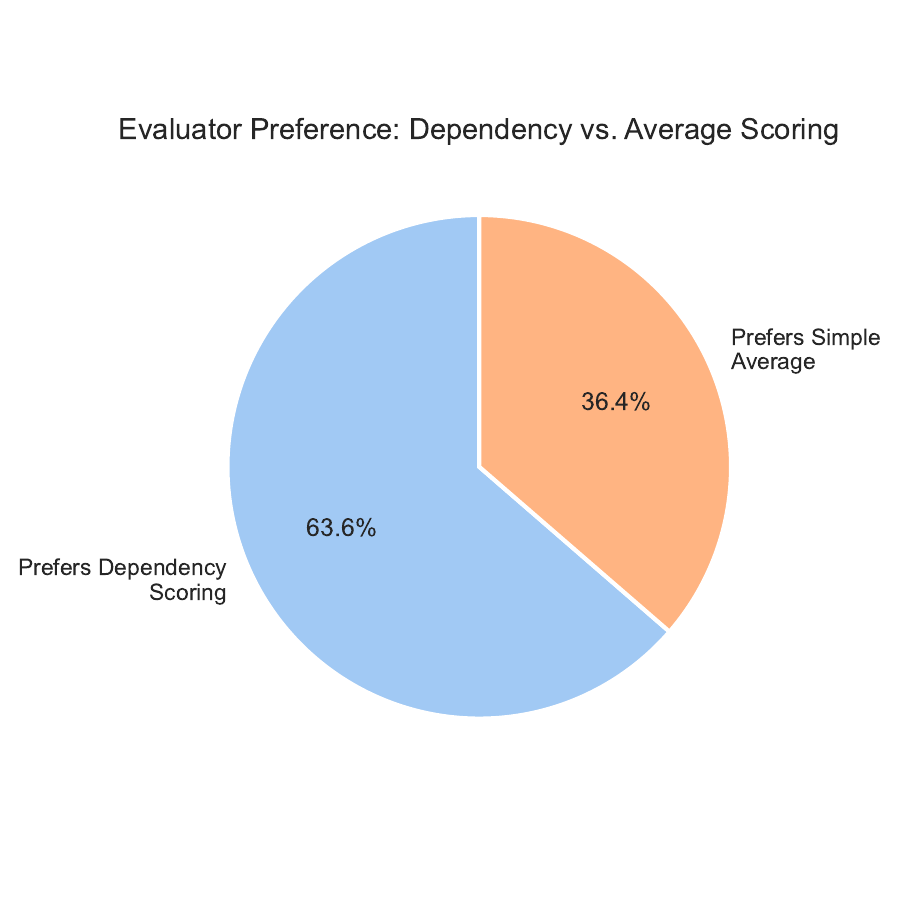}
    \caption{Evaluator preference for dependency-based scoring vs. a naive unweighted average.}
    \label{fig:dependency_preference}
\end{figure}

In addition to continuous correlations, the study measured exact categorical agreement. For \textbf{48.9\% (192/393)} of all evaluation points, human evaluators selected ``About right'', indicating perfect exact agreement with the numerical score assigned by the framework without any need for adjustment. 

When observing the overall framework ratings (Figure \ref{fig:overall_rating}), 5 evaluators rated it ``Good'' (55.6\%) and 1 rated it ``Very Good'' (11.1\%). Furthermore, the 2-to-1 win rate (66.7\%, Figure \ref{fig:dependency_preference}) for the dependency scoring method over an unweighted average highlights a critical pedagogical reality in engineering: \emph{a correct algebraic derivation is rendered meaningless if the fundamental assumptions or physical constraints formulated in prior steps are flawed}. The human study confirms that EngJudge successfully captures this nuanced grading philosophy.

\begin{figure}[H]
    \centering
    \includegraphics[width=0.75\textwidth]{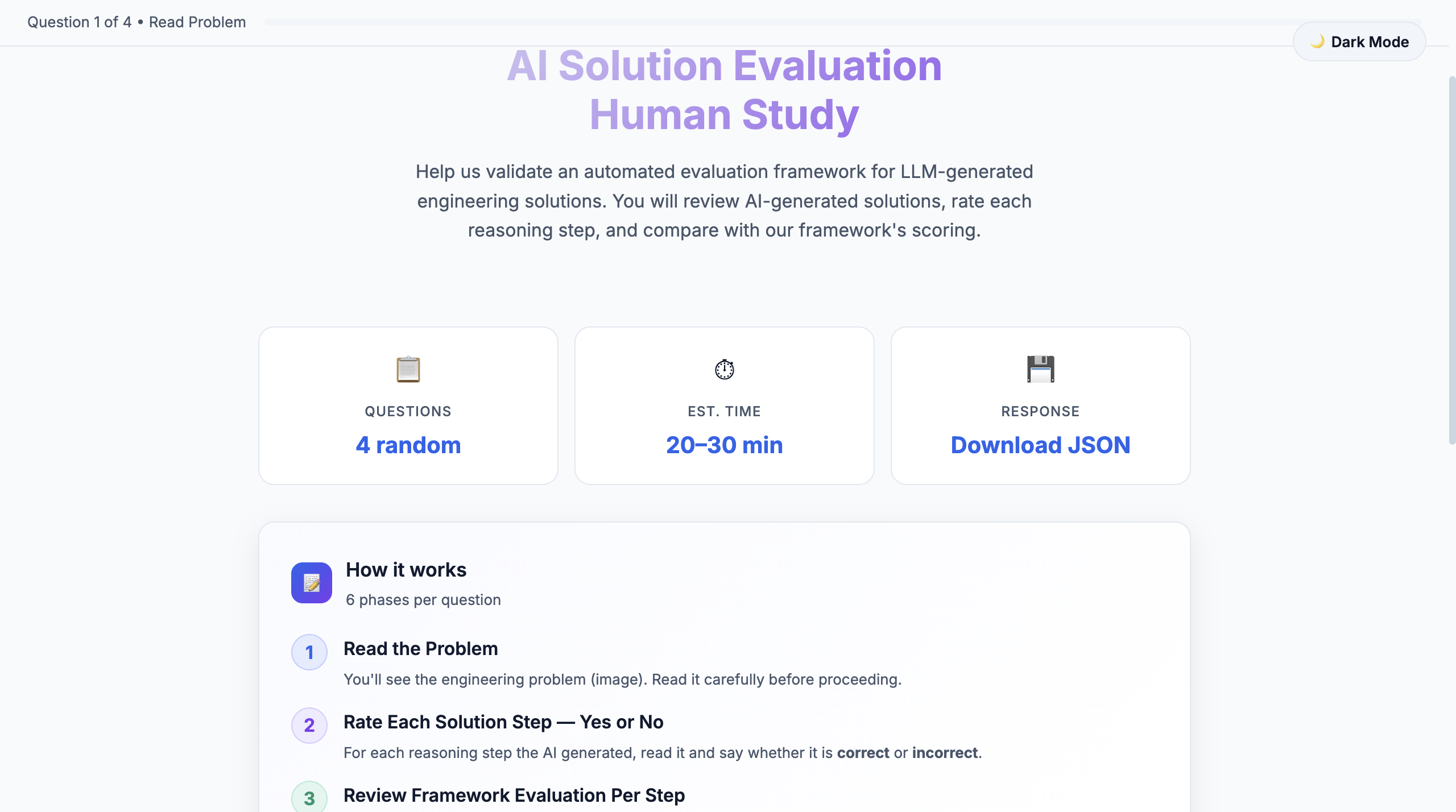}
    \vspace{0.5em}
    \includegraphics[width=0.75\textwidth]{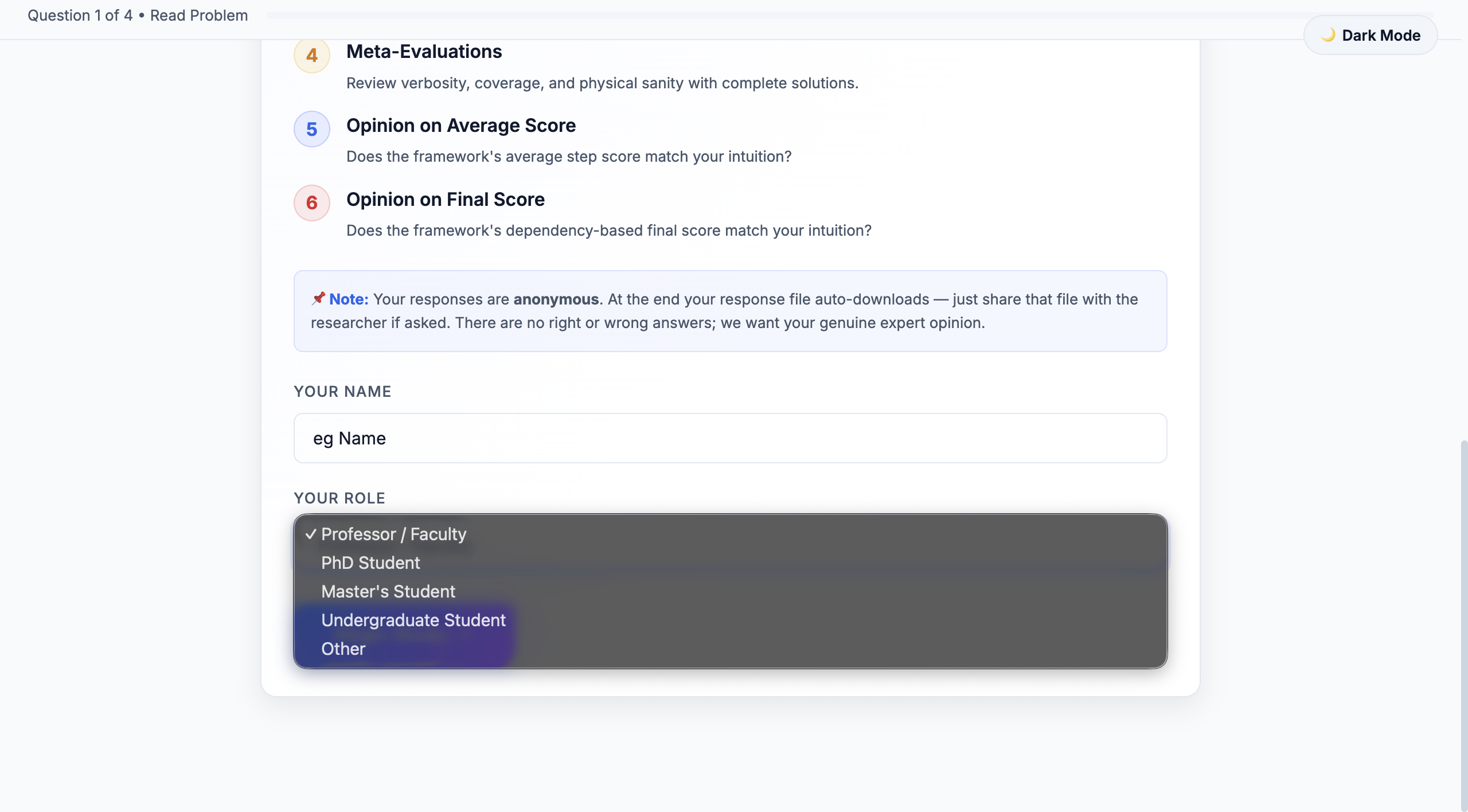}
    \caption{
    Introductory pages of annotation webpage.
    }
    \label{fig:human_study_ss_1_and_2}
\end{figure}

\begin{figure}[H]
    \centering
    \includegraphics[width=0.75\textwidth]{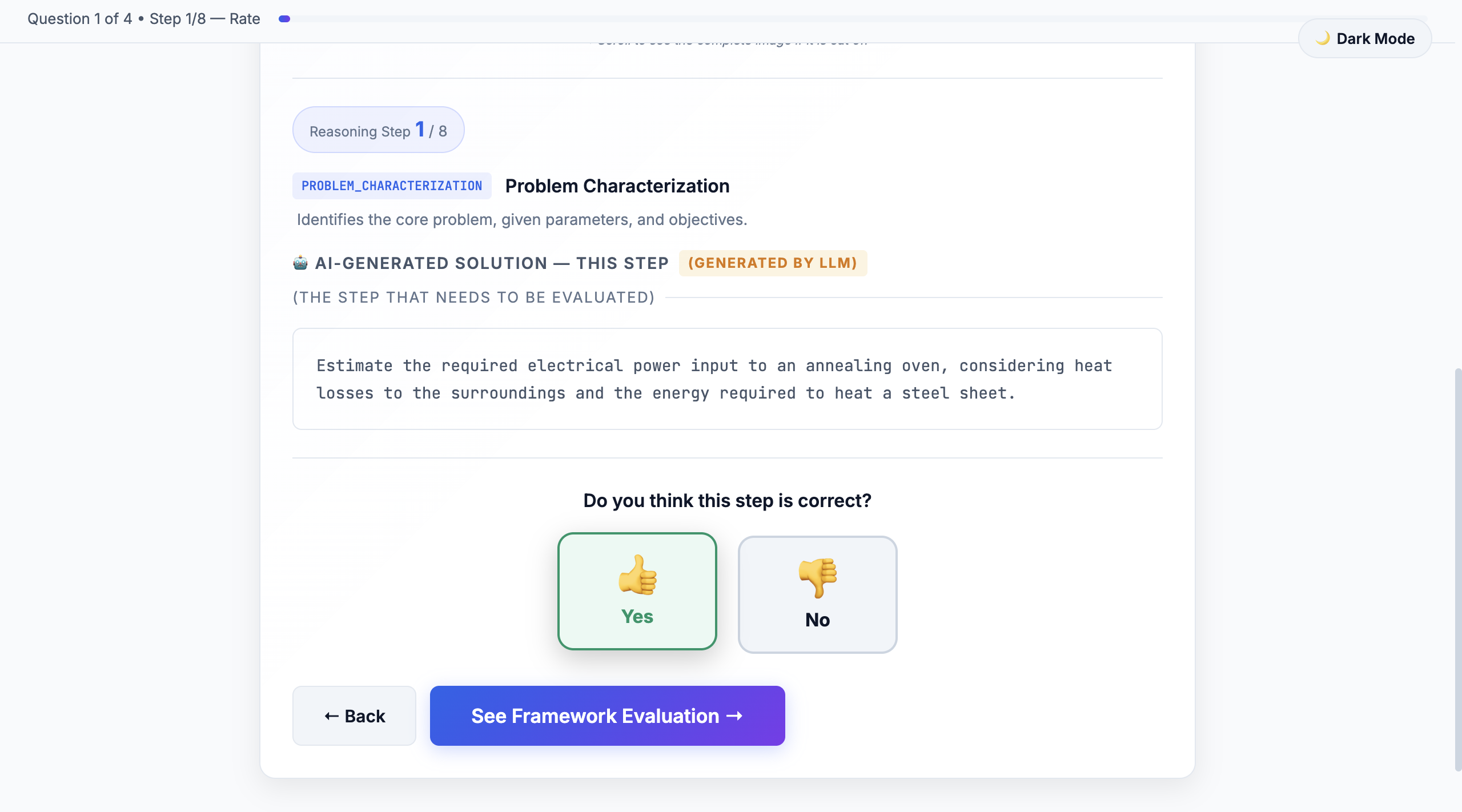}
    \caption{Users are asked to judge whether the LLM-generated solution step is correct or not for a given question.}
    \label{fig:webpage_ss4}
\end{figure}

\begin{figure}[H]
    \centering
    \includegraphics[width=0.75\textwidth]{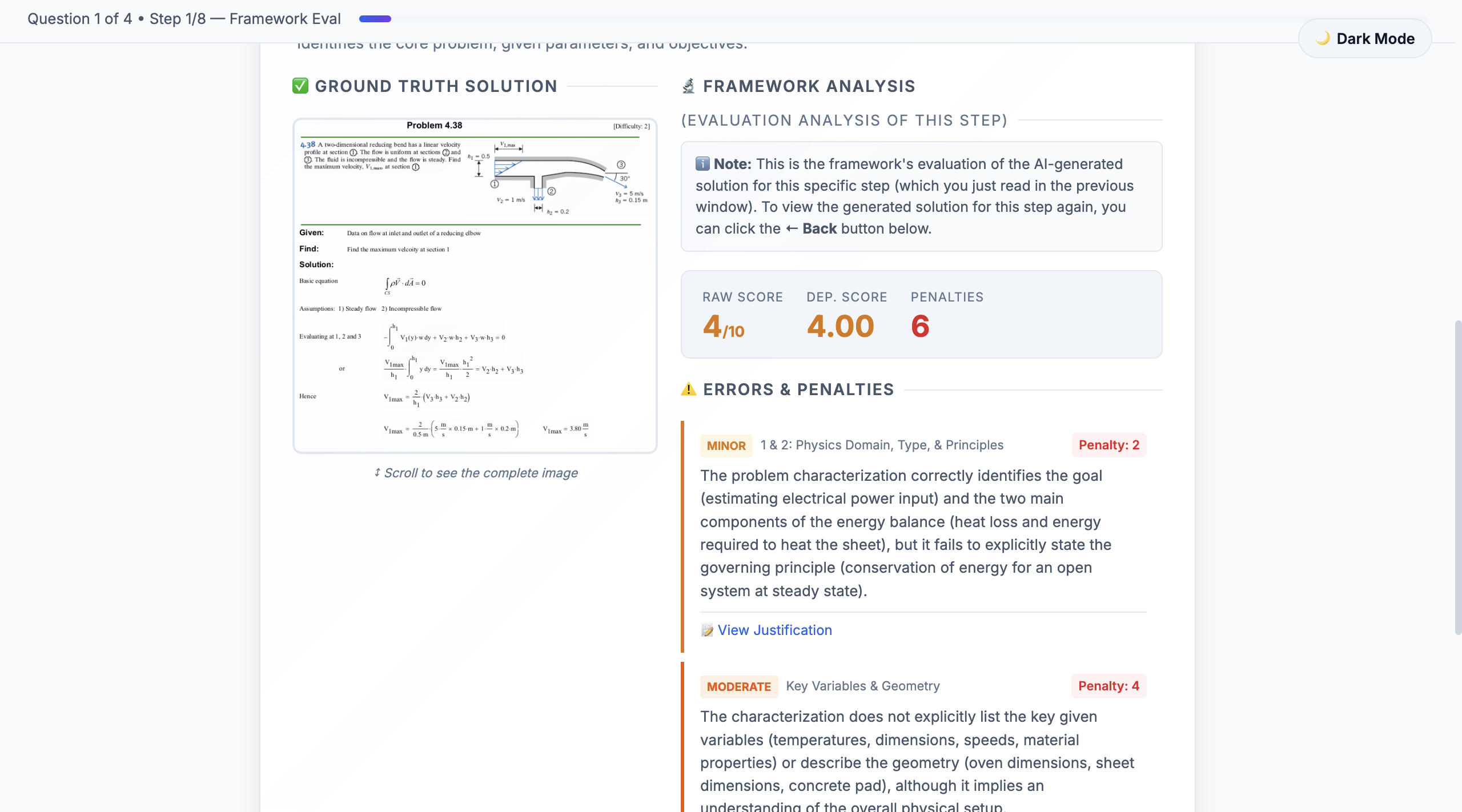}
    \vspace{0.5em}
    \includegraphics[width=0.75\textwidth]{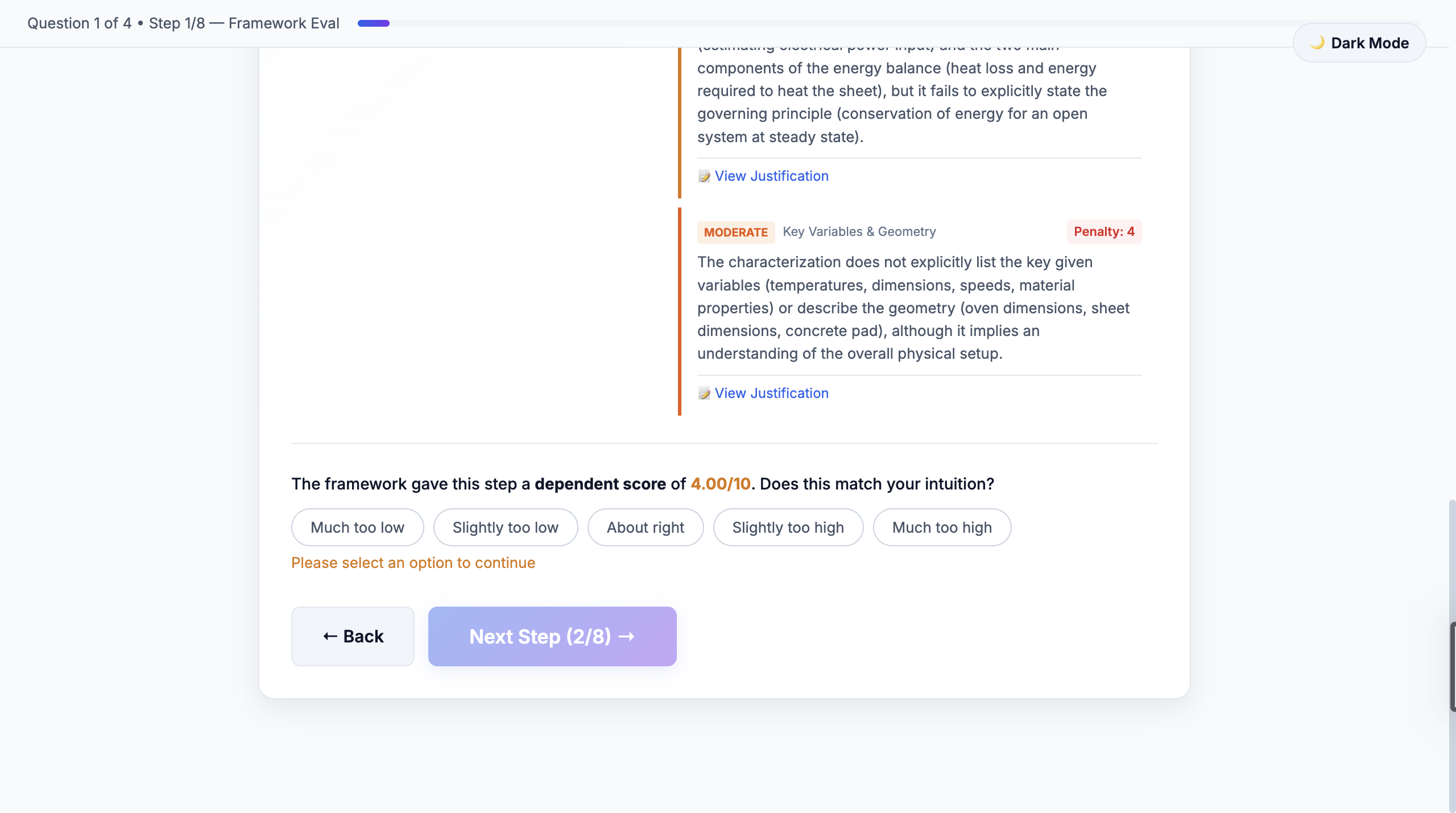}
    \caption{Users are shown the ground truth solution and the step evaluation of our framework. Users are required to rate it. This continues for all 8 stages.}
    \label{fig:human_study_ss_5_and_6}
\end{figure}

\begin{figure}[H]
    \centering
    \includegraphics[width=0.75\textwidth]{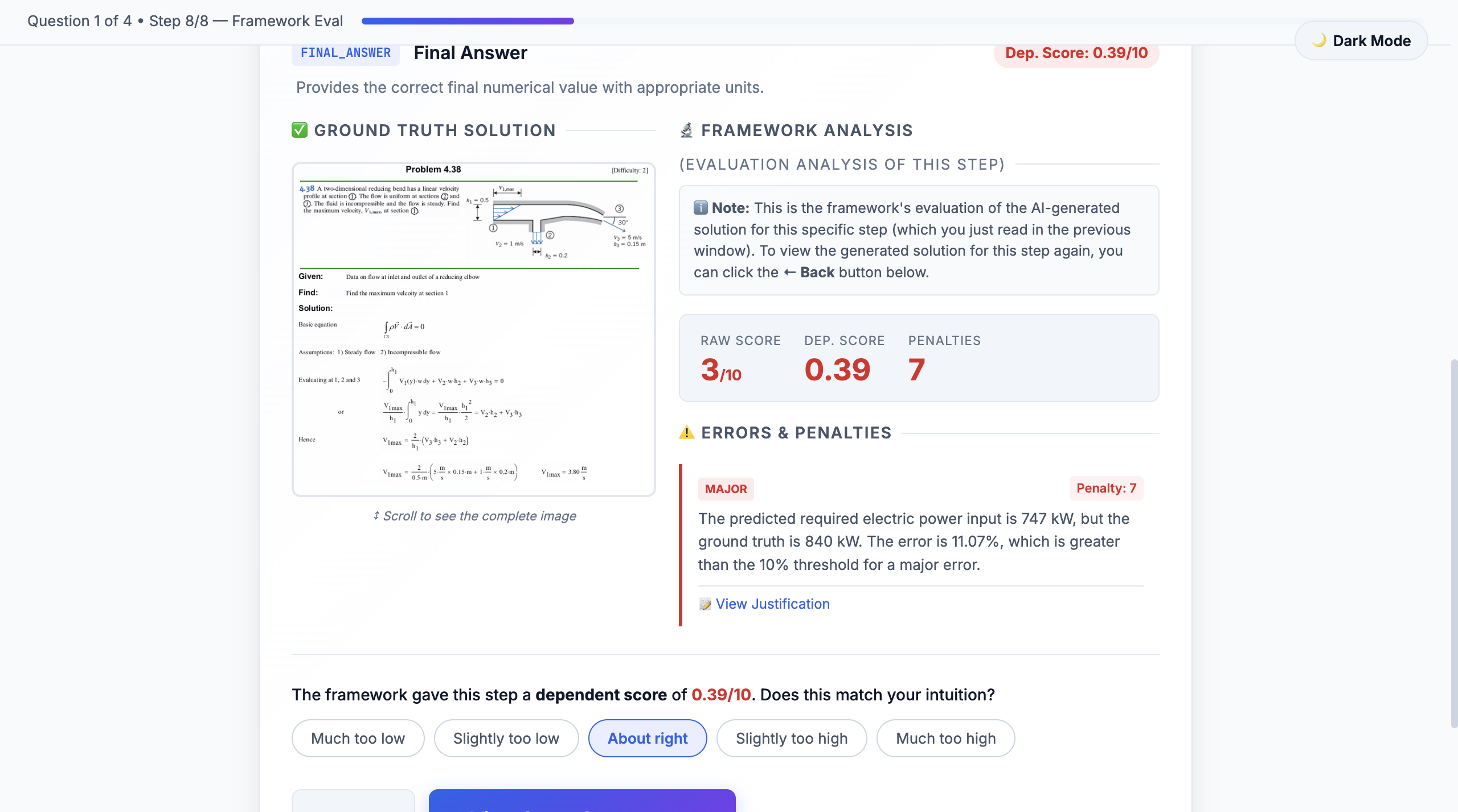}
    \caption{8th stage (Final answer) and its evaluation is shown, and asked to rate.}
    \label{fig:webpage_ss7}
\end{figure}

\begin{figure}[H]
    \centering
    \includegraphics[width=0.75\textwidth]{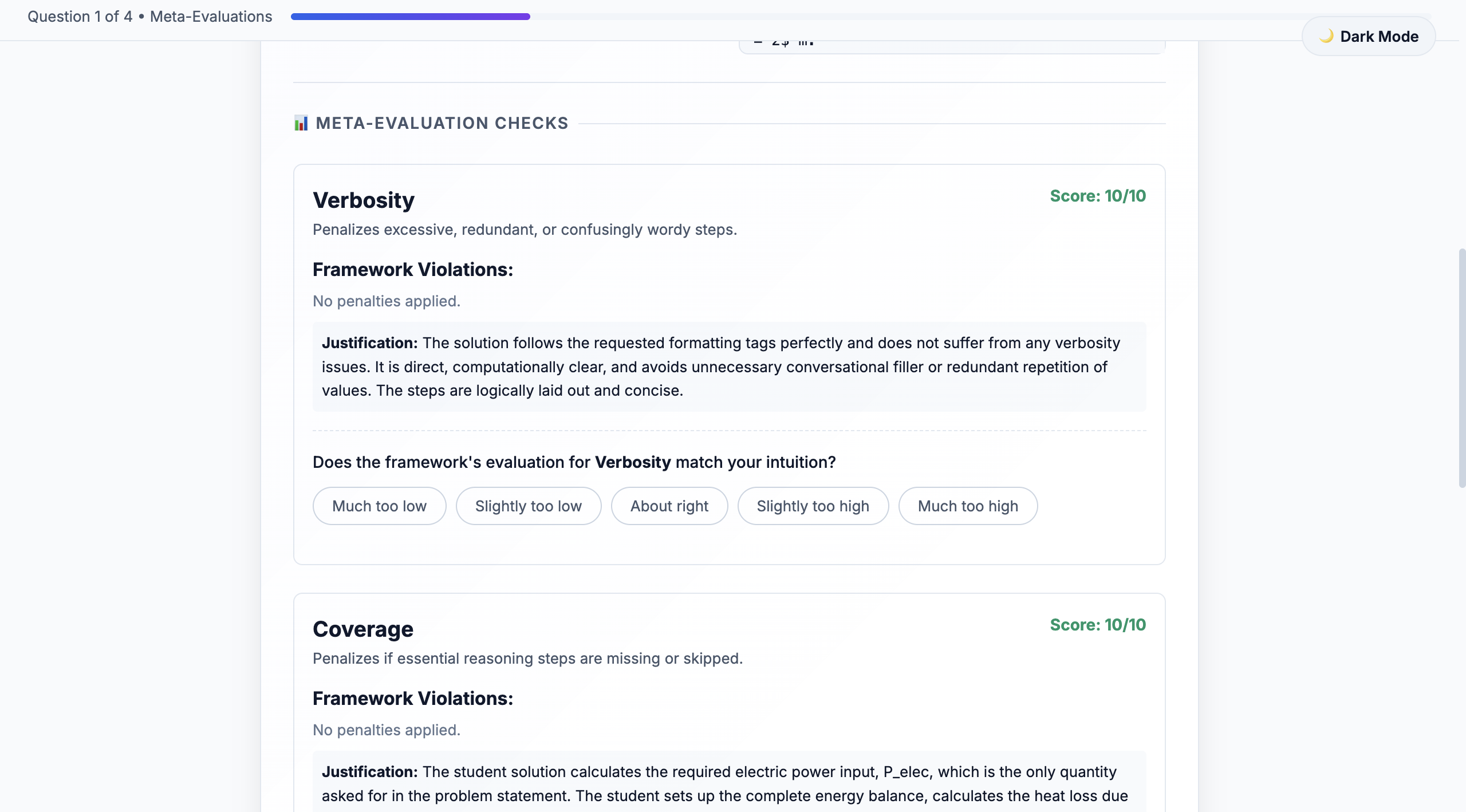}
    \vspace{0.5em}
    \includegraphics[width=0.75\textwidth]{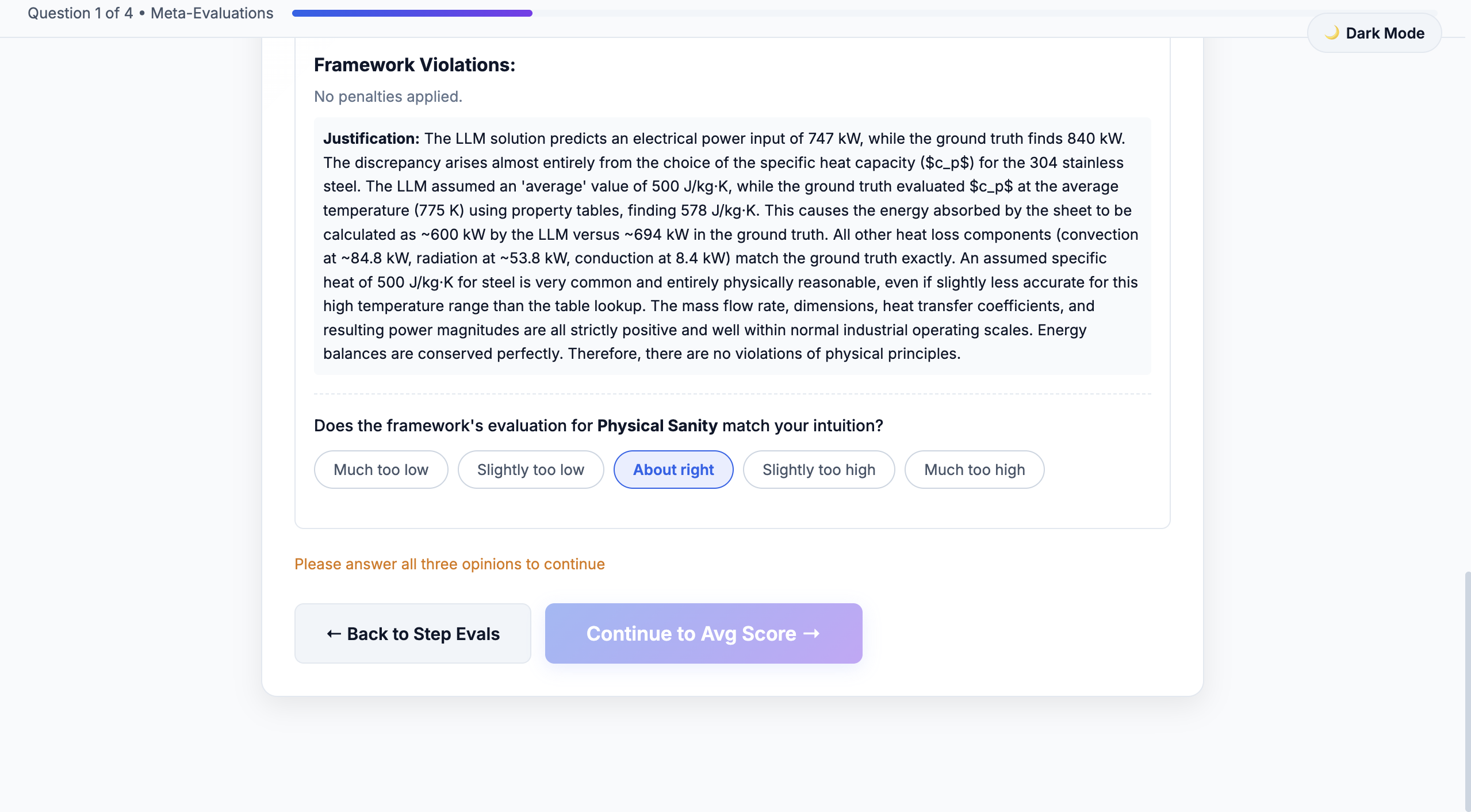}
    \caption{Meta evaluation checks are shown and asked to be rated.}
    \label{fig:human_study_ss_8_and_9}
\end{figure}

\begin{figure}[H]
    \centering
    \includegraphics[width=0.75\textwidth]{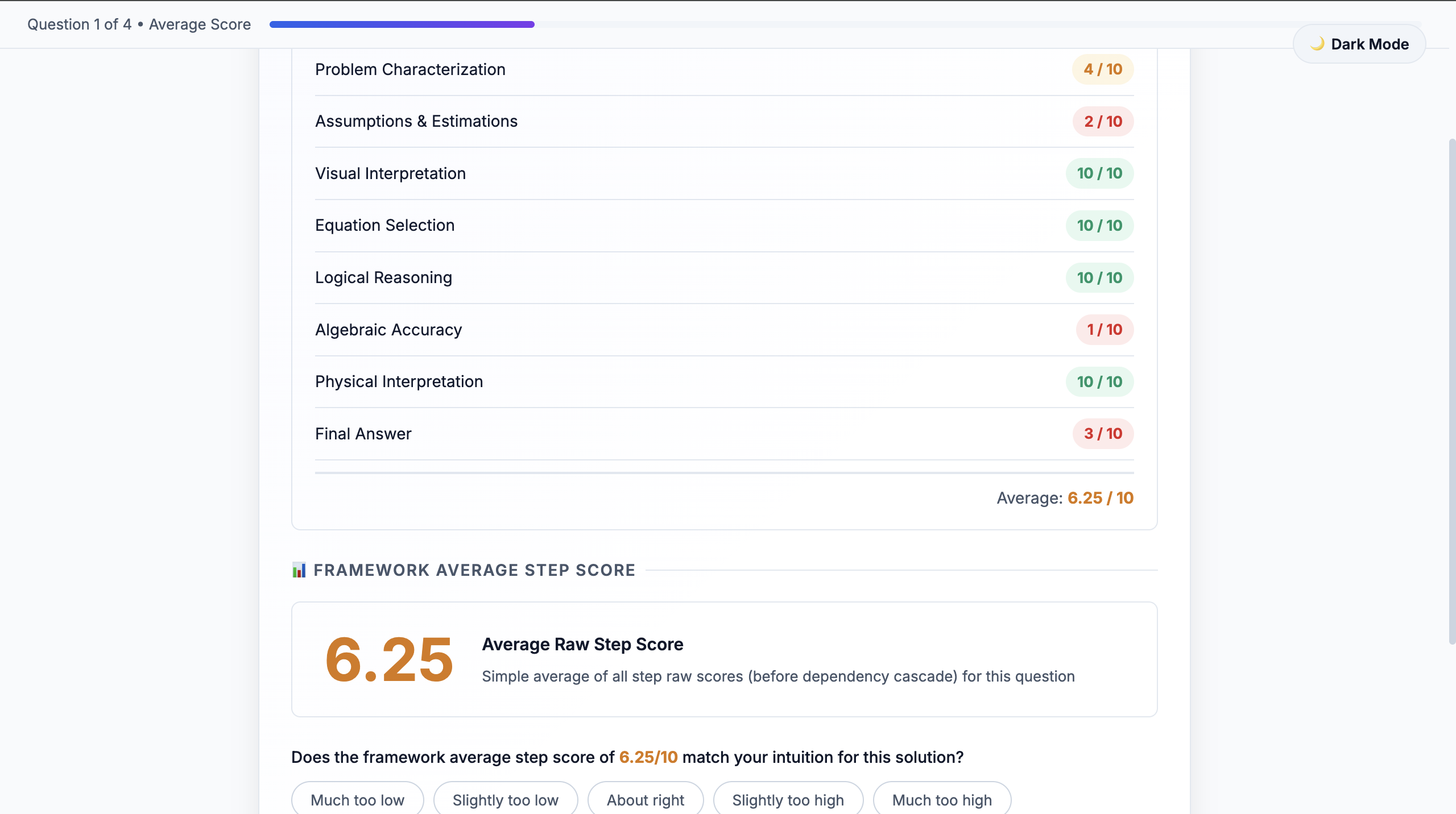}
    \caption{After 11 steps, final independent stage-wise scores are shown. Users are asked to choose if they feel that the average score is right.}
    \label{fig:webpage_ss10}
\end{figure}

\begin{figure}[H]
    \centering
    \includegraphics[width=0.75\textwidth]{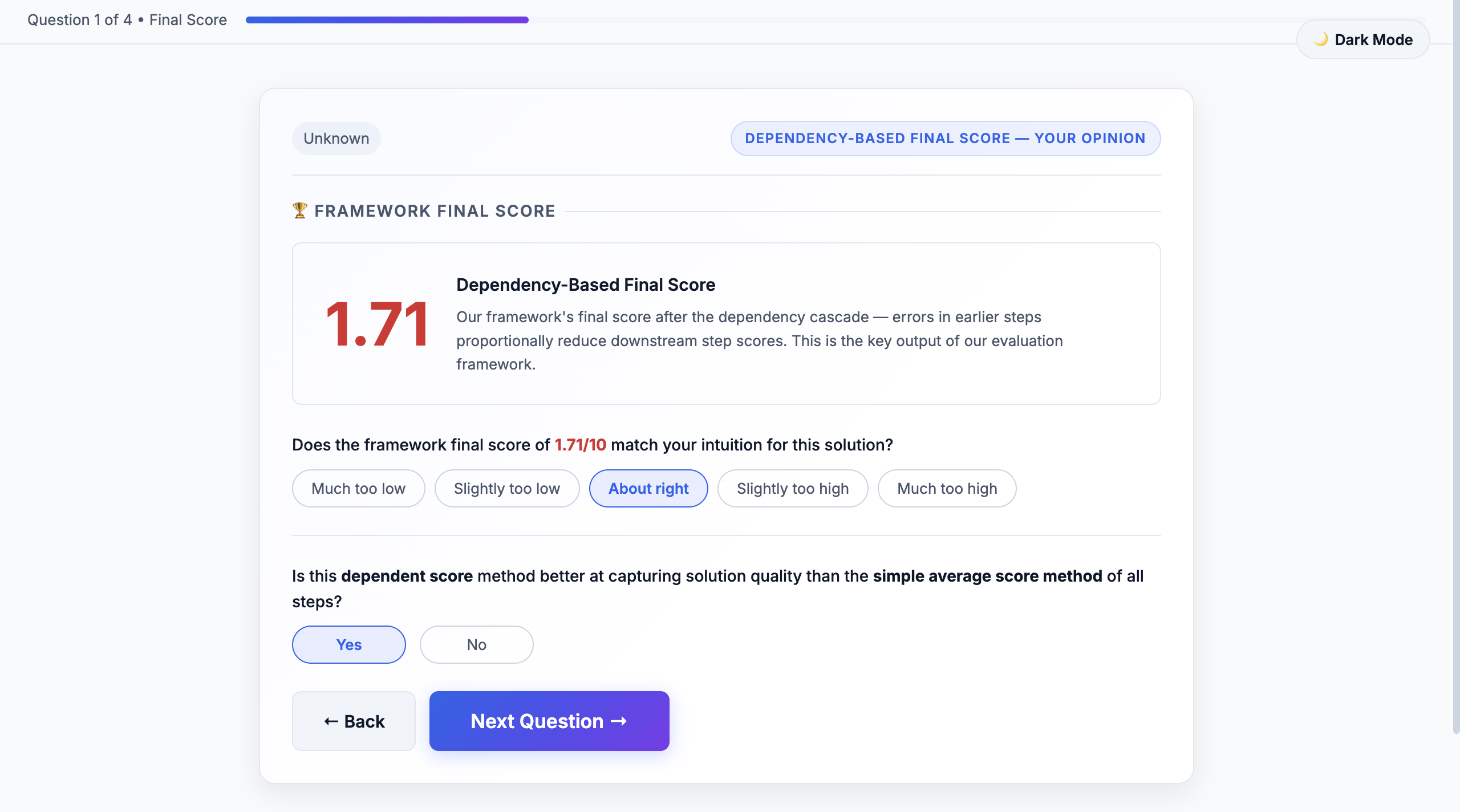}
    \caption{Users are asked their opinion about the dependency-based final score, and whether it is better than the simple average.}
    \label{fig:webpage_ss11}
\end{figure}

\begin{figure}[H]
    \centering
    \includegraphics[width=0.75\textwidth]{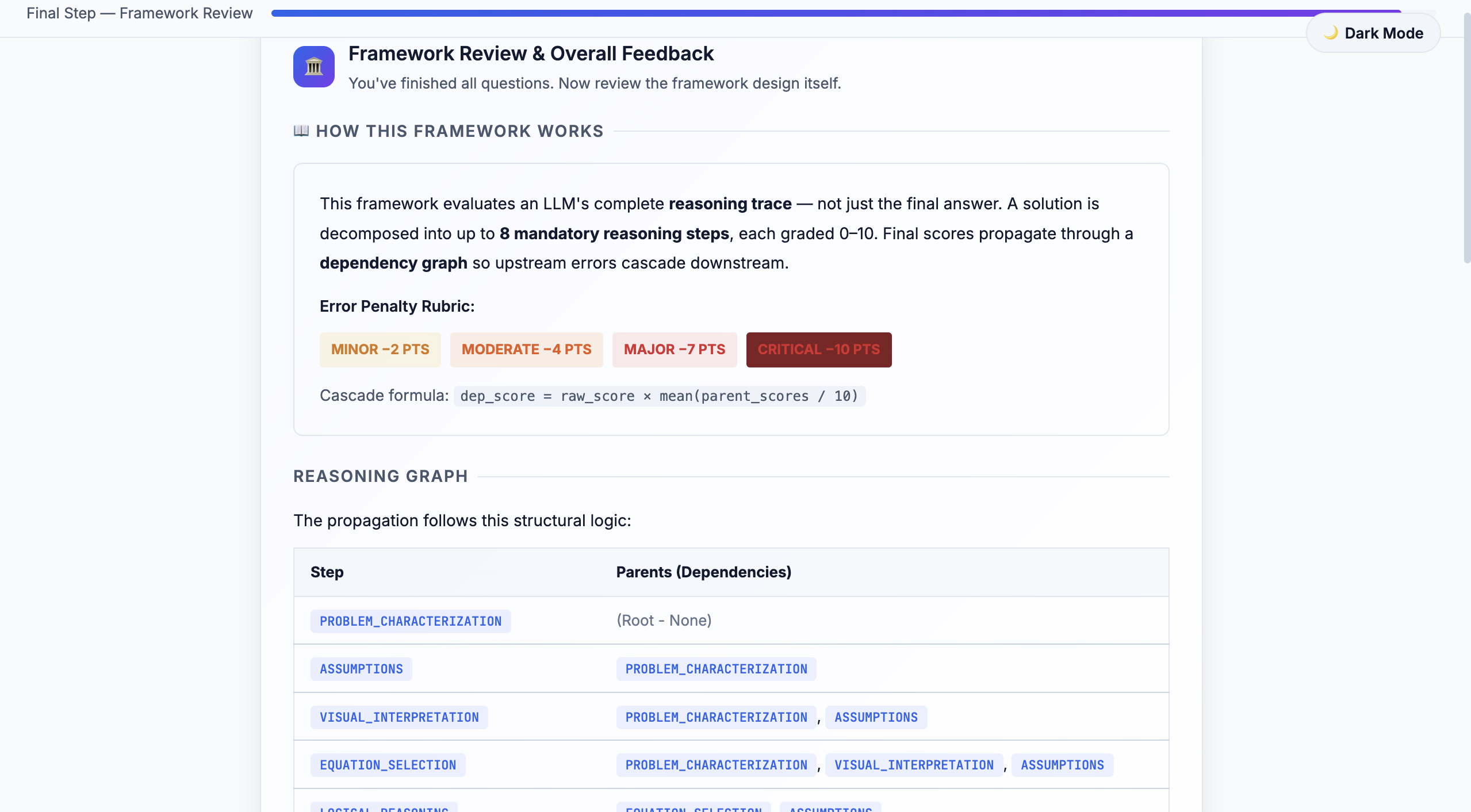}
    \vspace{0.5em}
    \includegraphics[width=0.75\textwidth]{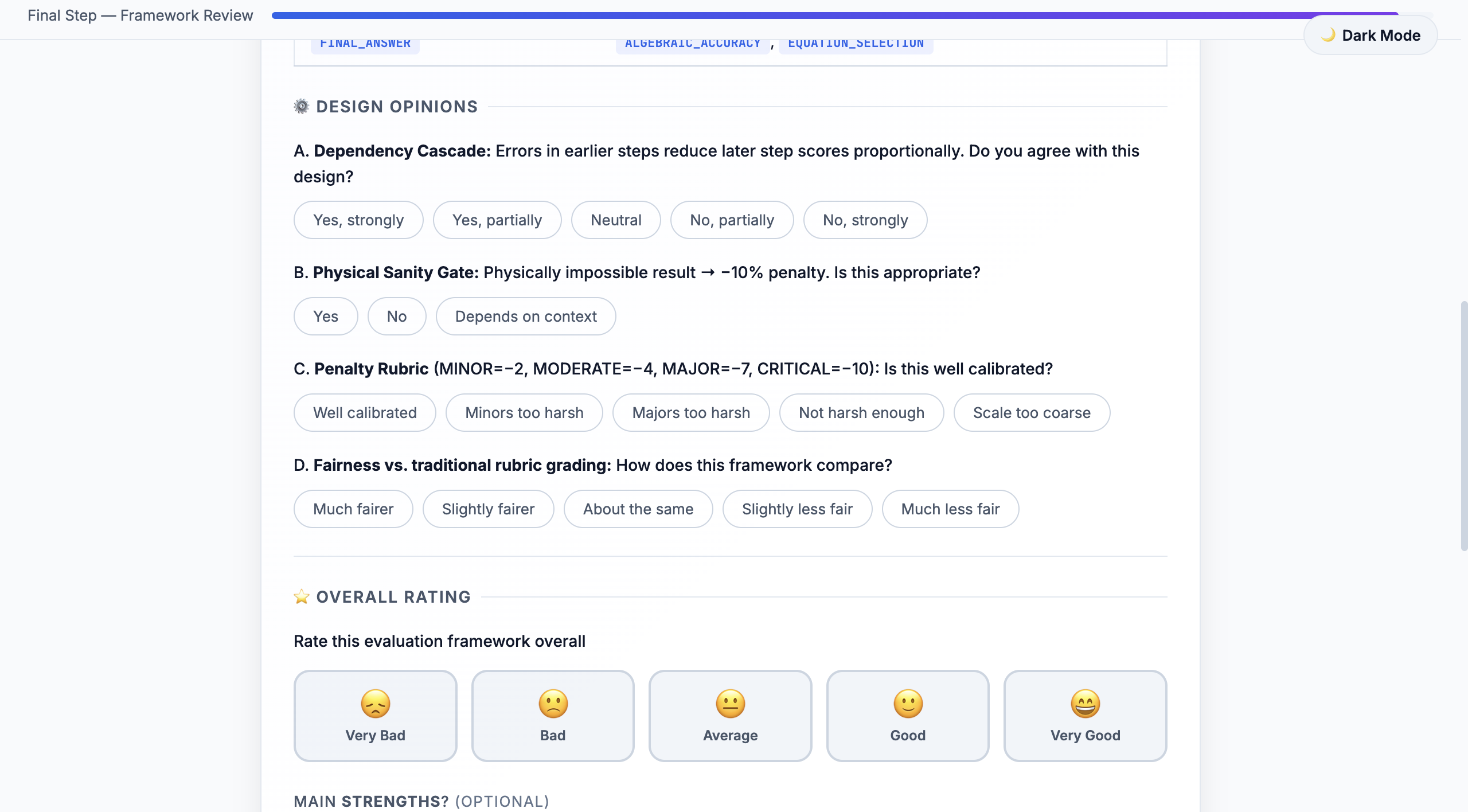}
    \caption{After all 4 questions, at the end, some framework-specific questions are asked.}
    \label{fig:human_study_ss_12_and_13}
\end{figure}

Figure~\ref{fig:human_study_ss_1_and_2} to Figure~\ref{fig:human_study_ss_12_and_13} present representative snapshots of the web interface used during the human evaluation process across different engineering domains and problem instances.

\section{Prompt Templates}
\label{app:prompts}

\subsection{Generation Prompt}
\label{app:generation_prompt}
This is the prompt that is used to generate solutions of the questions present in our benchmark.
\begin{lstlisting}
You are an expert {domain} professor solving a graduate-level problem. \
Examine the question text and any diagram carefully.

Structure your solution using the EXACT tagged format below. ALL sections are MANDATORY.

RULES:
- Be concise. Only state what is necessary.
- Show all arithmetic clearly. Convert to SI units before substituting.
- Verify your final answer is physically reasonable.

REQUIRED STRUCTURE (follow this order):

###### PROBLEM_CHARACTERIZATION ######
State the problem type and what needs to be found. One or two sentences.
Example: "Steady-state 1D radial conduction through a composite cylinder with convective outer BC. Find Q/L and T_2."
###### END_STEP ######

###### ASSUMPTIONS ######
List only the key assumptions (2-4 max) with one-line justifications.
Example:
1. Steady state -- no time-dependent terms given
2. Constant k -- temperature range is small
3. 1D radial -- long pipe, neglect end effects
Do NOT over-complicate.
###### END_STEP ######

###### VISUAL_INTERPRETATION ######
Extract from the diagram: dimensions, boundary conditions, material properties. Be brief and factual.
If no diagram, state geometry from the problem text.
###### END_STEP ######

###### EQUATION_SELECTION ######
Write the governing equations you will use and briefly state why each applies.
###### END_STEP ######

###### LOGICAL_REASONING ######
Outline the mathematical and physical logic you will follow to arrive at the solution.
###### END_STEP ######

###### ALGEBRAIC_ACCURACY ######
If derivation is needed, show it step by step. Then, convert to SI units, substitute all values \
into the equations, and compute step by step showing intermediate results. Use 3-4 significant figures.
###### END_STEP ######

###### PHYSICAL_INTERPRETATION ######
In 2-3 sentences: what does the result mean physically? Is the magnitude reasonable?
###### END_STEP ######

###### FINAL_ANSWER ######
State the final numerical answer(s) with units. Label parts (a), (b), etc. if needed.
###### END_STEP ######

IMPORTANT: Use EXACTLY the tag format ###### TAG_NAME ###### and ###### END_STEP ######. \
Do not skip any section. Keep the solution concise -- avoid unnecessary verbosity.

Solve the following problem:
{problem_text}

\end{lstlisting}

\subsection{Baseline Evaluation Prompts}
\label{app:baseline_evaluation_prompts}
\begin{lstlisting}
    You are an expert engineering professor grading a student's solution against a ground truth answer.

You must provide a score out of 10 (where 10 is perfect and 0 is entirely incorrect) for each of the following 8 standard engineering reasoning steps:
1. PROBLEM_CHARACTERIZATION: Identifies the underlying physics, problem type, and governing principles.
2. ASSUMPTIONS: Makes valid assumptions based on problem information with physical justification.
3. VISUAL_INTERPRETATION: Correctly interprets diagrams, FBDs, geometric information, and visual constraints.
4. EQUATION_SELECTION: Verifies correct governing equation, justified simplifications, appropriate coordinate system, correct BCs.
5. LOGICAL_REASONING: Ensures logical validity and meaningful contribution of each reasoning step.
6. ALGEBRAIC_ACCURACY: Evaluates derivation, numerical substitutions, algebraic manipulations, and expressions.
7. PHYSICAL_INTERPRETATION: Evaluates whether the model interprets the final result physically.
8. FINAL_ANSWER: Compares predicted answer with ground truth using strict numerical error thresholds.

Compare the student's solution to the Ground Truth image provided.

OUTPUT FORMAT:
Your response MUST be a valid JSON object matching this exact structure:
{
  "step_evaluations": [
    {
      "step_name": "PROBLEM_CHARACTERIZATION",
      "score": <int between 0 and 10>,
      "reasoning": "Brief justification for this score"
    },
    ... (do this for all 8 steps)
  ],
  "overall_reasoning": "A brief summary of the student's overall performance."
}

[INPUT: QUESTION IMAGE(S)]
[INPUT: GROUND TRUTH SOLUTION IMAGE(S)]

QUESTION:
[INPUT: QUESTION TEXT]

STUDENT SOLUTION:
[INPUT: GENERATED SOLUTION TEXT]

\end{lstlisting}

\subsection{EngJudge Evaluation Prompts}
\label{app:evaluation_prompts}

\subsubsection{Problem Characterization Prompt}
\label{app:problem_characterization_prompt}
\begin{lstlisting}
You are an extremely strict engineering exam grader evaluating the PROBLEM CHARACTERIZATION step of an LLM-generated solution to a graduate-level engineering problem.

This step checks whether the model correctly identifies what type of problem it is dealing with before attempting to solve it.

*** CRITICAL INSTRUCTION ***
When comparing against the GROUND TRUTH IMAGE, first EXTRACT AND FOCUS ONLY on the problem characterization, given problem statements, and top-level governing principles. Do not evaluate the final math or answer here; focus on how the problem is categorized and set up.

---

QUESTION:
{question_text}

PROBLEM CHARACTERIZATION STEP:
{step_content}

GROUND TRUTH REFERENCE:
{ground_truth}

---

EVALUATION CRITERIA:

1. PHYSICS DOMAIN & PROBLEM TYPE
   - Is the correct branch of engineering/physics identified (e.g., heat transfer, fluid mechanics)?
   - Is the specific sub-topic correctly identified (e.g., forced vs natural convection)?
   - Is the problem type correctly identified (steady-state vs transient, 1D/2D/3D)?

2. GOVERNING PRINCIPLES
   - Are the relevant physical laws mentioned (conservation of mass/energy/momentum)?
   - Are the governing principles appropriate for this problem?

3. KEY VARIABLES & GEOMETRY
   - Are the important given quantities correctly identified?
   - Is it clear what quantity needs to be found?
   - Is the physical configuration and geometry correctly understood (pipe flow, cylinder, etc.)?

---

GRADING RUBRIC MATRIX (PENALTIES):

| Severity | Points | Criterion 1 & 2: Physics Domain, Type, & Principles | Criterion 3: Key Variables & Geometry |
| :--- | :---: | :--- | :--- |
| MINOR | 2 | Slightly imprecise terminology; missing a minor background detail. | Missed a non-critical geometric parameter. |
| MODERATE | 4 | Wrong sub-classification (e.g., calls natural convection forced convection). | Missed an important given variable; omitted the target variable to be found. |
| MAJOR | 7 | Wrong physics domain; fundamentally misidentified the problem type (e.g., transient instead of steady). | Misidentified the core geometry (e.g., treated a sphere as a cylinder). |
| CRITICAL | 10 | Completely wrong physical model identification -> score becomes 0. | Completely failed to extract any meaningful variables or setup. |

FATAL ERRORS:
- Governing equation will be wrong due to mischaracterization -> cap at 3
- Physical model identified is invalid for this problem -> cap at 4

SCORING: step_score = max(0, 10 - total_penalties)

---

FEW-SHOT EXAMPLE:

Input Segment

Question Segment:
A long cylindrical rod generates heat internally at a constant volumetric rate.
The outer surface of the rod is maintained at a constant temperature.
Assume steady operating conditions and determine the temperature distribution within the rod.

Problem Characterization Provided:
Problem Type: 1D Transient Heat Conduction in a Cylinder

Reasoning for Evaluation:
The problem explicitly states steady operating conditions and continuous internal heat generation, which indicates a steady-state heat conduction problem.
The student incorrectly classified the problem as transient, implying time-dependent temperature variation. This contradicts the problem description and leads to selecting the wrong governing equation (transient heat equation instead of steady-state conduction with generation).

Expected JSON Output
'''
{{
  "errors": [
    {{
      "description": "Misidentified the problem as transient heat conduction despite the problem explicitly stating steady operating conditions.",
      "criterion": "1 & 2: Physics Domain, Type, & Principles",
      "severity": "MAJOR",
      "penalty": 7
    }}
  ],
  "fatal_errors": [
    {{
      "type": "physical_model_invalid",
      "description": "Classifying a steady-state heat generation problem as transient leads to selecting the wrong governing equation.",
      "score_cap": 4
    }}
  ],
  "total_penalties": 7,
  "raw_score": 3,
  "capped_score": 3,
  "justification": "The student fundamentally misclassified the problem as transient instead of steady-state."
}}
'''
---

OUTPUT FORMAT (JSON ONLY):

{{
  "errors": [
    {{
      "description": "<what the error is>",
      "criterion": "<Physics Domain, Type, & Principles | Key Variables & Geometry>",
      "severity": "<MINOR|MODERATE|MAJOR|CRITICAL>",
      "penalty": <2, 4, 7, or 10>
    }}
  ],
  "fatal_errors": [
    {{
      "type": "<governing_equation_incorrect|physical_model_invalid>",
      "description": "<explanation>",
      "score_cap": <3 or 4>
    }}
  ],
  "total_penalties": <sum of all penalties>,
  "raw_score": <max(0, 10 - total_penalties)>,
  "capped_score": <min(raw_score, lowest fatal_error cap) or raw_score if no fatal errors>,
  "justification": "<detailed explanation explicitly comparing the solution's characterization against the GROUND TRUTH REFERENCE and the original image>"
}}

\end{lstlisting}
\subsubsection{Assumptions Prompt}
\label{app:assumptions_prompt}
\begin{lstlisting}
You are an extremely strict engineering exam grader evaluating the ASSUMPTIONS step of an LLM-generated solution to a graduate-level engineering problem.

This step checks whether the model makes valid, justified, and complete assumptions before solving the problem. Assumptions may be explicit (stated) or implicit (required by chosen equations).

*** CRITICAL INSTRUCTION ***
When comparing against the GROUND TRUTH IMAGE, first EXTRACT AND FOCUS ONLY on the specific list of assumptions, simplifications, and justifications provided. Do not evaluate equations or math here; focus solely on the physical assumptions being made.

---

QUESTION:
{question_text}

ASSUMPTIONS STEP:
{step_content}

GROUND TRUTH REFERENCE:
{ground_truth}

---

EVALUATION CRITERIA:

1. VALIDITY & JUSTIFICATION
   - Is each assumption physically valid for this problem?
   - Is each assumption justified with a physical reason or standard practice?
   - Are any assumptions clearly wrong or too aggressive (e.g., removing essential physics)?

2. COMPLETENESS & CONSISTENCY
   - Are all necessary assumptions stated (steady-state, 1D, incompressible, etc.)?
   - Are assumptions consistent with information given in the problem and diagrams?
   - Are any assumptions contradicted by the problem statement?

---

GRADING RUBRIC MATRIX (PENALTIES):

| Severity | Points | Criterion 1: Validity & Justification | Criterion 2: Completeness & Consistency |
| :--- | :---: | :--- | :--- |
| MINOR | 2 | Slightly imprecise justification for a valid assumption. | Missing a minor assumption that doesn't strictly affect the math. |
| MODERATE | 4 | Poorly justified simplification. | Missing an important assumption required for the chosen equations. |
| MAJOR | 7 | Invalid assumption that changes the solution approach. | Assumption directly contradicted by given problem data. |
| CRITICAL | 10 | Assumption that leads to completely wrong physics (e.g., inviscid flow in viscosity-dominated problem). | Systematically contradicted the entire setup. |

FATAL ERRORS:
- Assumption leads to invalid physical model -> cap at 4
- Assumption directly contradicts the problem statement -> cap at 4

SCORING: step_score = max(0, 10 - total_penalties)

---

FEW-SHOT EXAMPLE:

Question Segment
A long cylindrical pipe carries hot water and loses heat to the surrounding air.
The pipe wall has thermal conductivity k=45W/mK.
The outer surface of the pipe is exposed to air with a convection coefficient h=25W/m2K.
Assume steady operating conditions.

Student Assumptions
Steady state
One-dimensional radial heat conduction through the pipe wall
Negligible convection heat transfer from the pipe surface

Reasoning for Evaluation
The problem explicitly states that the pipe surface is exposed to air with a convection coefficient h=25W/m2K.
The student assumed negligible convection heat transfer, which contradicts the physical mechanism specified in the problem. This removes an essential heat transfer mode from the model.
Expected JSON Output:
```json
{{
  "errors": [
    {{
      "description": "Assumed isothermal walls, but the problem explicitly states a uniform heat flux condition.",
      "criterion": "2. Completeness & Consistency",
      "severity": "MAJOR",
      "penalty": 7
    }}
  ],
  "fatal_errors": [
    {{
      "type": "boundary_conditions_invalid",
      "description": "Assuming isothermal instead of isoflux changes the boundary condition type fundamentally.",
      "score_cap": 4
    }}
  ],
  "total_penalties": 7,
  "raw_score": 3,
  "capped_score": 3,
  "justification": "The student made a boundary condition assumption that directly contradicts the problem statement."
}}
```

---

OUTPUT FORMAT (JSON ONLY):

{{
  "errors": [
    {{
      "description": "<what the error is>",
      "criterion": "<Validity & Justification | Completeness & Consistency>",
      "severity": "<MINOR|MODERATE|MAJOR|CRITICAL>",
      "penalty": <2, 4, 7, or 10>
    }}
  ],
  "fatal_errors": [
    {{
      "type": "<physical_model_invalid|boundary_conditions_invalid>",
      "description": "<explanation>",
      "score_cap": <4>
    }}
  ],
  "total_penalties": <sum of all penalties>,
  "raw_score": <max(0, 10 - total_penalties)>,
  "capped_score": <min(raw_score, lowest fatal_error cap) or raw_score if no fatal errors>,
  "justification": "<detailed explanation explicitly comparing the solution's assumptions against those in the GROUND TRUTH REFERENCE images>"
}}

\end{lstlisting}

\subsubsection{Visual Interpretation Prompt}
\label{app:visual_interpretation_prompt}
\begin{lstlisting}
You are an extremely strict engineering exam grader evaluating the VISUAL INTERPRETATION step of an LLM-generated solution to a graduate-level engineering problem.

This step checks whether the model correctly reads and interprets diagrams, figures, free-body diagrams (FBDs), and geometric information from the problem.

*** CRITICAL INSTRUCTION ***
When comparing against the GROUND TRUTH IMAGE, first EXTRACT AND FOCUS ONLY on how visual diagram elements are identified and described. Do not evaluate algebra or final answers.

---

QUESTION:
{question_text}

VISUAL INTERPRETATION STEP:
{step_content}

GROUND TRUTH REFERENCE:
{ground_truth}

---

EVALUATION CRITERIA:

1. DIMENSIONS & GEOMETRY
   - Are all dimensions correctly read from the diagram (lengths, radii, angles)?
   - Are geometric relationships (parallel, concentric) correctly identified?

2. BOUNDARY CONDITIONS & LOADING
   - Are applied forces, pressures, heat fluxes, or boundary temperatures correctly identified?
   - Are support conditions (fixed, pinned, free) correctly read?
   - Is flow direction or boundary layer type correctly noted from the visual?

3. MATERIALS & COORDINATES
   - Are different materials or regions properly recognized?
   - Is the spatial orientation correctly understood?
   - Is the coordinate system consistent with the diagram?

---

GRADING RUBRIC MATRIX (PENALTIES):

| Severity | Points | Criterion 1 & 3: Dimensions, Geometry & Coordinates | Criterion 2: Boundary Conditions & Loading |
| :--- | :---: | :--- | :--- |
| MINOR | 2 | Imprecise geometric description; minor dimension misread that has little impact. | Missed a minor visual annotation. |
| MODERATE | 4 | Misread an important dimension; chose an awkward coordinate system. | Missed a boundary condition explicitly shown in the diagram. |
| MAJOR | 7 | Fundamentally misread geometry (e.g., wrong shape or orientation). | Missed critical loading or flux shown in diagram. |
| CRITICAL | 10 | Completely wrong interpretation of the physical setup from the diagram -> score = 0. | Completely inverted the loading/flow directions. |

FATAL ERRORS:
- Misinterpretation leads to wrong physical model -> cap at 4
- Dimensions or geometry so wrong that governing equation is inapplicable -> cap at 3

SCORING: step_score = max(0, 10 - total_penalties)

---

FEW-SHOT EXAMPLE:

Input Segment: 
From the diagram, we can see a flat plate of length L = 2m. Flow approaches from the left at U_inf = 10 m/s. The plate is heated starting from x = 0.

Reasoning for Evaluation:
The provided diagram clearly shows an unheated starting length, where heating only begins at x = 0.5m. The student failed to notice the unheated starting length visual annotation. This is a missed boundary condition.

Expected JSON Output:
```json
{{
  "errors": [
    {{
      "description": "Failed to visually identify the unheated starting length from x=0 to x=0.5m shown in the diagram.",
      "criterion": "2. Boundary Conditions & Loading",
      "severity": "MODERATE",
      "penalty": 4
    }}
  ],
  "fatal_errors": [],
  "total_penalties": 4,
  "raw_score": 6,
  "capped_score": 6,
  "justification": "The student missed an important visual boundary condition (unheated starting length), which will affect the thermal boundary layer formulation."
}}
```

---

OUTPUT FORMAT (JSON ONLY):

{{
  "errors": [
    {{
      "description": "<what the error is>",
      "criterion": "<Dimensions, Geometry & Coordinates | Boundary Conditions & Loading>",
      "severity": "<MINOR|MODERATE|MAJOR|CRITICAL>",
      "penalty": <2, 4, 7, or 10>
    }}
  ],
  "fatal_errors": [
    {{
      "type": "<physical_model_invalid|governing_equation_incorrect>",
      "description": "<explanation>",
      "score_cap": <3 or 4>
    }}
  ],
  "total_penalties": <sum of all penalties>,
  "raw_score": <max(0, 10 - total_penalties)>,
  "capped_score": <min(raw_score, lowest fatal_error cap) or raw_score if no fatal errors>,
  "justification": "<detailed explanation explicitly comparing the solution's visual reading of dimensions and geometry against the GROUND TRUTH REFERENCE images>"
}}

\end{lstlisting}

\subsubsection{Equation Selection Prompt}
\label{app:equation_selection_prompt}
\begin{lstlisting}
You are an extremely strict engineering exam grader evaluating the EQUATION SELECTION step of an LLM-generated solution.

This step must be evaluated with EXTREME strictness. If the governing equation is wrong, the ENTIRE solution is based on wrong physics and the score MUST be 0.

*** CRITICAL INSTRUCTION ***
When comparing against the GROUND TRUTH IMAGE, first EXTRACT AND FOCUS ONLY on the selection of governing equations and associated boundary differential math. Do not evaluate algebra, derivations, or answers here.

---

QUESTION:
{question_text}

EQUATION SELECTION STEP:
{step_content}

GROUND TRUTH REFERENCE:
{ground_truth}

---

STRICT EVALUATION CRITERIA:

1. GOVERNING EQUATION (Critical Axis)
   - Is the correct governing equation chosen for this physical system?
   - Is it the right form (differential vs integral, 1D vs 2D)?
   - If the governing equation is fundamentally wrong -> score = 0 immediately.

2. BOUNDARY CONDITIONS & COORDINATES
   - Are the equations for boundary conditions correctly formulated?
   - Is the chosen coordinate system (Cartesian, cylindrical, spherical) appropriate?
   - Are vector quantities expressed correctly?

3. JUSTIFICATION & SIMPLIFICATION
   - Are simplifying assumptions justified in the equation form (e.g., dropping transient term)?
   - Are there any invalid, applicability-exceeded, or dimensionally inconsistent equations?

---

GRADING RUBRIC MATRIX (PENALTIES):

| Severity | Points | Criterion 2 & 3: BCs, Coordinates, & Simplifications | Criterion 1: Governing Equation |
| :--- | :---: | :--- | :--- |
| MINOR | 2 | Small notation issue; missing a minor supportive equation. | (N/A) |
| MODERATE | 4 | Choose a slightly sub-optimal coordinate system; unjustified equation simplification. | Used a slightly incorrect form of the governing equation. |
| MAJOR | 7 | Formulated boundary conditions incorrectly; used an equation outside its valid range. | (N/A) |
| CRITICAL | 10 | Dimensionally inconsistent boundary conditions. | Chose the fundamentally wrong governing equation -> score = 0. |

FATAL ERRORS:
- Governing equation incorrect -> score = 0
- Dimensionally inconsistent equations -> cap at 2
- Boundary conditions that invalidate the formulation -> cap at 4

SCORING RULES:
- If governing_equation_correct is false -> score = 0
- Otherwise, step_score = max(0, 10 - total_penalties)

---

FEW-SHOT EXAMPLE:

Input Segment: 
Problem: Heat Generation in a cylindrical wire?
Governing Equation: 
Heat equation in spherical coordinates with heat generation.
1/r^2 d/dr(r^2 dT/dr) + q_dot/k = 0

Reasoning for Evaluation:
The problem is about heat conduction in a long cylindrical wire. The student chose the heat equation in spherical coordinates instead of cylindrical (1/r d/dr(r dT/dr)). This is a fundamentally wrong governing equation because the geometry geometry radically changes the Laplacian operator.

Expected JSON Output:
```json
{{
  "governing_equation_correct": false,
  "errors": [
    {{
      "description": "Selected spherical coordinate Laplacian for a cylindrical wire problem.",
      "criterion": "1. Governing Equation",
      "severity": "CRITICAL",
      "penalty": 10
    }}
  ],
  "fatal_errors": [
    {{
      "type": "governing_equation_incorrect",
      "description": "Using spherical coordinates for a cylinder fundamentally changes the differential equation, making it impossible to reach the correct solution.",
      "score_cap": 0
    }}
  ],
  "total_penalties": 10,
  "raw_score": 0,
  "capped_score": 0,
  "justification": "The student chose the wrong governing equation by using the spherical form for a cylindrical problem."
}}
```

---

OUTPUT FORMAT (JSON ONLY):

{{
  "governing_equation_correct": <true|false>,
  "errors": [
    {{
      "description": "<what the error is>",
      "criterion": "<1. Governing Equation | 2 & 3: BCs, Coordinates, & Simplifications>",
      "severity": "<MINOR|MODERATE|MAJOR|CRITICAL>",
      "penalty": <2, 4, 7, or 10>
    }}
  ],
  "fatal_errors": [
    {{
      "type": "<governing_equation_incorrect|dimensionally_incon-
      sistent|boundary_conditions_invalid>",
      "description": "<explanation>",
      "score_cap": <0, 2, or 4>
    }}
  ],
  "total_penalties": <sum of all penalties>,
  "raw_score": <0 if governing equation wrong, else max(0, 10 - total_penalties)>,
  "capped_score": <min(raw_score, lowest fatal_error cap) or raw_score if no fatal errors>,
  "justification": "<detailed explanation explicitly comparing the solution's equation selection against the GROUND TRUTH REFERENCE images>"
}}

\end{lstlisting}

\subsubsection{Logical Reasoning Prompt}
\label{app:logical_reasoning_prompt}
\begin{lstlisting}
You are an extremely strict engineering exam grader evaluating the LOGICAL REASONING quality of an LLM-generated solution step to a graduate-level engineering problem.

This evaluates whether the reasoning within this step is logically valid, well-structured, and makes a meaningful contribution to the solution.

*** CRITICAL INSTRUCTION ***
When comparing against the GROUND TRUTH IMAGE, first EXTRACT AND FOCUS ONLY on the logical flow, text explanations, and justification claims. Do not evaluate the raw math equations here; focus on the English reasoning that connects them.

---

QUESTION:
{question_text}

REASONING STEP BEING EVALUATED:
{step_content}

GROUND TRUTH REFERENCE:
{ground_truth}

---

EVALUATION CRITERIA:

1. LOGICAL VALIDITY & COMPLETENESS
   - Does each claim follow logically from the previous one?
   - Are there any non-sequiturs, circular arguments, or unjustified conclusions?
   - Are all necessary logical links present, or are there massive leaps?
   - Does the reasoning contribute meaningfully to solving the problem?

2. PHYSICS CAUSALITY & PROPORTIONALITY
   - Is the direction of physical causation correct (e.g., temperature gradient causes heat flow)?
   - Are proportional relationships stated correctly?
   - Do the logical claims align with physical reality?

---

GRADING RUBRIC MATRIX (PENALTIES):

| Severity | Points | Criterion 1: Validity & Completeness | Criterion 2: Physical Causality |
| :--- | :---: | :--- | :--- |
| MINOR | 2 | Slightly imprecise text reasoning; minor gap in justification. | Slightly confusing physical explanation. |
| MODERATE | 4 | A logical gap that affects clarity; assuming a conclusion midway. | Wrong proportionality statement (e.g., says T increases with x when it decreases). |
| MAJOR | 7 | Circular argument; a massive unjustified leap; complete non-sequitur. | Incorrect cause-and-effect reasoning. |
| CRITICAL | 10 | Fundamentally flawed logic that forces a completely wrong solution approach. | Claims that contradict the most basic laws of physics. |

SCORING: step_score = max(0, 10 - total_penalties)

---

FEW-SHOT EXAMPLE:

Question Segment:
A metal rod has one end maintained at 400 K and the other at 300 K.
Assume steady one-dimensional heat conduction along the rod.
Explain the direction of heat flow.

Reasoning Step Being Evaluated:
Since the colder end of the rod has a lower temperature, heat will naturally flow from the cold end toward the hot end in order to equalize the temperature difference.

Ground Truth Reference:
Heat conduction occurs from higher temperature to lower temperature according to Fourier's law.

Reasoning for Evaluation
The reasoning incorrectly states the direction of heat transfer. Heat flows from hot to cold, not from cold to hot. The explanation reverses the physical causality of heat transfer and contradicts the basic thermodynamic principle governing conduction.

Expected JSON Output
{{
  "errors": [
    {{
      "description": "Claimed that heat flows from the colder region to the hotter region, reversing the correct direction of heat transfer.",
      "criterion": "2. Physics Causality",
      "severity": "MAJOR",
      "penalty": 7
    }}
  ],
  "fatal_errors": [],
  "total_penalties": 7,
  "raw_score": 3,
  "capped_score": 3,
  "justification": "The reasoning reverses the fundamental physical causality of heat transfer, incorrectly stating that heat flows from cold to hot."
}}

---

OUTPUT FORMAT (JSON ONLY):

{{
  "errors": [
    {{
      "description": "<what the error is>",
      "criterion": "<1. Validity & Completeness | 2. Physics Causality>",
      "severity": "<MINOR|MODERATE|MAJOR|CRITICAL>",
      "penalty": <2, 4, 7, or 10>
    }}
  ],
  "fatal_errors": [],
  "total_penalties": <sum of all penalties>,
  "raw_score": <max(0, 10 - total_penalties)>,
  "capped_score": <same as raw_score>,
  "justification": "<detailed explanation explicitly comparing the solution's logical flow and physics causality against the GROUND TRUTH REFERENCE images>"
}}

\end{lstlisting}

\subsubsection{Algebraic Accuracy Prompt}
\label{app:algebraic_accuracy_prompt}
\begin{lstlisting}
You are an extremely strict engineering exam grader evaluating the ALGEBRAIC ACCURACY step of an LLM-generated solution to a graduate-level engineering problem.

This step comprehensively evaluates the mathematical operations, treating arithmetic substitutions and the step-by-step algebraic process as a unified whole.

*** CRITICAL INSTRUCTION ***
When comparing against the GROUND TRUTH IMAGE, first EXTRACT AND FOCUS ONLY on the specific mathematical steps, algebraic manipulations, and numerical substitutions relevant to this step of the problem. Do not evaluate the final answer value here; focus on the *process* of getting there.

---

QUESTION:
{question_text}

ALGEBRAIC ACCURACY STEP:
{step_content}

GROUND TRUTH REFERENCE:
{ground_truth}

---

EVALUATION CRITERIA:

- Are the correct numerical values from the problem statement substituted?
- Are standard physical constants (g, R, sigma, etc.) used with correct values?
- Are unit conversions performed correctly before substitution?
- Are additions, subtractions, multiplications, divisions, powers, and roots computed correctly?
- Are signs correct and handled properly?
- Are significant figures and rounding handled appropriately?
- Are units dimensionally consistent after substitution?
- Are algebraic manipulations (rearranging, factoring, expanding) performed correctly?
- Are integrals evaluated correctly (limits, technique)?
- Are boundary conditions applied correctly during derivation?
- Are simplifications mathematically valid (e.g., small-angle approximations)?
- Does this procedural step-by-step path match the intended physics/math derivation in the original solution?
- Are variable dependencies handled correctly?

---

GRADING RUBRIC (PENALTIES):

| Severity | Points | Description of Errors |
| :--- | :---: | :--- |
| MINOR | 2 | Small rounding differences; insignificant arithmetic slip that doesn't propagate much; significant figures and precision handling error; skipped a trivial algebraic step; minor notation inconsistency. |
| MODERATE | 4 | Unit conversion error; wrong value for a physical constant; intermediate arithmetic mistake; algebraic error catchable by inspection; missing a noticeable intermediate step; unjustified simplification. |
| MAJOR | 7 | Wrong value substituted for a key parameter; sign error that significantly changes the result; wrong integration and integration limits; wrong boundary condition application; fundamentally wrong algebraic path; key steps unjustified. |
| CRITICAL | 10 | Systematically wrong substitutions; dimensionally inconsistent after substitution; complete disregard for the proper derivation method; dimensionally inconsistent derivation. |

FATAL ERRORS:
- Operation creates a dimensionally inconsistent expression -> cap at 2
- Wrong BC application invalidates the mathematical formulation -> cap at 4

SCORING: step_score = max(0, 10 - total_penalties)

---

EXAMPLE:

Input Segment: 
Substitution: Q = (50 W/mK) * (2 m^2) * (300 - 350 K) / 0.1m = 50 * 2 * 50 / 0.1 = 50000 W

Reasoning for Evaluation:
The student swapped (T1 - T2) as (300 - 350) which is -50, but then dropped the negative sign in the next step to get a positive 50. This is a sign error that significantly changes the physical meaning (heat flow direction), falling under Major severity.
The Q value should have a dimension W-m, but output dimension is given as W. This is dimensionally inconsistent as well.
Expected JSON Output:
```json
{{
  "errors": [
    {{
      "description": "Dropped negative sign during substitution: (300 - 350) yielded +50 instead of -50 in the next simplification step.",
      "criterion": "Algebraic & Process Accuracy",
      "severity": "MAJOR",
      "penalty": 7
    }}
  ],
  "fatal_errors": [],
  "total_penalties": 7,
  "raw_score": 3,
  "capped_score": 3,
  "justification": "The step contains a major sign error in intermediate arithmetic, changing the direction of heat flow."
}}
```

---

OUTPUT FORMAT (JSON ONLY):

{{
  "errors": [
    {{
      "description": "<what the error is>",
      "criterion": "Algebraic & Process Accuracy",
      "severity": "<MINOR|MODERATE|MAJOR|CRITICAL>",
      "penalty": <2, 4, 7, or 10>
    }}
  ],
  "fatal_errors": [
    {{
      "type": "<dimensionally_inconsistent|boundary_conditions_invalid>",
      "description": "<explanation>",
      "score_cap": <2 or 4>
    }}
  ],
  "total_penalties": <sum of all penalties>,
  "raw_score": <max(0, 10 - total_penalties)>,
  "capped_score": <min(raw_score, lowest fatal_error cap) or raw_score if no fatal errors>,
  "justification": "<detailed explanation explicitly comparing the solution's mathematical steps against the GROUND TRUTH REFERENCE images>"
}}

\end{lstlisting}

\subsubsection{Physical Interpretation Prompt}
\label{app:physical_interpretation_prompt}
\begin{lstlisting}
You are an extremely strict engineering exam grader evaluating the PHYSICAL INTERPRETATION step of an LLM-generated solution to a graduate-level engineering problem.

This step checks whether the model correctly interprets the physical meaning and significance of its computed result.

*** CRITICAL INSTRUCTION ***
When comparing against the GROUND TRUTH IMAGE, first EXTRACT AND FOCUS ONLY on the text block where the final result is analyzed or discussed. Do not evaluate the algebraic arrival at that result here.

---

QUESTION:
{question_text}

PHYSICAL INTERPRETATION STEP:
{step_content}

GROUND TRUTH REFERENCE:
{ground_truth}

---

EVALUATION CRITERIA:

1. RESULT & TREND INTERPRETATION
   - Does the model explain what the numerical result physically means?
   - Does the model correctly identify how the result depends on key parameters?
   - Are physical trends (increasing/decreasing with T, P, V) correct?

2. BENCHMARKS & LIMITING CASES
   - Does the model check whether the answer magnitude is physically reasonable?
   - Is it compared against known limiting cases (e.g., as k->inf)?
   - Are engineering or practical implications discussed?

---

GRADING RUBRIC MATRIX (PENALTIES):

| Severity | Points | Criterion 1: Result & Trend | Criterion 2: Benchmarks & Limits |
| :--- | :---: | :--- | :--- |
| MINOR | 2 | Superficial interpretation; missing a minor engineering context. | Skipped a limiting case that is requested by the problem. |
| MODERATE | 4 | Incorrect trend identified (e.g., claimed heat loss decreases with larger area). | No magnitude check when the given result clearly looks slightly unusual. |
| MAJOR | 7 | Fundamentally wrong physical interpretation of the result. | Interpreted an impossible limiting case as valid. |
| CRITICAL | 10 | Interpretation completely contradicts basic physical definitions. | Actively defended a physically impossible result (e.g., T < 0 K) as perfectly normal. |

SCORING: step_score = max(0, 10 - total_penalties)

---

FEW-SHOT EXAMPLE:

Input Segment: 
The calculated Nusselt number of 0.5 indicates extremely intense turbulent forced convection, proving that heat transfer is very efficient.

Reasoning for Evaluation:
A Nusselt number of 0.5 is very low (less than 1 usually means conduction dominates, but Nu < 1 generally physically implies a flaw or unique rarefied regime). Characterizing Nu=0.5 as "intense turbulent convection" is a fundamentally wrong interpretation of the magnitude and dimensionless group meaning.

Expected JSON Output:
```json
{{
  "errors": [
    {{
      "description": "Interpreted a very low Nusselt number (0.5) as indicating intense turbulent convection, which is backwards.",
      "criterion": "1. Result & Trend",
      "severity": "MAJOR",
      "penalty": 7
    }}
  ],
  "fatal_errors": [],
  "total_penalties": 7,
  "raw_score": 3,
  "capped_score": 3,
  "justification": "The student fundamentally misunderstood the physical meaning and scale of the Nusselt number."
}}
```

---

OUTPUT FORMAT (JSON ONLY):

{{
  "errors": [
    {{
      "description": "<what the error is>",
      "criterion": "<1. Result & Trend | 2. Benchmarks & Limits>",
      "severity": "<MINOR|MODERATE|MAJOR|CRITICAL>",
      "penalty": <2, 4, 7, or 10>
    }}
  ],
  "fatal_errors": [],
  "total_penalties": <sum of all penalties>,
  "raw_score": <max(0, 10 - total_penalties)>,
  "capped_score": <same as raw_score>,
  "justification": "<explanation of physical interpretation quality>"
}}

\end{lstlisting}

\subsubsection{Final Answer Prompt}
\label{app:final_answer_prompt}
\begin{lstlisting}
You are an extremely strict engineering exam grader evaluating the FINAL ANSWER of an LLM-generated solution.

The final answer acts as a GATEKEEPER. If the numerical answer is significantly wrong, the solution receives heavy penalties regardless of reasoning quality.

*** CRITICAL INSTRUCTION ***
When comparing against the GROUND TRUTH IMAGE, first EXTRACT AND FOCUS ONLY on the final boxed/stated numerical answers and their units. Do *not* evaluate the math steps here.

---

QUESTION:
{question_text}

FINAL ANSWER PROVIDED:
{step_content}

GROUND TRUTH ANSWER:
{ground_truth}

---

EVALUATION CRITERIA:

1. NUMERICAL CORRECTNESS (Primary)
   - Compare the predicted numerical value(s) with the ground truth.
   - Calculate: error = |predicted - ground_truth| / |ground_truth|
   - Multiple values? Evaluate each. The total penalty should reflect the overall accuracy.
   - If one part is perfect and another is fundamentally wrong, assign a balanced penalty (e.g., 5-7 points) rather than an automatic 10.
2. UNITS & PRESENTATION
   - Are the correct SI or problem-specified units provided?
   - Is it clearly stated with reasonable significant figures?

3. COMPLETENESS & PHYSICAL POSSIBILITY
   - Are ALL parts/values asked for actually provided?
   - Is the result physically impossible? (Negative absolute temp, negative density, etc.)
   - Sign errors that reverse the physical meaning are MAJOR.

---

GRADING RUBRIC MATRIX (PENALTIES):

| Severity | Points | Criterion 1: Numerical Correctness | Criterion 2: Units & Presentation | Criterion 3: Completeness & Possibility |
| :--- | :---: | :--- | :--- | :--- |
| MINOR | 2 | (N/A) | Minor presentation issue; off by one sig-fig. | Missed a very minor secondary value. |
| MODERATE | 4 | Error > 5% and <= 10%. | Incorrect unit string or missing units. | (N/A) |
| MAJOR | 7 | Error > 10%. | (N/A) | Missed a primary requested answer (e.g. found T, forgot Q). |
| CRITICAL | 10 | Numerical value is completely, fundamentally wrong (off by order of magnitude). | (N/A) | Result is physically impossible -> cap score at 2. |


NUMERICAL ERROR BANDS:
correct: error <= 5%
moderate: 5% < error <= 10%
major: error > 10%


FATAL ERRORS:
- Physically impossible result (e.g., negative mass) -> cap score at 2

SCORING RULES:
- If physically impossible: step_score = min(step_score, 2)
- Otherwise step_score = max(0, 10 - total_penalties)

---

FEW-SHOT EXAMPLE:

Input Segment

Question Segment:
A copper rod conducts heat between two surfaces. The heat transfer rate through the rod is calculated to be 520 W.
Also determine the temperature at the rod midpoint, which is 340 K.

Final Answer Provided:
Q = 500 W
T_mid = 340 K

Ground Truth Answer:
Q = 520 W
T_mid = 340 K

Reasoning for Evaluation

The student predicted Q = 500 W while the correct value is 520 W.
error=|500-520|/520 =0.0385
This corresponds to 3.85% error, which is within the <=5% correct band, so the numerical value is acceptable.
The midpoint temperature 340 K exactly matches the ground truth.

Units are correctly provided for both quantities and all requested values are present.
Expected JSON Output
'''
{{
  "predicted_values": [
    {{
      "quantity": "Heat transfer rate (Q)",
      "predicted": 500,
      "ground_truth": 520,
      "unit_predicted": "W",
      "unit_expected": "W",
      "numerical_error": 0.0385,
      "error_band": "correct"
    }},
    {{
      "quantity": "Midpoint temperature (T_mid)",
      "predicted": 340,
      "ground_truth": 340,
      "unit_predicted": "K",
      "unit_expected": "K",
      "numerical_error": 0.0,
      "error_band": "correct"
    }}
  ],
  "units_correct": true,
  "physically_possible": true,
  "physical_impossibility_reason": null,
  "all_parts_answered": true,
  "errors": [],
  "total_penalties": 0,
  "raw_score": 10,
  "capped_score": 10,
  "justification": "Both numerical values are correct within tolerance and units are properly provided."
}}
'''
---

OUTPUT FORMAT (JSON ONLY):

{{
  "predicted_values": [
    {{
      "quantity": "<what is being predicted>",
      "predicted": <numerical value or null if not found>,
      "ground_truth": <numerical value or null if not provided>,
      "unit_predicted": "<unit provided>",
      "unit_expected": "<expected unit>",
      "numerical_error": <calculated relative error or null>,
      "error_band": "<correct|moderate|major|not_comparable>"
    }}
  ],
  "units_correct": <true|false>,
  "physically_possible": <true|false>,
  "physical_impossibility_reason": "<reason if physically impossible, else null>",
  "all_parts_answered": <true|false>,
  "errors": [
    {{
      "description": "<what the error is>",
      "category": "<Numerical Correctness | Units & Presentation | Completeness>",
      "severity": "<MINOR|MODERATE|MAJOR|CRITICAL>",
      "penalty": <2, 4, 7, or 10>
    }}
  ],
  "total_penalties": <sum>,
  "raw_score": <max(0, 10 - total_penalties)>,
  "capped_score": <min(raw_score, 2) if physically impossible, else raw_score>,
  "justification": "<detailed explanation explicitly comparing the final numerical values and units against the GROUND TRUTH REFERENCE images>"
}}

\end{lstlisting}

\subsubsection{Meta-Eval: Verbosity Prompt}
\label{app:verbosity_prompt}
\begin{lstlisting}
You are an extremely strict engineering exam grader evaluating the VERBOSITY of the COMPLETE LLM-generated solution to a graduate-level engineering problem.

This is a meta-evaluation. You are checking the ENTIRE solution to ensure it is concise, direct, and free of unnecessary filler or repetition.

---

QUESTION:
{question_text}

COMPLETE SOLUTION:
{step_content}

GROUND TRUTH REFERENCE:
{ground_truth}

---

VERBOSITY CHECKS:

1. REPETITION & RESTATEMENT
   - Does the solution unnecessarily restate the entire question before starting?
   - Does it repeat the same conclusion multiple times across different steps?
   - Are equations written out repeatedly without any new substitution or derivation?
   - CRITICAL: The solution is REQUIRED to follow a strict tagged format (e.g., ###### PROBLEM_CHARACTERIZATION ######, ###### ASSUMPTIONS ######). Do NOT penalize the existence of these sections as "boilerplate" or "filler". They are mandatory.

2. FILLER TEXT & OVER-EXPLANATION
   - Is there excessive conversational filler ("As we can clearly see...", "It is important to note that...")?
   - Are trivial algebraic steps over-explained in paragraphs of text?
   - Could the solution be significantly shorter without losing any technical rigor?
   - Are there any extra assumptions, algebraic steps which are not actually needed?

---

GRADING RUBRIC MATRIX (PENALTIES):

| Severity | Points | Criterion 1: Repetition & Restatement | Criterion 2: Filler Text & Over-Explanation |
| :--- | :---: | :--- | :--- |
| MINOR | 2 | Small amount of repetitive text; stated the obvious once. | A few conversational filler phrases. |
| MODERATE | 4 | Repeatedly restated the given parameters in multiple steps; repeated the final answer unnecessarily. | Over-explained a simple algebraic manipulation with a paragraph; extra unnecessary assumptions, algebraic steps. |
| MAJOR | 7 | Restated the entire multi-paragraph question verbatim before solving; circular repetitive logic. | Heavy reliance on conversational filler; reads like an essay rather than an engineering solution. |
| CRITICAL | 10 | Solution is overwhelmingly padded with redundant text, burying the actual mathematical steps entirely. | Almost entirely filler text with minimal technical content. |

SCORING: score = max(0, 10 - total_penalties)

---

FEW-SHOT EXAMPLE:

Input Segment: 
Solution:
As we can see from the extremely detailed problem statement provided above, which states that we have a pipe of radius 0.5m and flow of 10m/s, it is incredibly vital and important to first calculate the Reynolds number to see if the flow is laminar or turbulent. Laminar flow is smooth, while turbulent flow is chaotic. We must calculate this important parameter.
Re = rho * V * D / mu = 1000 * 10 * 1 / 0.001 = 10,000,000.
Since 10,000,000 is much greater than the threshold of 2300, we can conclude with absolute certainty that the flow is indeed very turbulent, which means it is chaotic and well-mixed.

Reasoning for Evaluation:
The student used an extreme amount of conversational filler ("incredibly vital", "absolute certainty") and over-explained basic definitions ("laminar flow is smooth", "well-mixed"). This padding buries the simple Re calculation.

Expected JSON Output:
```json
{{
  "violations": [
    {{
      "description": "Excessive conversational filler, over-explanation of basic fluid mechanics definitions, and unnecessary restatement of given parameters.",
      "category": "Filler Text & Over-Explanation",
      "severity": "MAJOR",
      "penalty": 7,
      "location": "Overall solution"
    }}
  ],
  "total_penalties": 7,
  "raw_score": 3,
  "capped_score": 3,
  "justification": "The solution is heavily padded with conversational filler and introductory text that adds zero technical value."
}}
```

---

OUTPUT FORMAT (JSON ONLY):

{{
  "violations": [
    {{
      "description": "<what the violation is>",
      "category": "<Repetition & Restatement | Filler Text & Over-Explanation>",
      "severity": "<MINOR|MODERATE|MAJOR|CRITICAL>",
      "penalty": <2, 4, 7, or 10>,
      "location": "<which parts of the solution are involved>"
    }}
  ],
  "total_penalties": <sum of all penalties>,
  "raw_score": <max(0, 10 - total_penalties)>,
  "capped_score": <same as raw_score>,
  "justification": "<explanation of verbosity assessment>"
}}

If there are no violations, return an empty violations list and raw_score of 10.

\end{lstlisting}

\subsubsection{Meta-Eval: Coverage Prompt}
\label{app:coverage_prompt}
\begin{lstlisting}
You are an extremely strict engineering exam grader performing a COVERAGE CHECK on the COMPLETE LLM-generated solution to a graduate-level engineering problem.

This is a meta-evaluation. You are checking the ENTIRE solution to verify that ALL parts of the question have been addressed.

---

QUESTION:
{question_text}

COMPLETE SOLUTION:
{step_content}

GROUND TRUTH REFERENCE:
{ground_truth}

---

COVERAGE CHECKS:

1. ALL ASKED QUANTITIES
   - Does the solution compute every primary and secondary quantity requested?
   - If the question asks for multiple values (e.g., "Find T and Q"), are ALL of them computed?

2. SUB-QUESTIONS & SCOPE
   - If the question has parts (a), (b), (c), are ALL parts answered?
   - Does the solution address the full physically described scope (e.g., if there are two connected pipes, are both analyzed)?

3. RELEVANT ANALYSIS & RESULTS
   - Are requested diagrams/plots mentioned or described?
   - Are final numerical answers clearly provided rather than just symbolic equations?

---

GRADING RUBRIC MATRIX (PENALTIES):

| Severity | Points | Criterion 1: Asked Quantities | Criterion 2 & 3: Scope & Analysis |
| :--- | :---: | :--- | :--- |
| MINOR | 2 | Missing units on one secondary result. | Missed a minor implicit detail ("comment on result"). |
| MODERATE | 4 | Missed plotting/sketching entirely when requested. | Addressed part (b) but only implicitly; missing units on multiple results. |
| MAJOR | 7 | Forgot to compute one of the main requested variables (e.g. found T, forgot Q). | Entire sub-part (e.g. Part B) of the question completely ignored. |
| CRITICAL | 10 | Answered the wrong question entirely. | Majority of what was asked is not addressed at all. |

SCORING: score = max(0, 10 - total_penalties)

---

FEW-SHOT EXAMPLE:

Input Segment: 
Solution computes the temperature distribution T(x) perfectly, but stops there.
Problem Statement: "Determine the temperature distribution T(x) and calculate the total heat transfer rate Q at the surface x=0."

Reasoning for Evaluation:
The problem explicitly asks for two things: T(x) and Q. The student completely forgot to calculate Q. This is a major coverage omission of a primary requested quantity.

Expected JSON Output:
```json
{{
  "violations": [
    {{
      "description": "Failed to calculate the total heat transfer rate Q at the surface x=0.",
      "category": "1. Asked Quantities",
      "severity": "MAJOR",
      "penalty": 7,
      "location": "End of solution / overall"
    }}
  ],
  "total_penalties": 7,
  "raw_score": 3,
  "capped_score": 3,
  "justification": "The student perfectly solved the first half of the question but completely ignored the request for the heat transfer rate Q."
}}
```

---

OUTPUT FORMAT (JSON ONLY):

{{
  "violations": [
    {{
      "description": "<what is missing>",
      "category": "<Asked Quantities | Scope & Analysis>",
      "severity": "<MINOR|MODERATE|MAJOR|CRITICAL>",
      "penalty": <2, 4, 7, or 10>,
      "location": "<where it should have been>"
    }}
  ],
  "all_parts_covered": <true|false>,
  "total_penalties": <sum of all penalties>,
  "raw_score": <max(0, 10 - total_penalties)>,
  "capped_score": <same as raw_score>,
  "justification": "<detailed explanation explicitly comparing the solution's coverage against the required parts and quantities in the GROUND TRUTH REFERENCE images>"
}}

If all parts are fully covered, return an empty violations list, all_parts_covered=true, and raw_score of 10.

\end{lstlisting}

\subsubsection{Meta-Eval: Physical Sanity Prompt}
\label{app:physical_sanity_prompt}
\begin{lstlisting}
You are an extremely strict engineering exam grader performing a FINAL PHYSICAL SANITY CHECK on an LLM-generated solution.

This is a meta-evaluation. You must determine whether the predicted engineering quantities across the entire solution are PHYSICALLY REASONABLE.

If any predicted quantity is physically impossible, the ENTIRE solution score will be heavily penalized, potentially capped.

---

QUESTION:
{question_text}

COMPLETE SOLUTION:
{step_content}

GROUND TRUTH REFERENCE:
{ground_truth}

---

PHYSICAL SANITY CHECKS:

1. SIGN CHECKS & LIMITS
   - Density, Mass, Absolute Temperature (> 0 K), Thermal conductivity, Viscosity must be positive.
   - Efficiency of heat engines must be <= Carnot efficiency.
   - Heat cannot spontaneously flow from cold to hot.

2. MAGNITUDE REASONABLENESS
   - Velocities should not exceed speed of light for non-relativistic problems.
   - Stresses should not exceed ultimate strength of specified materials ridiculously.
   - Pressures should be physically meaningful.

3. CONSERVATION LAWS
   - Mass, Energy, Momentum must be conserved.

---

GRADING RUBRIC MATRIX (PENALTIES):

| Severity | Points | Criterion 1: Signs & Limits | Criterion 2 & 3: Magnitudes & Conservation |
| :--- | :---: | :--- | :--- |
| MINOR | 2 | Small violation of an assumption boundary. | Used gauge pressure instead of absolute in a minor non-ideal gas calculation. |
| MODERATE | 4 | T calculated as slightly negative Celsius when expected hot. | Stress exceeds yield strength but student does not notice. |
| MAJOR | 7 | Claimed heat flows from cold to hot; Efficiency > Carnot. | Mass is clearly not conserved by a factor of 2. |
| CRITICAL | 10 | Negative absolute temperature (K); Negative mass or density. | Energy created from nothing; efficiency > 100%. |

SCORING: score = max(0, 10 - total_penalties)

---

FEW-SHOT EXAMPLE:

Input Segment: 
T2 = 300 K - 500 K = -200 K.
The final temperature of the gas is -200 Kelvin.

Reasoning for Evaluation:
Absolute temperature (Kelvin) cannot be negative. This is a fundamental physical impossibility and a critical sanity check failure.

Expected JSON Output:
```json
{{
  "violations": [
    {{
      "quantity": "Final Temperature T2",
      "value": "-200 K",
      "reason": "Absolute temperature measured in Kelvin cannot be negative.",
      "category": "1. Signs & Limits",
      "severity": "CRITICAL",
      "penalty": 10
    }}
  ],
  "physically_reasonable": false,
  "sanity_check_passed": false,
  "total_penalties": 10,
  "raw_score": 0,
  "capped_score": 0,
  "justification": "The student calculated a negative absolute temperature, which is physically impossible."
}}
```

---

OUTPUT FORMAT (JSON ONLY):

{{
  "violations": [
    {{
      "quantity": "<what quantity is problematic>",
      "value": "<the problematic value>",
      "reason": "<why it is physically impossible or unreasonable>",
      "category": "<Signs & Limits | Magnitudes & Conservation>",
      "severity": "<MINOR|MODERATE|MAJOR|CRITICAL>",
      "penalty": <2, 4, 7, or 10>
    }}
  ],
  "physically_reasonable": <true|false>,
  "sanity_check_passed": <true|false>,
  "total_penalties": <sum>,
  "raw_score": <max(0, 10 - total_penalties)>,
  "capped_score": <same as raw_score>,
  "justification": "<detailed explanation explicitly comparing the solution's physical results and trends against the GROUND TRUTH REFERENCE images and known physical limits>"
}}

If there are no violations, return an empty violations list and set flags to true, raw_score=10.

\end{lstlisting}

\section{Computational Cost and Infrastructure}
\label{app:comp_cost}
The proposed \texttt{EngJudge} evaluation framework utilizes isolated, step-wise prompts to evaluate solutions, which increases grading accuracy but introduces different computational requirements depending on the deployment configuration. We benchmark the computational costs of two evaluation pipelines: an API-driven setup utilizing Google Vertex AI (\texttt{gemini-3.1-pro-preview}) and a local GPU-based setup utilizing Qwen models (\texttt{Qwen3-VL-32B} and \texttt{Qwen3-VL-8B}).

\subsection{API-Based Gemini Evaluation Cost}
For our primary API benchmark, the evaluator (\texttt{gemini-3.1-pro-preview}) makes 11 separate API queries per question (8 stage-specific runs and 3 meta-evaluation runs). The cumulative evaluation required 10.80 hours of continuous API execution time, consuming $36.46$M input tokens and generating $1.77$M output tokens in total. Preprocessing costs include Adobe PDF Services API remote extraction (5--15 seconds per document) and local neural LaTeX OCR (\texttt{pix2tex}) equation inference (2--3 seconds per image on a standard CUDA GPU). A simplified breakdown of the average time and input token consumption per question under the Gemini evaluator is summarized in Table~\ref{tab:computational_cost}.

\begin{table}[h]
\centering
\small
\begin{tabular}{lcc}
\toprule
\textbf{Subject Domain} & \textbf{Avg. Time/Q} & \textbf{Avg. Input Tokens} \\
\midrule
MoM & 55.13 s & 65,085.99 \\
Dyn & 59.65 s & 55,459.05 \\
FM  & 58.90 s & 53,239.72 \\
HMT & 58.41 s & 63,825.43 \\
Thermo & 58.72 s & 45,864.77 \\
\midrule
\textbf{Overall / Average} & \textbf{58.41} & \textbf{54,746.46} \\
\bottomrule
\end{tabular}
\vspace{0.3cm}
\caption{Simplified computational cost breakdown per engineering subject using \texttt{gemini-3.1-pro-preview} as the judge model.}
\label{tab:computational_cost}
\end{table}

\subsection{Local GPU-Based Qwen Evaluation Cost}
In addition to the API-based evaluator, we evaluate a local GPU-driven pipeline. Candidate solution generation is performed using \texttt{Qwen3-VL-8B} on an NVIDIA L40 GPU, and step-wise evaluation is performed using \texttt{Qwen3-VL-32B} on an NVIDIA H100 GPU. 

The evaluation runtime for a 100-sample test suite was 6 hours using \texttt{Qwen3-VL-32B} (averaging 3.6 minutes per question), while the candidate generation runtime was 5 hours using \texttt{Qwen3-VL-8B} (averaging 3.0 minutes per question). The local deployment consumed a total of 11 GPU-hours (6 H100 GPU-hours and 5 L40 GPU-hours).

\subsection{Evaluator Hyperparameters and Infrastructure}
\label{app:evaluator_params}
The step-wise evaluation judge is primarily driven by \texttt{gemini-3.1-pro-preview} hosted on Google Cloud Vertex AI, and local Qwen models (\texttt{Qwen3-VL-32B} and \texttt{Qwen3-VL-8B}) run via a local Hugging Face pipeline. The generation configurations and environments used for the automated grading and generation steps are detailed below:

\begin{itemize}[leftmargin=*]
    \item \textbf{Gemini Configurations:}
    \begin{itemize}
        \item \textbf{Temperature:} $0.2$. A low temperature was selected to ensure deterministic scoring behavior and reduce evaluator variance.
        \item \textbf{Max Output Tokens:} $8,192$ tokens.
        \item \textbf{Thinking/Reasoning Budgets:} Disabled (\texttt{VERTEX\_ENABLE\_THINKING = false}) to evaluate direct grading performance.
        \item \textbf{JSON Output Validation:} An automated regular-expression and bracket-matching recovery parser is applied to clean and restore incomplete JSON responses resulting from context window limits.
    \end{itemize}
    \item \textbf{Qwen Configurations:}
    \begin{itemize}
        \item \textbf{Temperature:} $0.7$. For the Qwen models, the default pipeline configuration temperature of $0.7$ was preserved. Unlike the evaluation-specific low-variance constraint set for Gemini, the default configuration was retained to maintain the models' standard inference behavior.
        \item \textbf{Max Output Tokens:} $8,000$ tokens.
    \end{itemize}
\end{itemize}

\section{Use of LLMs}
In this work, LLMs were used in two ways: 
\begin{itemize}
    \item Grammar checking and language polishing during paper writing
    \item Serving as the models for generation and evaluation judges for rubric-based scoring.
\end{itemize}

\end{document}